\documentclass[12pt]{report}

\usepackage{pdfpages}
\usepackage{lipsum}
\usepackage[top=1in, bottom=1in, inner=1.5in, outer=1in,twoside]{geometry}
\usepackage[hidelinks,urlcolor=blue]{hyperref}
\usepackage{url}
\usepackage{graphicx}
\usepackage{float}
\usepackage[none]{hyphenat}
\usepackage{xfrac}
\usepackage{pgfplots}
\newlength\figureheight
\newlength\figurewidth
\usepackage{gensymb}
\usepackage{pdfpages}
\usepackage{overcite}
\usepackage{tabularx}
\usepackage{amsmath}
\usepackage[parfill]{parskip}

\usepackage{tikz}
\usetikzlibrary{arrows, positioning}
\usepackage{tikz-3dplot}
\usepackage{leftidx}

\usepackage{enumitem}
\usepackage{subcaption}
\usepackage{pdflscape}

\usepackage[framed,numbered,autolinebreaks,useliterate]{mcode}

\usepackage[nottoc]{tocbibind}

\usepackage[]{algorithm2e}
\usepackage{mdframed}
\newcommand{\listofalgorithmes}{\tocfile{\listalgorithmcfname}{loa}}

\newcommand{\subsubsubsection}[1]{\begin{small}\noindent\textit{\textbf{#1}}\end{small}\\}

\newcommand{\degr}{$^o$ }
\newcommand{\attributions}[1]{\begin{mdframed}\textit{\textbf{Attributions: }}\textit{#1}\end{mdframed}}

\usepackage{pagecolor,lipsum}
\definecolor{grey}{rgb}{0.3, 0.3, 0.3}
\definecolor{default}{rgb}{0, 0, 0}

\usepackage{appendix}
\usepackage{titlesec}
\titlespacing\chapter{0pt}{12pt}{0pt}

\usepackage{tocloft}
\usepackage{setspace}
\usepackage{fancyhdr}

\begin{document}
\pagecolor{white}
\color{default}
	\cleardoublepage
	\begin{titlepage}
		\begin{center}
			
			\vspace{-1cm}
			\vspace{-0.8cm}
			
			\line(1,0){450}\\
			\LARGE{\bfseries Hybrid Aerial-Ground Vehicle Autonomy\\ in GPS-denied Environments}
			
			\vspace*{-0.5cm}
			\line(1,0){450}\\
			\vspace{1cm}
			\textsc{\Large Tara Bartlett, 450198331}\\
			\vspace{1cm}

			\begin{table}[H]
				\centering
				\begin{tabular}{rl}
					\large \textbf{Supervisor:} & \large Assoc. Prof. Peter Gibbens \\
					\large \textbf{Advisors:}    & \large Dr. Ali-akbar Agha-mohammadi               \\
					& \large Mr. Rohan Thakker         
				\end{tabular}
			\end{table}
			
			\begin{figure}[H]
				\centering
				\includegraphics[width=0.4\linewidth]{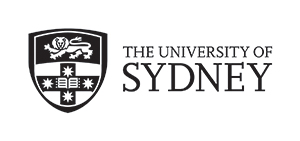}
			\end{figure}
			\normalsize
			\textsc{\large{School of Aerospace, Mechanical and Mechatronic Engineering\\}}
			\vspace{1.5cm}
			A thesis submitted to The University of Sydney in partial fulfilment of the requirements for the degree of Bachelor of Engineering Honours (Aeronautical) (Space))\\
			
			\vspace{1cm}
			This research was carried out at the Jet Propulsion Laboratory, California Institute of Technology, and was sponsored by the University of Sydney Industry Placement Scholarship and the National Aeronautics and Space Administration.
		\end{center}
	\end{titlepage}
	
	\pagenumbering{roman}
	
	\chapter*{Declaration of Authorship}\addcontentsline{toc}{chapter}{Declaration of Authorship}

I, Tara Bartlett, declare that this thesis titled Hybrid Aerial-Ground Vehicle Autonomy in GPS-denied Environments and the work presented in it are my own.

I confirm that:
\begin{itemize}
	\item The work was completed entirely while in candidature for a Bachelor of Engineering at the University of Sydney.
	\item When applicable, consultation of published literature has been clearly attributed and referenced.
	\item The work I have completed myself in this thesis is as follows:
	\begin{itemize}
		\item I completed a literature review into perception and global localisation, collision avoidance, navigation without localisation, and traversability analysis methods.
		\item I integrated a stereo and depth camera with an existing local planning architecture
		\item I improved and thoroughly tested the robustness local planning architecture and integrated it with the hardware and controller
		\item I developed a wall following algorithm and tested this on the vehicle
		\item I aided in the development and testing of the hybrid wall following development
		\item I developed a traversability analysis algorithm and completed handheld tests
		\item I developed a multiple-environment unit-test simulator, except for the 3D model of the vehicle and a plugin for the sensor
		\item I worked with a team to collaboratively test all experiments which were towards this thesis and the project at the NASA Jet Propulsion Laboratory
	\end{itemize}
\end{itemize}

\begin{centering}
	\begin{table}[H]
		\begin{tabular}{lll}
			\textbf{Author (Tara Bartlett)} &  & \textbf{Supervisor (A/Prof. Peter W. Gibbens)} \\
		&  &                                       \\
		&  &                                       \\ \cline{1-1} \cline{3-3} 
		&  &                                       \\
		\textbf{Date}                   &  & \textbf{Date}                                  \\
		&  &                                       \\
		&  &                                       \\ \cline{1-1} \cline{3-3} 
		\end{tabular}
	\end{table}
\end{centering}

\section*{Acknowledgements}\addcontentsline{toc}{chapter}{Acknowledgements}
I would like to acknowledge the work of the Co-STAR Team of the DARPA Subterranean Challenge, especially the other members of the Rollocopter Guidance and Control team, consisting of Rohan Thakker, David Fan, and Meriem Ben Miled. The development of the planning and control systems was a collaborative effort and all testing in this thesis was conducted by this sub-team.

All other subsystems, upon which the Guidance and Control work depends, were developed by the wider Co-STAR team.
	\cleardoublepage
	\chapter*{Abstract}\addcontentsline{toc}{chapter}{Abstract}
The DARPA Subterranean Challenge is leading the development of robots capable of mapping underground mines and tunnels up to 8km in length and identify objects and people. Developing these autonomous abilities paves the way for future planetary cave and surface exploration missions. The Co-STAR team\cite{costar}, competing in this challenge, is developing a hybrid aerial-ground vehicle, known as the Rollocopter. The current design of this vehicle is a drone with wheels attached. This allows for the vehicle to roll, actuated by the propellers, and fly only when necessary, hence benefiting from the reduced power consumption of the ground mode and the enhanced mobility of the aerial mode. This thesis focuses on the development and increased robustness of the local planning architecture for the Rollocopter.

The first development of thesis is a local planner capable of collision avoidance. The local planning node provides the basic functionality required for the vehicle to navigate autonomously. The next stage was augmenting this with the ability to plan more reliably without localisation. This was then integrated with a hybrid mobility mode capable of rolling and flying to exploit power and mobility benefits of the respective configurations. A traversability analysis algorithm as well as determining the terrain that the vehicle is able to traverse is in the late stages of development for informing the decisions of the hybrid planner. A simulator was developed to test the planning algorithms and improve the robustness of the vehicle to different environments.

The results presented in this thesis are related to the mobility of the rollocopter and the range of environments that the vehicle is capable of traversing. Videos are included in which the vehicle successfully navigates through dust-ridden tunnels, horizontal mazes, and areas with rough terrain. Additional results include videos depicting the functionality of the unit test simulator which has already been used in improving the robustness of the methods used for navigating without localisation.
	\cleardoublepage
	\chapter*{Executive Summary}\addcontentsline{toc}{chapter}{Executive Summary}
With the increasing use of drones comes more potential applications to different fields. The DARPA Subterranean Challenge is encouraging the development of robots capable of mapping underground mines and tunnels up to 8km in length and identify objects and people. Developing these autonomous abilities paves the way for future planetary cave and surface exploration missions. The Co-STAR team\cite{costar} competing in this challenge is developing a hybrid aerial-ground vehicle, known as the Rollocopter, capable of long duration missions in unknown extreme environments. The current design of this vehicle is a drone with wheels attached. This allows for the vehicle to roll, actuated by the propellers, and fly only when necessary, hence benefiting from the reduced power consumption of the ground mode and the enhanced mobility of the aerial mode. This thesis focuses on the development and increased robustness of the local planning architecture for the Rollocopter.

A literature review was completed to provide some background into methods which have been developed and tested in previous studies. This covered common perception and planning algorithms for drones and ground robots in addition to approaches used for navigation without localisation and terrain classification. The conclusions drawn from the comparisons of the different methods helped inform the decisions made during the development completed in this thesis and provided a baseline from which to improve.

The first development of thesis is a local planner capable of collision avoidance. The local planning node provides the basic functionality required for the vehicle to navigate autonomously by avoiding obstacles and planning short-term paths towards a specified goal. The main developments through this section of the thesis was to integrate a previously written planning architecture with the vehicle and make the algorithms more robust by testing on hardware and improving the system based on the findings. The system works with instantaneous point clouds which allows for the vehicle to plan without a reliance on localisation.

The next stage was the enhancement of the local planner to navigate more reliably without localisation by exploiting the structure of the underground environment. Hence, the local planner was augmented to support a wall following behaviour. This was first created for the ground mode then extended to the aerial mode and adjusted to reap the full benefits of the increased mobility. This allowed for the vehicle to navigate without localisation while rolling or flying. Finally, the two modes were combined into a hybrid mobility mode capable of rolling and flying to exploit the advantages of each mode.

To fully benefit from the hybrid functionality, the vehicle must also consider the local terrain and thus use an improved estimate of when rolling is more efficient than flying. A node capable of performing traversability analysis was created to provide more information to the planner. This was create to allow for the vehicle to use a local map of the terrain to determine when to roll or fly. The roughness and slope of the terrain, combined with knowledge of the hardware capabilities in this regard, allow for the planner to make more informed decisions of when to transition between the aerial and ground modes. While the development of this section of the project was stopped, the work is still applicable to the vehicle and is likely to be combined with a local map of the environment in future.

Finally, to simplify the process of this development and testing, a low fidelity simulator was developed in which the planning algorithms could be tested. A simple model of the vehicle was created with the sensors required by the local planner. While this tool is not directly applicable to the vehicle, it will provide significant aid in the process of ensuring the planning and autonomy algorithms are robust to different types of environments. The simulator is set up to perform unit tests. A state machine was created such that the developer can set the vehicle to be tested in a specified range of environments and receive information about whether the vehicle was stuck, collided with the environment, or successfully completed the course.

The results presented in this thesis are related to the mobility of the rollocopter and the range of environments that the vehicle is capable of traversing. Videos are included in Section \ref{sec:wall_following} in which the vehicle successfully navigates through dust-ridden tunnels, horizontal mazes, and areas with rough terrain. Additional results include videos depicting the functionality of the unit test simulator which has already been used in improving the robustness of the methods used for navigating without localisation.
	\chapter*{Log of Changes}\addcontentsline{toc}{chapter}{Log of Changes}

\begin{table}[H]
	\begin{tabular}{|l|p{3.5cm}|p{8.5cm}|}
		\hline
		Date          & Version & Change                   \\ \hline
		14 March 2019 & V1.0    & Initial Submission       \\ \hline
		16 April 2019 & V1.1    & Removed a reference in the chapter of Occupational Health and Safety [Requested by JPL] \\ \hline
		23 April 2019 & \flushleft{V1.2 [Primary Version for Submission] }  & Adjusted Reference to the JPL Internal Intranet [Requested by JPL] \\ \hline	
		13 May 2019 &\flushleft{ V1.3 [Primary Version for Publication] }   & Minor adjustments \\ \hline
	\end{tabular}
\end{table}
	\cleardoublepage
	\tableofcontents
	\cleardoublepage
	\newpage

	\phantomsection
	\listoffigures
	
	\newpage
	\phantomsection
	\listoftables
	
	
	\newpage
%
	\newpage
	\listofalgorithmes
	\newpage
	
	\pagenumbering{arabic}
	\setcounter{page}1
	
	\chapter{Introduction}
\section{Thesis Overview}

Autonomy is a rapidly developing field allowing vehicles to explore areas that are unreachable by humans. This thesis is part of larger project in which a hybrid aerial-ground vehicle is to explore and map underground tunnels. This is applicable to exploring caves and tunnels on other planets in the continued search for life\cite{ali_intro1,ali_intro2}. This thesis details the development of a local planner designed to plan around obstacles and follow walls. Due to the requirement of functioning underground, the robots are unable to receive a global positioning system (GPS) signal. This leads to a reliance on alternative forms of localisation, such as vision-based methods, for navigation. However, if the camera cannot identify enough features in a given frame, the algorithm loses track of the position. It is therefore important to have alternative methods of navigation\cite{ali_intro4,ali_intro5} available to the vehicle. This local planner is therefore being designed to work when SLAM loses track. By following the walls, this vehicle is able to navigate through the tunnel when localisation information is unreliable.

The vehicle is designed as a drone with wheels. This allows for the flight and rolling. Rolling allows the drone to traverse areas with smooth ground with a lower power consumption. This is due to the weight of the vehicle being supported by the ground. The ability to fly improves the mobility of the vehicle, providing the option of traversing rougher terrain or patches of rocks. 

To optimise the use of these features, a sophisticated trajectory generation algorithm is required.  In addition, there is a need to localise to be able to follow paths. An algorithm known as simultaneous localisation and mapping (SLAM) can be used to estimate the position and orientation of the vehicle. This helps for keeping track of the location of the vehicle in the world frame, but it is also necessary to be aware of local features near the robot to avoid crashing into obstacles and planning a trajectory between waypoints accordingly. In addition, SLAM relies on extracting features from images. If there is bad lighting, a lack of features, or the images are affected by motion blur, the algorithm can lose track. 

If this occurs, it is necessary to use the latest sensory information for an instantaneous solution that does not rely on temporal fusion. This is because the position and orientation of the vehicle when the sensory information is recorded is unknown with respect to previous measurements. Thus, the information in the current body frame needs to be used to choose an optimal velocity or know when to send the signal for an emergency stop.

This sensory information is be obtained from the sensors on the rollocopter, such as the Intel RealSense RBGD (Red-Green-Blue-Depth) Camera (with stereo and infrared depth capabilities) and a 360\degr Velodyne LIDAR. Both sensors create a point cloud based on the features in the current field of view. For collision avoidance, a set of motion primitives are checked with the point cloud to ensure that there are no obstacles close enough to the path for the robot to collide with. A wall following algorithm was also developed to allow for the vehicle to navigate without localisation.

These algorithms will be usable in a variety of situations, especially where localisation is unavailable. Collision avoidance is a requirement for any autonomous vehicle. This thesis looks into the improvement of the current collision avoidance and trajectory generation algorithms. In addition to making these features more robust and efficient, the wall following algorithm was developed and made to be suitable for a variety of walls from flat man-made walls to natural rough or rocky walls. This was also extended to be used in the hybrid aerial and ground mode to allow for increased mobility and a reduction in power usage.

A strong focus of this thesis was to thoroughly test each of these algorithms and improve the robustness. A low fidelity simulator was developed to allow for testing the autonomy algorithms in a variety of complicated environments including three-dimensional mazes and tunnel networks, doorways, and corridors of different widths.

This thesis will also develop ground traversability analysis methods. The area will be discretised into a grid and a path will be chosen to go through the patches which are more traversable. If there is no traversable path to the next goal, the rollocopter will transition to flight. This will allow for greater mobility while still using less power than would be used flying the whole mission. This will be used for deciding when to transition between the aerial and ground modes.

An interface to the planner was also developed such that a user, and in future a state machine, will be able to command the vehicle to take off, land, fly or drive to a goal, or continue forwards until more information about the environment is available.

\section{Objectives}
The main objectives of this thesis are as follows.
\begin{itemize}
	\item Define motion primitives which can represent appropriate trajectories
	\item Create an algorithm for defining the optimal direction of travel and sending commands to controller
	\item Improve the robustness of a local planner
	\item Follow walls for instantaneous navigation and path selection
	\item Quantify traversability to choose between flying and rolling for a given terrain
	\item Develop a simulator capable of testing the above algorithms and any higher level development
\end{itemize}
\newpage

\section{Report Structure}
This thesis is structured as follows.
\begin{itemize}[leftmargin=1cm,rightmargin=1cm]

\item[] \textit{Chapter 2:  Case Study: The Rollocopter} - 
\indent This chapter provide an overview of the challenge the vehicle is being designed for, and the vehicle itself. This will also discuss how the learning attributes of \textit{Advanced Aircraft Design Analysis} was covered.

\item[]\textit{Chapter 3:  Case Study: Flight Testing and Evaluation} - 
\indent The learning attributes of \textit{Flight Mechanics Advanced Testing and Evaluation} are covered in this section.

\item[]\textit{Chapter 4:  Occupational Health and Safety} - 
\indent This chapter discusses the procedures in place to ensure safety in the workplace.

\item[]\textit{Chapter 5:  Literature Review} - 
\indent This provides an overview of related work for perception and global localisation, collision avoidance, trajectory planning, navigation without localisation, and traversability analysis.

\item[]\textit{Chapter 6:   Local Planning for Collision Avoidance} - 
\indent This chapter explains the the development relating to local planning and collision avoidance.

\item[]\textit{Chapter 7:   Navigation without Localisation} - 
\indent This chapter details the solution for navigating without localisation (wall following) in the aerial, ground and hybrid modes.

\item[]\textit{Chapter 8:   Ground Traversability} - 
\indent Traversability analysis was also a part of this thesis which will be used to help determine when to roll or fly and to locate potential landing sites.

\item[]\textit{Chapter 9:   Autonomy Unit Test Simulator} - 
\indent This simulator was developed for testing the autonomy algorithms in a variety of complicated environments before testing on hardware.

\item[]\textit{Chapter 10:   Conclusion and Future Work} - 
\indent Finally, the contributions of this thesis are highlighted and compared to related work. In addition, conclusions are drawn and future work outlined.

\end{itemize}
	\cleardoublepage
	\chapter{Case Study: The Rollocopter}
\attributions{The content of this chapter is the work of the author of this thesis, except for Section \ref{sec:AircraftDesignProcedures} which contains images created and/or acquired by the JPL SubT Team Hardware Development Manager \cite{matt_anderson}. The design of the rollocopter was also completed by the Co-STAR\cite{costar} team.}
\section{Overview}
This chapter provides detail on the design of the Rollocopter\cite{ali_rollo2,ali_rollo5}, which is being designed for the DARPA Subterranean Challenge. An overview of the design procedure is also provided.

\section{The DARPA Subterranean Challenge}
\begin{figure}[H]
	\centering
	\includegraphics[width=0.6\textwidth]{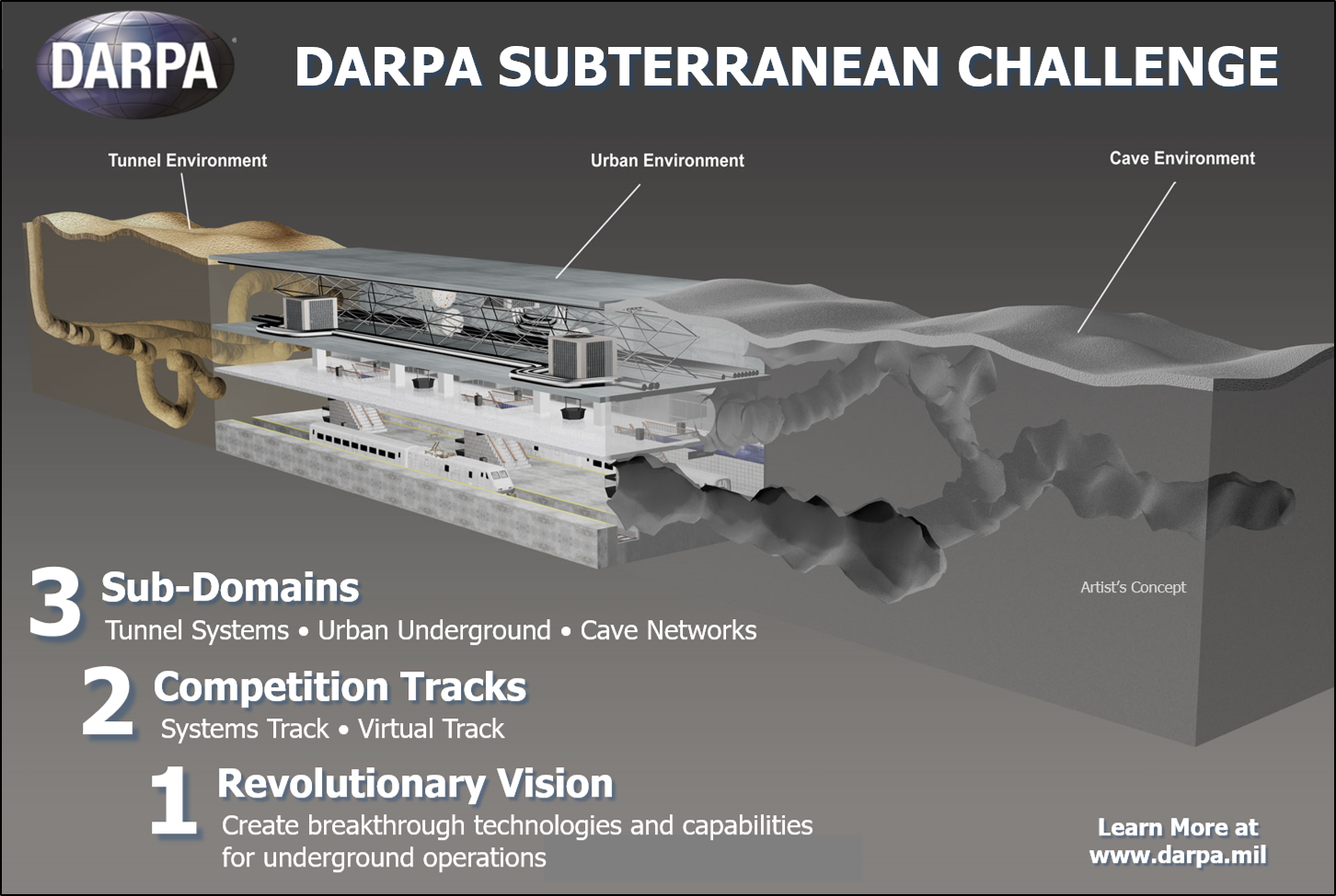}
	\caption{The DARPA Subterranean challenge overview\cite{darpa_subt}}
	\label{fig:darpa_subt}
\end{figure}

The work of this thesis is towards the DARPA Subterranean Challenge\cite{darpa_subt}, where a swarm of robots are required to navigate through an underground environment. The first stage involves navigating through a network of tunnels and identifying objects. A possible application of this would be for search and rescue, for instance. A robot could navigate through a collapsed mine, for example, and return with the locations of survivors. The environments likely to be encountered are illustrated in Figure \ref{fig:darpa_subt}

\section{Vehicle Design}
This section details the hardware-related design of the rollocopter in addition to discussing how the learning attributes of \textit{Advanced Aircraft Design Analysis} were covered through this thesis.

The unit of study covers the following learning attributes.

\begin{itemize}
	 \item Develop an understanding of practical aircraft design processes expected in industry
	 \item Evaluate and  perform case studies of existing aircraft designs
	 \item Develop  modifications to aircraft
	 \item Gain familiarity with international aviation regulations and certification processes will be expected with respect to the design of aircraft
\end{itemize}
 
Hence, this section will discuss the processes used when designing the rollocopter, a case study of the vehicle design, modifications which have been implemented or are to be implemented, and how the design of the vehicle relates to the aviation regulations that this project is governed by.

\newpage
\subsection{Aircraft Design Procedures}\label{sec:AircraftDesignProcedures}
In this project, the vehicle design procedure is as follows.

\begin{enumerate}
	\item When designing a vehicle, the first consideration is the mission. In this case, the vehicle is to transport the sensors from a start location to an end location while collecting data.
	\item One must then discuss the desired sensors and their respective positions on the vehicle with all relevant sub-teams and calculate the total mass of the of the sensors required. An example of this spreadsheet is shown in Figure \ref{fig:mass_table}.
	
	\begin{figure}[H]
		\centering
		\includegraphics[width=0.8\textwidth]{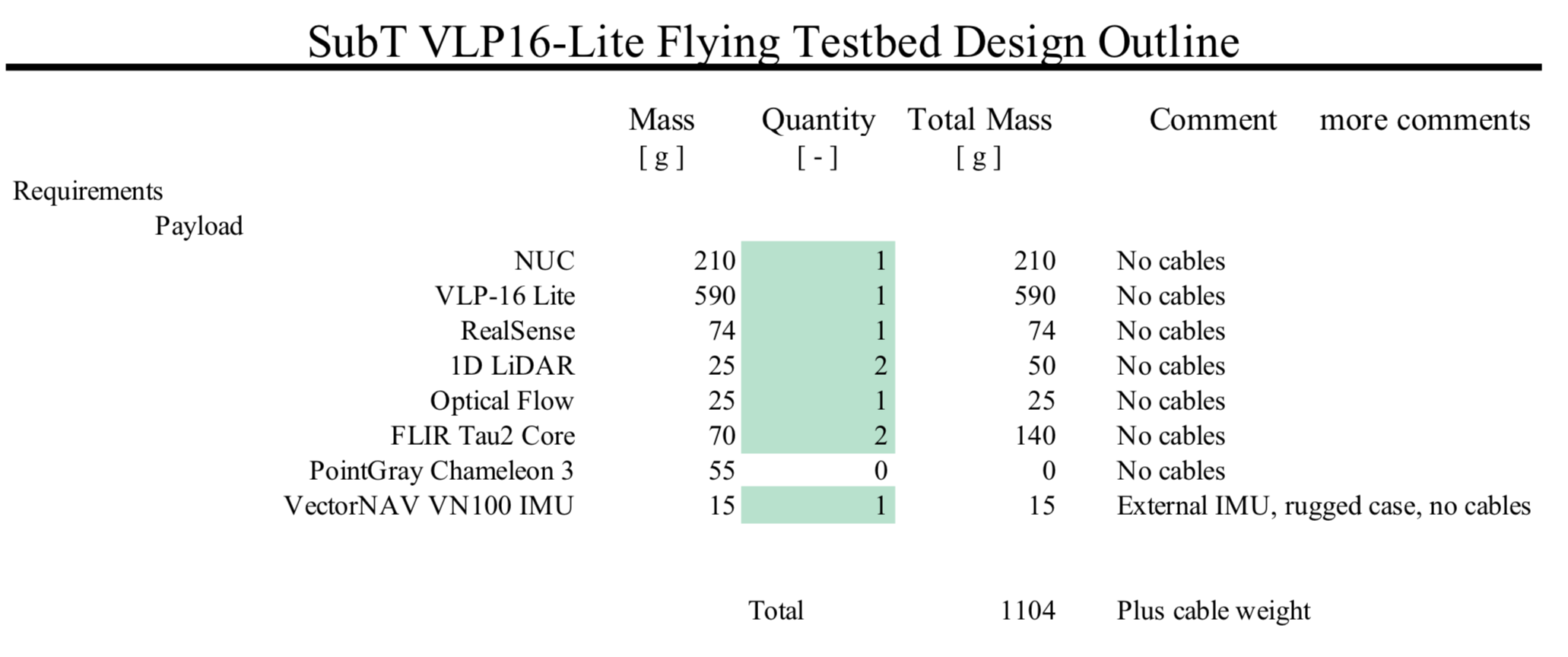}
		\caption{Sample payload mass table for the rollocopter \cite{matt_anderson}}
		\label{fig:mass_table}
	\end{figure}
	
	\item One can then estimate the weight of the vehicle from this by using the rule of thumb that the weight of the aircraft should be approximately three times the weight of the payload (sensors)\cite{matt_anderson}. The other two thirds are reserved for the battery and all components needed for flight (including the frame, motors and propellers).
	\item Next, one must consider the size of the propellers that is required to be able to lift this weight. An efficient way of achieving this is to complete a brief literature search into drones that can carry a similar payload. In the case of the latest design of this vehicle, the Matrice100\cite{drone_matrice100}, the Matrice200\cite{drone_matrice200}, and Alpha Aero Lotus \cite{drone_lotus} were examined.
	\item Once the propeller size has been determined, the size of the vehicle can be chosen. One must note that there must be at least one inch between the propeller disks (the areas the propellers will spin through).
	\item The motor selection can be achieved by considering the desired maximum thrust and the chosen propeller size. A table such as that shown in Figure \ref{fig:tmotors_table} can be used to choose the motor and propeller combination. 
	
	\begin{figure}[H]
		\centering
		\includegraphics[width=0.7\textwidth]{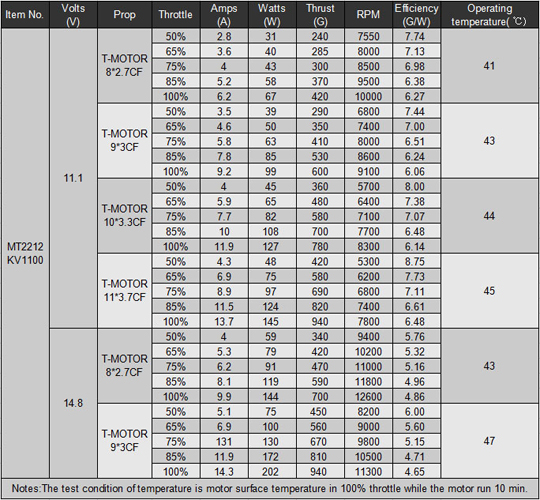}
		\caption{Thrust table for the motors on the rollocopter, TMotor UAV Brushless Motor MT2212 1100Kv\cite{tmotors}}
		\label{fig:tmotors_table}
	\end{figure}
	
	The goal is to have the desired thrust as the thrust value in the table at 60\% throttle for the selected motor and propeller.
	
	The desired thrust can be determined as follows.
	\begin{enumerate}
		\item To ensure control authority over the vehicle, the thrust required to hover should be approximately 60\% of the maximum thrust. Given that the vehicle does not accelerate when hovering, by definition, the hover thrust force is equal to the weight of the vehicle.
		\item Another important consideration is the location of the flight. The thrust values provided for the motors are valid when the vehicle is flown at sea level. Given the lift equation (Equation \ref{eqn:lift_equation}, where $C_L$ is the lift coefficient of the vehicle, $\rho$ is the air density, $V$ is the airspeed of the vehicle, and $S$ is the wetted area).
		\begin{align}
		L = \frac{1}{2}C_L \rho V^2S\label{eqn:lift_equation}
		\end{align}
		\item The desired maximum thrust can then be calculated using Equation \ref{eqn:rollo_desired_max_thrust}, where $T_{max_{SL}}$ is the maximum thrust at sea level and $W$ is the weight of the vehicle.
		\begin{align}
		T_{max_{SL}} = \frac{\rho}{\rho_{SL}}\frac{1}{0.6}\times W\label{eqn:rollo_desired_max_thrust}
		\end{align}
	\end{enumerate}
	\item Finally, one must organise the placement of the components of the vehicle to obtain the desired mass balance. This involves placing all components near the centre of the vehicle, where possible.
\end{enumerate}

\newpage
\subsection{The Rollocopter Hardware Configuration}
This section details the actuation and sensor configurations\cite{ali_rollo3} on the vehicle.

\subsubsection{Actuation}
\begin{figure}[H]
	\centering
	\includegraphics[width=0.5\textwidth]{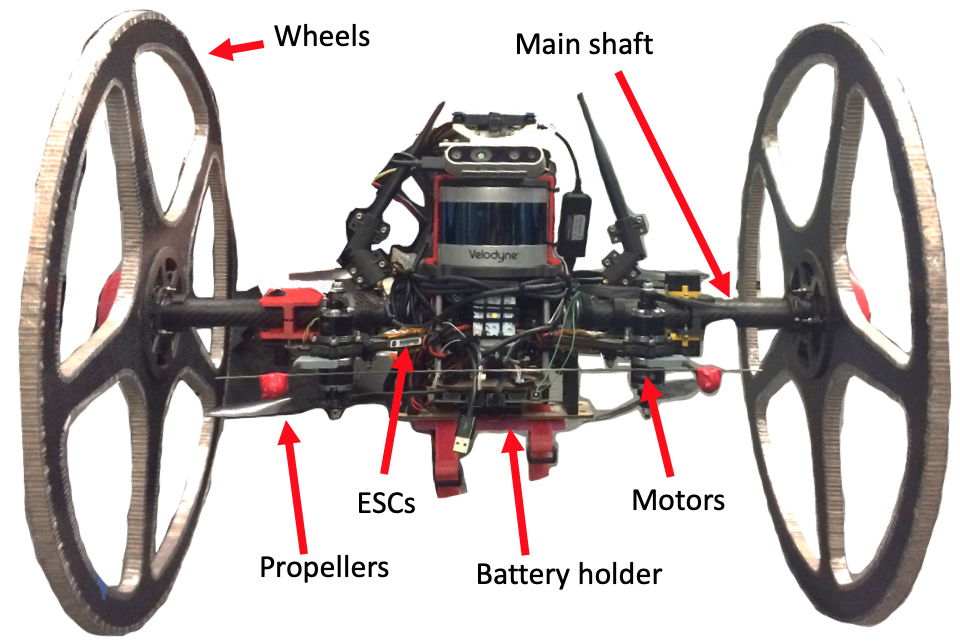}
	\caption{Rollocopter hardware}
	\label{fig:rollocopter_hardware}
\end{figure}

The rollocopter is shown in Figure \ref{fig:rollocopter_hardware}.

As shown, the following components of the rollocopter are used for actuation.

\begin{itemize}[rightmargin=1cm, leftmargin=1cm]
 \item[]\textit{Wheels} - 
  The wheels are made of a carbon-fibre honeycomb structure, with high-traction tape around the edges. These are not actuated, but allow the vehicle to roll forward when provided with a force in the positive x direction from the propellers.

\item[]\textit{Electronic Speed Controllers (ESCs)} - 
 The ESCs are used to control the velocity of the motors by sending the required power, as requested by the flight controller (Pixhawk).

\item[]\textit{Motors} - 
The motors spin the propellers, allowing for thrust to be generated.

\item[]\textit{Propellers} - 
The propellers produce the thrust force when spun, by pulling the air down, through the propellers

\item[]\textit{Battery and holder} - 
 While not specifically for actuation, the battery is held in the velcro straps indicated in Figure \ref{fig:rollocopter_hardware} and is the power source for all actuation and computation.

\item[]\textit{Main shaft} - 
The main shaft connects the wheels to the airframe. While not technically related to actuation, it is an important structural component which allows for the wheels to be part of the vehicle.
\end{itemize}

\subsubsection{Sensors and Perception}
\begin{figure}[H]
	\centering
	\includegraphics[width=0.5\textwidth]{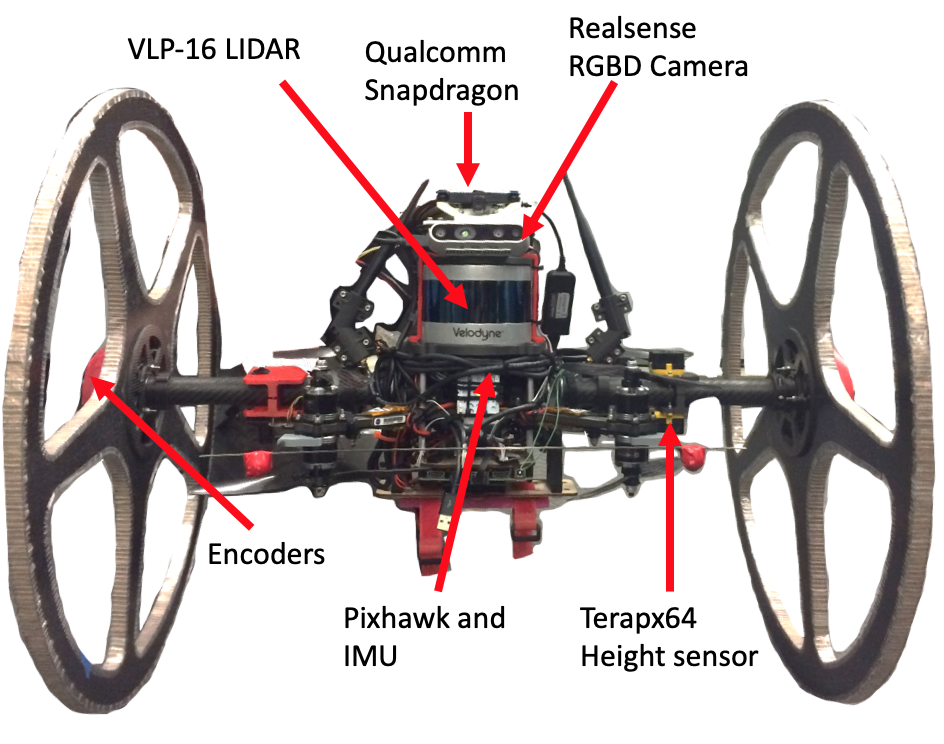}
	\caption{Rollocopter sensors}
	\label{fig:rollocopter_perception}
\end{figure}

The perception subsystem has access to the following sensors. This is shown on Figure \ref{fig:rollocopter_perception}.

The sensors are used as follows.
\begin{itemize}[rightmargin=1cm, leftmargin=1cm]
\item[]\textit{RealSense RGBD Camera} - 
 This camera provides the stereo and RGBD images which can be used for SLAM algorithms such as ORBSLAM.

\item[]\textit{Velodyne 360$^o$ VLP-16 LIDAR} - 
 The LIDAR provides a point cloud around the vehicle with a 360\degr azimuth angle and $\pm15^o$ elevation angle field of view. This is used for collision checking and for LIDAR SLAM algorithms.

 \item[]\textit{LIDAR-based height sensor (Terapx64)} - 
  There are two Terapx64 height sensors on the vehicle, one pointing up and one pointing down. This allows for the vehicle to consider the bottom and top clearances when planning. It is desired that the vehicle maintain a constant distance from the ground while not crashing into the ceiling. These sensors are LIDAR-based and provide a point cloud of 64 points above and below the vehicle. Some traversability analysis can be used to determine the slope and roughness of the terrain and hence can be used to determine if the ground under the vehicle is a good candidate for a landing site.

 \item[]\textit{Pixhawk and the Inertial Measurement Unit (IMU)} - 
  The Pixhawk is the flight controller board and also contains an IMU. This allows for the acceleration and attitude to be determined. This cannot be seen in Figure \ref{fig:rollocopter_perception} because it is at the centre of the vehicle.

 \item[]\textit{Encoders} - 
  The encoders are used to determine the angle of the wheels. When in rolling mode, the angular velocities of the wheels can be estimated, allowing for the yaw rate and forward velocity to be determined.

 \item[]\textit{Qualcomm Snapdragon} - 
  This device is used for SLAM. The current solution for localisation which is used for state estimation is ORBSLAM on the images from the RealSense camera. However, it has been found that this algorithm can easily lose track even with less aggressive manoeuvres. Hence, this algorithm, while closed source, is being tested on the vehicle. A node on the vehicle, known as the \textit{Resiliency Logic} is being created to switch between the different localisation sources depending on which is the most reliable. If the SLAM algorithm on the Snapdragon fails, the SLAM algorithm being used on the LIDAR data may have survived. When in a dusty environment, such as a mine, it has been found that the LIDAR can see some points past the dust while the RGB cameras cannot. Hence, there are some cases where the LIDAR odometry will be more reliable than the Snapdragon odometry.
\end{itemize}
\subsection{Aviation Regulations}
At the NASA Jet Propulsion Laboratory, all flight is regulated by NASA. Given that the vehicles are publicly owned, the Federal Aviation Authority (FAA) regulations, along with the NASA regulations must be adhered to. When testing outside, three people are required. There must be an observer, a pilot in command and, if the vehicle is flying autonomously, a ground station operator. The pilot must be certified by the NASA Armstrong facility. 

There are four different testing cases. The test could be conducted inside or outside, within the JPL campus or at an external location. Each of these cases has different requirements, which are detailed in Chapter \ref{sec:OHS}.

\section{Summary}
The rollocopter hardware configuration was designed as detailed in this chapter to compete in the DARPA Subterranean Challenge and meet the requirements specified above. To ensure the rollocopter is capable of completing the tasks required for the challenge, the software system needed to be developed and tested. This is covered in Chapter \ref{sec:flight_testing}.

\chapter{Case Study: Flight Testing and Evaluation}\label{sec:flight_testing}

\attributions{The safety requirements in Section \ref{rollo_flight_testing} are based on safety regulations learnt at the University of Sydney. All subsystems not covered in other chapters of this thesis were developed by other members of the Co-STAR team\cite{costar}. Section \ref{sec:controller} is based on work written by another member of the Guidance and Control Team\cite{hybrid_paper}. All other work was completed by the author of this thesis.}

\section{Overview}
This chapter covers the flight software system design and the testing procedures used for system identification and autonomy testing. This case study also details how the material in the \textit{Flight Mechanics Advanced Test and Evaluation} subject was covered in work relating to this thesis. 

The following lists the learning attributes as detailed in the handbook on the Course and Unit of Study Portal (CUSP).

\begin{itemize}
	\item Understanding of aircraft flight test, validation and verification
	\item Development of modern flight control, guidance, and navigation systems
	\item Aircraft System Identification and Control
\end{itemize}

This section will discuss aircraft flight testing procedures, an overview of the autonomy subsystems, and aircraft dynamic system identification and control.

\section{Rollocopter Flight Testing}\label{rollo_flight_testing}
Testing hardware can be extremely dangerous, especially when testing autonomous capabilities. Hence, it is important to adopt an incremental approach.

\subsection{Hardware Testing}
When first testing the hardware, the following approach is used.

\begin{enumerate}
	\item Test the connection to the ground station software, such as \textit{QGroundControl}.
	\item Test the commands with the propellers removed. Ensure that the correct motors are responding with the correct intensity and not responding unstably.
	\item Put the propellers back on the vehicle. Start the throttle at a very low setting. This should be approximately half the thrust required to take off.
	\item Hold the vehicle down (with gloves) while testing small roll, pitch and yaw commands. Check that the vehicle attempts the correct motion.
	\item Gradually increase the throttle and repeat the previous step. Stop this test at the slightest sign of instability. Once close to take off thrust, end this test.
	\item In the next test, take off and slowly increase the roll and pitch commands to see how the vehicle responds. This is referred to as envelope expansion. Once completed, the vehicle is ready for use.
\end{enumerate}

\subsection{Autonomy Testing}
When testing the autonomy system, one must also test incrementally. The following testing procedure must be used. 

\begin{enumerate}
	\item Tune the controller. More detail is provided  in Section \ref{sec:controller}.
	\item Next, perform hand held tests with the planner. Check that the correct commands are being sent to the controller.
	\item Test the local planner and the controller together.
	\item Test the mobility services node with the local planner. Test commanding the vehicle to 'Take off', 'Hover, and 'Land'.
	\item Finally, test the autonomy task managing system, known as the BPMN.
\end{enumerate}

Throughout this process, save and check through flight logs and keep track of and fix any interesting issues.

\section{Flight Autonomy Systems}
\begin{figure}[H]
	\centering
	\includegraphics[width=0.5\textwidth]{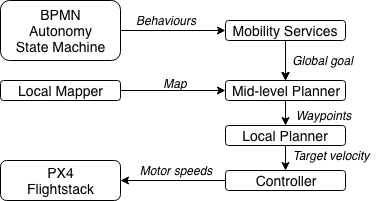}
	\caption{Autonomy system architecture}
	\label{fig:autonomy_system_architecture}
\end{figure}

The autonomy system is structured as shown in Figure \ref{fig:autonomy_system_architecture}.

\subsection{BPMN: Autonomy State Machine}
The 'Business Process Management' (BPMN) node is a system that works like a state machine. Given the current state and task of the system, the BPMN calculates the next desired behaviour of the system. This system is set up such that the following behaviours are possible.

\begin{itemize}
	\item Transition to a frontier (go to a goal position)
	\item Push a frontier forward (navigate forward without an explicitly defined goal position)
	\item Return to base station (based on the energy level of the vehicle and the mission status)
	\item Land (at the ending the mission or once the vehicle has returned to base)
\end{itemize}

These behaviours are then sent to the Mobility Services node.

In future work, this node will utilise previous work to become more sophisticated. For instance, a multi-robot planning architecture will be used\cite{ali_331_1} and advanced motion planning methods will be used to take localization uncertainty into account when planning \cite{ali_planning1,ali_planning2,ali_planning3}.

\subsection{Mobility Services}\label{sec:mobility_services_overview}
Mobility Services uses the behaviour sent from the BPMN to determine the goals for the vehicle and sends these goals to the mid-level planner.

The behaviours are converted into the services detailed below and translated into a series of goals which can be sent to the mid-level planner.

\begin{itemize}[rightmargin=1cm,leftmargin=1cm]
\item[]\textit{Idle Mode} - 
	This mode is used to hold position and is available when transitioning from a ground mode (drive to, drive forward, land).

\item[]\textit{Hover Mode} - 
	This mode is used to hold position and is only available when transitioning from an aerial mode (take off, fly forward, or fly to).

\item[]\textit{Take off} - 
	When in a ground mode (such as idle mode, drive to, or drive forward), the vehicle can be commanded to take off.

\item[]\textit{Land} - 
	Landing can be called from any aerial mode and causes the vehicle altitude to decrease until landed.
	
\item[]\textit{Fly to (x, y, z)} - 
	Once in the air, this mode can be used to fly to a specific coordinate in the world frame.
	
\item[]\textit{Fly Forward} - 
	This mode sends a goal in the body frame to the local planner which leads to wall following.
	
\item[]\textit{Drive to (x, y)} - 
	Like `Fly to', drive to commands the vehicle to roll to a given coordinate in the world frame. This can be called from any ground mode.
	
\item[]\textit{Drive Forward} - 
	Drive forward can also be called from any ground mode and is the wall following mode for when rolling.
\end{itemize}
	
Each of these services defines a goal in the gravity-aligned body frame, or the world frame, which is sent to the local planner for short-term trajectory planning. The algorithms for these services are included in Appendix \ref{apdx:mobility_services}.

The mobility services node was set up as a state machine with each mode being a state of the vehicle and there being transitions defined between the modes, as specified in Table \ref{table:ms_transitions}. Note that this table details the transitions from the mode of the row to the mode of the column. A value of 1 indicates that the transition is possible. If the value is 0, the transition is not possible.

\begin{table}[H]
	\begin{tabular}{|p{0.1\textwidth}|p{0.07\textwidth}|p{0.07\textwidth}|p{0.07\textwidth}|p{0.07\textwidth}|p{0.1\textwidth}|p{0.1\textwidth}|p{0.1\textwidth}|p{0.1\textwidth}|}
		\hline
		\textbf{From/To}	   	& \textbf{Idle} & \textbf{Hover} & \textbf{Take off} & \textbf{Land} & \textbf{Fly to} & \textbf{Fly forward} & \textbf{Drive to} & \textbf{Drive forward} \\ \hline
		\textbf{Idle}          &      -         &      0          &    1              &       0        &       0          &  0     &     1   &   1            \\ \hline
		\textbf{Hover}         &      0         &  -              &    0              &       1        &   1              &   1    &    0    &   0            \\ \hline
		\textbf{Take off}      &      0         &      1          & -                 &        1       &     1            &  1     &   0     &  0             \\ \hline
		\textbf{Land}          &      0         &     1           &    1              &   -            &    1             &  1     &   0     & 0              \\ \hline
		\textbf{Fly to}        &      0         &     1           &     0             &       1        &   1              &    1   &    0    &    0           \\ \hline
		\textbf{Fly forward}   &       0        &     1           &     0             &      1         &   1              & 1      &   0     &    0           \\ \hline
		\textbf{Drive to}      &        1       &    1            &     1             &     0          &    0             & 0      &   1     &   1            \\ \hline
		\textbf{Drive forward} &         1      &     0           &     1             &     0          &  0               &  0     &    1     &  1            \\ \hline
	\end{tabular}
	\caption{Mobility services state machine transition}
	\label{table:ms_transitions}
\end{table}

\subsection{Local Mapper}
\begin{figure}[H]
	\centering
	\includegraphics[width=0.5\textwidth]{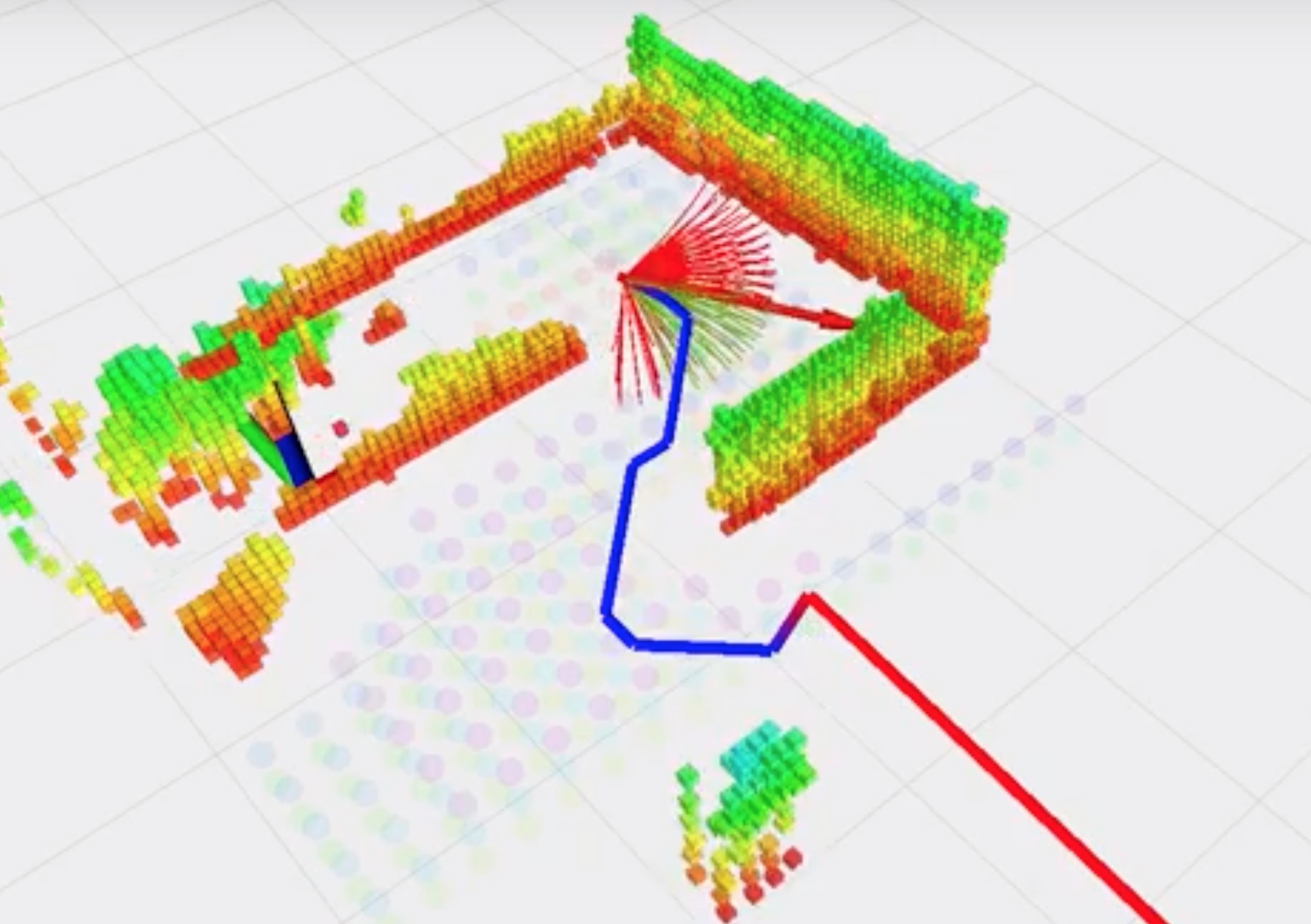}
	\caption{Example of the local map}
	\label{fig:local_map}
\end{figure}

The local mapper creates a sub-map based on the point cloud from the LIDAR. The latest state estimate is used to transform the latest cloud into the map frame. The new cloud is then combined with the previous clouds from within a given radius of the current position. Octomap is used to create a map of the environment with occupied and unoccupied cells, as shown in Figure \ref{fig:local_map}.

In future work, a confidence-rich mapping (CRM) technique will be used to capture the environment uncertainty and plan accordingly\cite{ali_mapping1,ali_mapping2,ali_mapping3}.

\subsection{Mid-level Planner}
The mid-level planner uses A* to plan an optimal path from the latest position of the vehicle to the goal from mobility services, based on the information in the local map. The planner then defines a waypoint which is 1m away from the vehicle, in the direction of the most optimal path.\\

\subsection{Local Planner}
The local planner receives the waypoint from the mid-level planner and generates a series of dynamically-feasible motion primitives from the current position. These primitives are discretised and checked for collisions. The primitive which is most towards the goal and collision free is then followed. The primitive is converted to a target velocity which can be sent to the controller. In future version of this, the planner will have the ability to perform contact-based exploration with a proprioceptive approach in environments where visual and LIDAR sensory information is unusable\cite{ali_rollo4}. \\

\subsection{Controller}
The controller receives the velocity setpoint from the local planner and calculates the desired acceleration accordingly. This acceleration is converted into motor speeds in revolutions per minute (RPM) which is then sent to the PX4 flightstack. The flightstack is run on the Pixhawk which sends the commands to the motors. More detail of this is provided in Section \ref{sec:rollo_controller}.

\section{System Identification and Control}\label{sec:rollo_controller}
This section covers the architecture of the controller and the methods used for system identification.

\subsection{The Controller}\label{sec:controller}
The controller is a key component of the system as the vehicle is then able to follow higher level commands from the autonomy system. The hybrid vehicle has two main controllers which correspond to the two mobility modes. 

\subsubsection{Flying Controller}
\begin{figure}[H]
	\centering
	\includegraphics[width=0.7\textwidth]{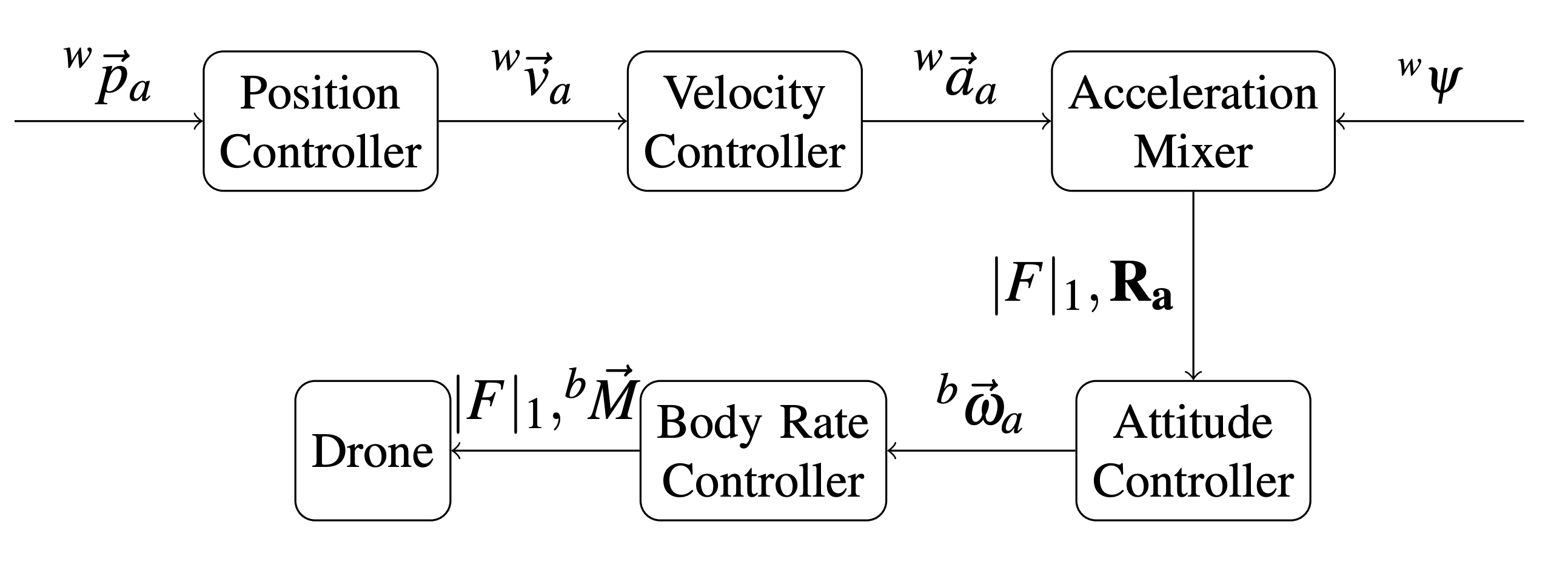}
	\caption{Architecture of the flying controller\cite{px4control}}
	\label{fig:flying_controller}
\end{figure}

The architecture of the controller used for the aerial mode is shown in Figure \ref{fig:flying_controller}.

The controller receives the desired state from the local planner. The desired velocity is then calculated based a proportional gain in the position controller and sent to the velocity controller which calculates the desired acceleration using the same method.

The acceleration mixer calculates the desired thrust, $|F|_1$, and attitude, $\mathbf{R}$, based on desired acceleration, a feed-forward acceleration term and the desired yaw. These values are then sent to the Pixhawk flight controller\cite{px4control}.

\subsubsection{Rolling Controller}
\begin{figure}[H]
	\centering
	\includegraphics[width=0.6\textwidth]{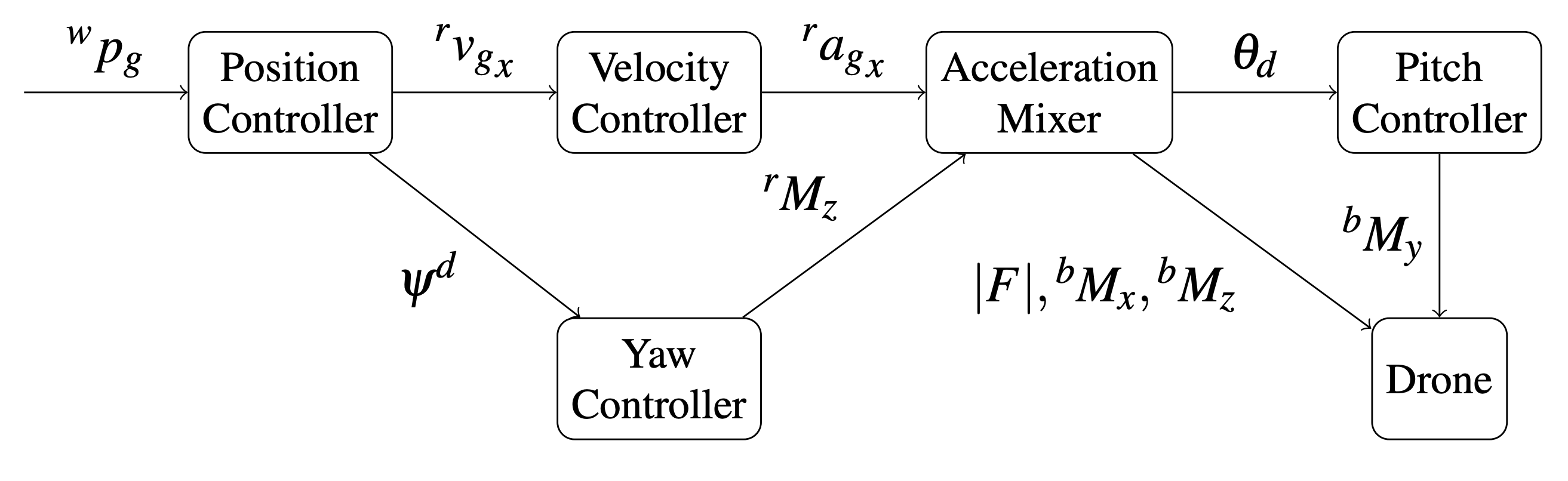}
	\caption{Controller architecture}
	\label{fig:controller_architecture}
\end{figure}

The rolling controller architecture is illustrated in Figure \ref{fig:controller_architecture}.
The main differences to the flying controller are the following
\begin{itemize}
	\item When rolling, there is a non-holonomic constraint, assuming there is not slip. This causes the vehicle to be unable to move in the body y direction and also limits the maximum yaw rate.
	\item It is assumed that the wheels are always in contact with the ground when moving.
\end{itemize}

The following shows the equations of the control law for rolling mode.

\textbf{Position Controller:}
The desired forward body velocity is calculated using a proportional gain on the desired position, as shown in the following equations.
\begin{align}
^we_g &= ^wp_g^d - ^w\hat{p_g} \\
^re_g &= R(\psi_{yaw}) ^we_g\\
{^rv_g}_x &=  ^{pos}k_p (^re_g)
\end{align}

\textbf{Yaw Controller:}
A desired moment is calculated using a PD controller.
\begin{align}
^w\psi^d &= \tan^{-1}({^we_g}_y/{^we_g}_x)\\
^we_{\psi} &= d_{S^1}(^w\psi^d, ^w\psi)\\
^rM_z &= ^{\psi}k_p  (^we_{\psi}) + ^{\psi}k_d(^w\dot{e}_{\psi})
\end{align}

\textbf{Velocity Controller:}
The desired acceleration is calculated based on the desired velocity and a feed-forward velocity term.
\begin{align}
^r\dot{e_g}_x &= ({^rv_g}_x^d + {^rv_g}_x^{ff} ) - {{^r\hat{v}}_g}_x\\
^ra_x &= ^{vel}k_d( ^r\dot{e}_x) + ^{vel}k_I\chi_1
\end{align}

\textbf{Acceleration Mixer:}
The acceleration mixer calculates the desired pitch angles and the desired yaw moment based on the desired acceleration from the velocity controller.
\begin{align}
^b\theta_{pitch}^d &= \sin^{-1}(^ra_x / (m |F|))\\
^bM_x,^bM_z &= R(^b\hat{\theta}_{pitch})^rM_z
\end{align}

\textbf{Pitch Controller:}
The pitch controller then calculates the desired pitch moment based on the calculated error.
\begin{align}
^be_{pitch} &= d_{S(1)}(^b\theta_{pitch}^d,^b\hat{\theta}_{pitch})\\
^bM_y &= ^{pitch}k_p(^be_{pitch})+^{pitch}k_d(-^b\dot{\theta}_{pitch}) + ^{pitch}k_I\chi_2
\end{align}

In the equations above, $^{pos}k_p$, $^{\psi}k_p $, $^{\psi}k_d$, $^{vel}k_d$, $^{\theta}k_p$, $^{\theta}k_d$ and $^{\theta}k_I$ are constant gains on the position, yaw, velocity and pitch respective.$\chi_1$ and $\chi_2$ are the error terms used by the integrator component of the controller.

\newpage
\subsection{System Identification}
When a new version of the vehicle with a slightly different sensor configuration, for instance, was ready for development, the following parameters would be measured and tested.

\begin{itemize}
	\item Mass of the vehicle
	\item Thrust rating
	\item Controller gains
\end{itemize}

The mass of the vehicle would be measured first. Next, the thrust rating would be measured. The thrust rating is the decimal-based proportion of the maximum thrust that is required for the vehicle to maintain a steady altitude.

Given Newton's second law of motion, Equation \ref{eqn:thrust_rating1} can be defined, assuming that the only external forces acting on the vehicle are the gravity force ($F_g$) and the thrust force from the motors ($F_T$). 

\begin{align}
	\Sigma F = ma\\
	F_g + F_T = 0\label{eqn:thrust_rating1}
\end{align}

The thrust rating can then be defined as in Equation \ref{eqn:thrust_rating2}, where $TR$ is the thrust rating, and $T_{max}$ is the maximum thrust available to the vehicle.

\begin{align}
	F_T = -mg\\
	TR = F_T/T_{max}\label{eqn:thrust_rating2}
\end{align}

\begin{figure}[H]
	\centering
	\includegraphics[width=0.9\textwidth]{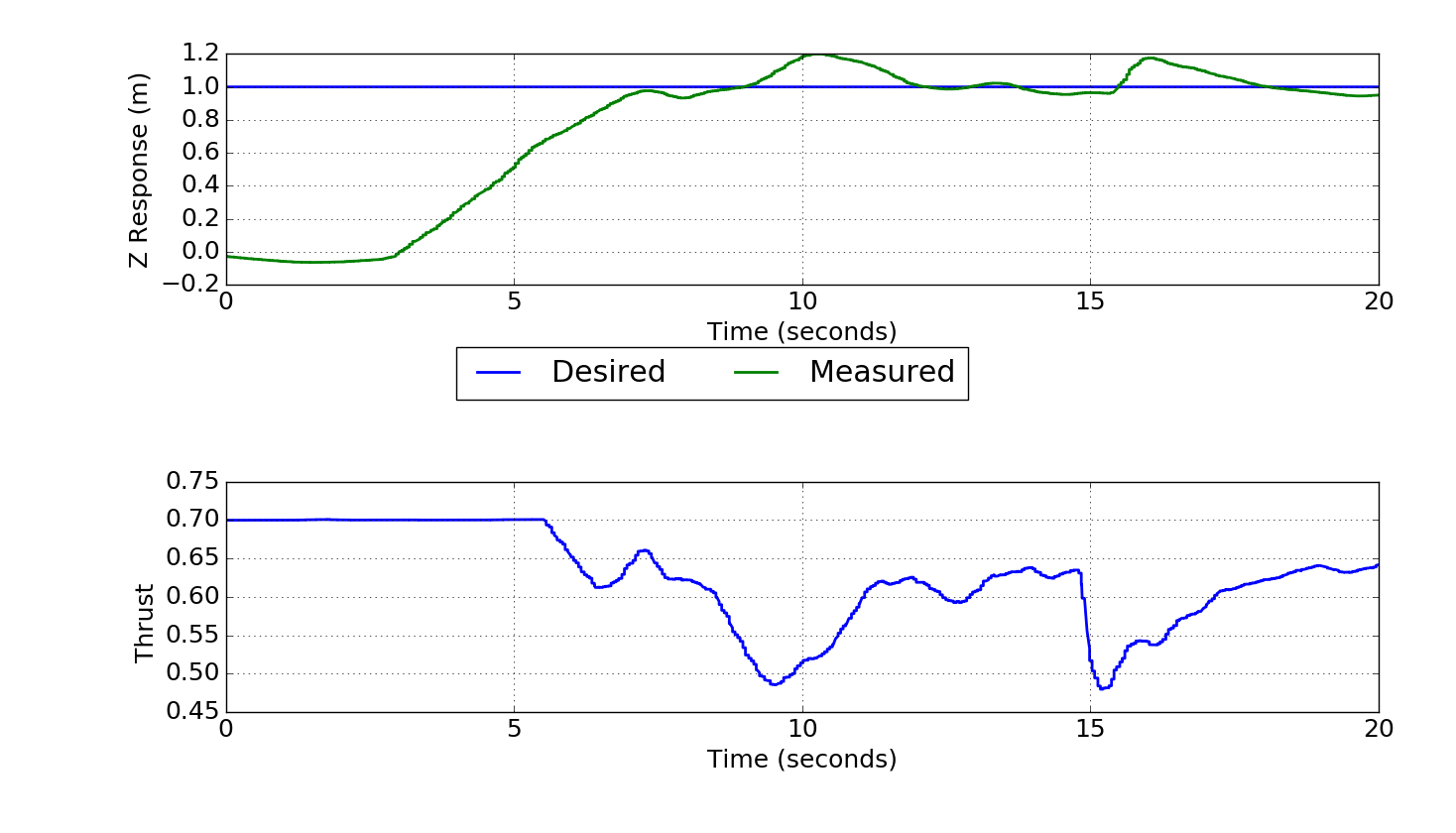}
	\caption{Measuring the thrust rating from a manual flight}
	\label{fig:rollo_system_id_thrust_rating}
\end{figure}

This would be achieved by recording the thrust measured while flying manually. The pilot would increase thrust until the vehicle was able to hover steadily at a given position. The thrust compared to the height is shown for an instance of this test flight, for the latest rollocopter, in Figure \ref{fig:rollo_system_id_thrust_rating}. Comparing the two plots at 12-15s, it is clear that the rollocopter hovering at a constant altitude of 1m occurs with a thrust rating of 0.63.

The controller would then be tested and the gains tuned accordingly. Given that this is a PD controller, there are only two gains per section of the controller which need to be tuned. In addition, this is gains of the previous rollocopter can be used for the initial estimates.

The procedure for this is as follows.

\begin{enumerate}
	\item \textbf{Tune the pitch controller:} Hold the rollocopter steady and send commands to pitch forward and back. 
	\begin{enumerate}
		\item If the rollocopter response is oscillatory, a higher differential gain is required.
		\item If the vehicle is not reaching the desired pitch, a lower differential gain is required.
		\item If the response is too slow, the proportional gain needs to be increased.
		\item If the response is too aggressive, the proportional gain needs to be decreased.
	\end{enumerate}
	\item \textbf{Tune the velocity controller:} Attach the vehicle to a tether and request that the vehicle move forward at a chosen velocity for a specified length of time (e.g. 0.5m/s for 4 seconds) and observe the response.
	\begin{enumerate}
		\item The tuning of the gains corresponds to that of the pitch controller.
		\item Assess the performance of the proportional controller on how the distance travelled compares to the corresponding distance based on the commanded velocity and time.
		\item Assess the performance of the differential controller on how oscillatory the motion is, going forward and backward.
	\end{enumerate}
	\item \textbf{Tune the yaw controller:} Command the vehicle to maintain a constant heading while moving forward on different terrain types, such as flat tile, dirt, and gravel.
	\item \textbf{Tune the position controller:} Test the response to requesting a range of desired positions for the rollocopter.
\end{enumerate}

Following the order above is correct as tuning the response to the pitch controller will adjust the response of the velocity controller. This is because the velocity controller sends commands to the pitch controller. Hence, the response of the velocity controller is heavily dependent on the performance of the pitch controller.

\begin{figure}[H]
	\centering
	\includegraphics[width=0.7\textwidth]{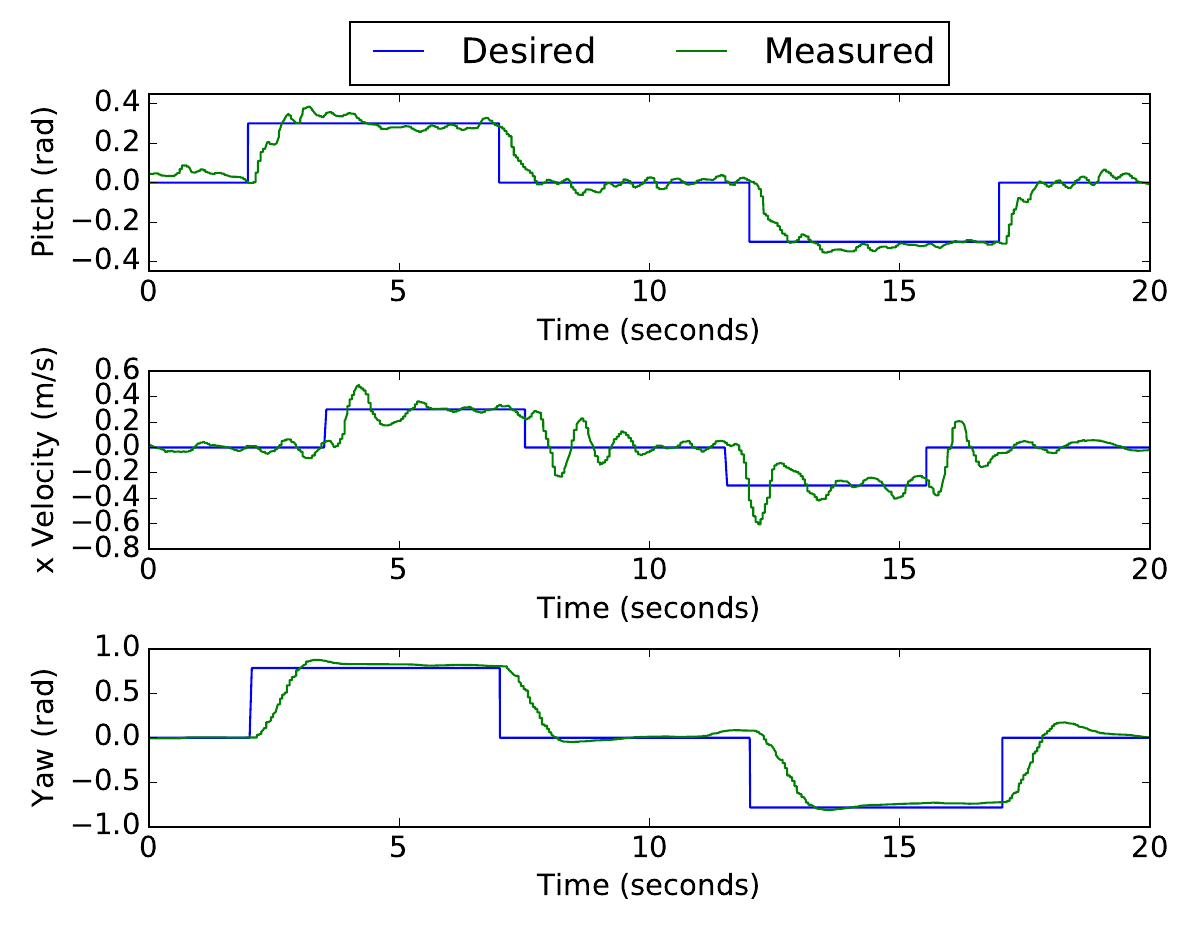}
	\caption{Pitch, velocity, and yaw responses}
	\label{fig:controller_pitch_vel}
\end{figure}

The results of tuning these parameters is shown in Figure \ref{fig:controller_pitch_vel}.

\chapter{Occupational Health and Safety}\label{sec:OHS}
\attributions{This chapter is based on the safety briefings provided to all JPL employees\cite{jpl_rules} and teachings at the University of Sydney.}
\section{Overview}
To ensure the safety of all team members involved in the testing of the previous chapter, the precautions detailed in this chapter need to be followed.

\section{Safety Training}
Each worker is required to undertake the following safety courses.

 \subsubsection{Personal Protective Equipment Overview}
  This session provided an overview of the equipment to use and when it should be used.
	
 \subsubsection{Eye and Face Protection}
 This detailed the cases in which one is required to where glasses and masks. While testing the rollocopter, all observers are required to wear goggles. If the vehicle is connected to a handheld tether or is being used in a handheld test with the propellers being used, the person holding the tether is required to wear a face mask.

\subsubsection{Hand Protection}
This session covered which gloves should be worn during a range of circumstances. The relevant section for our team was that the person holding the tether or rollocopter must be wearing gloves.

\subsubsection{Robotic Safety}
 This covered the importance of various aspects of safety protocols around robots. For instance, it is important to have emergency stop capabilities and designated testing areas that observers must not enter during tests.

\subsubsection{Electrical Safety} 
The electrical safety course covered the required procedures for avoiding electrical fires. Notable points include not connecting multiple multi-adapters or extension cords in series ('daisy chains') and ensuring all wires are well insulated.

\subsubsection{Fire Extinguisher Training}
The fire department taught us which extinguishers need to be used in what circumstances and demonstrated how to use them. Practice runs were also included such that all workers could understand the required hand motion and how the fire extinguisher works.

\section{Testing Procedures}
JPL has strict testing requirements which must be adhered to. Depending on the nature of the testing, the requirements can differ, as detailed below.

\subsection{Operating with a Fixed Tether}
	\begin{enumerate}
		\item The tether must be securely attached to a support surface, such as the ceiling
		\item Everyone in the vicinity of the test must wear glasses
		\item be ready to kill in case the propellers make contact with anything
		\item Stop the flight before the battery reaches 3.7V per cell
		\item In the case of a quadrotor, if its the first flight of the vehicle, hold the vehicle down (with gloves) and test that the response of the motors corresponds to the desired behaviour.
		\item Always have the tether attached to the vehicle and stay at least  the length of the tether away from vehicle
	\end{enumerate}

\subsection{Operating with a Moving Tether}
	\begin{enumerate}
		\item In addition to the above requirements, the tether must be held with two hands and must follow the vehicle
	\end{enumerate}

 \subsection{Operating Untethered (Indoors Only)}
	\begin{enumerate}
		\item In addition to the above requirements, the vehicle must be a testing area separated by a net.
		\item The pilot must stay away from vehicle, behind the net, if possible.
	\end{enumerate}
	\subsection{Autonomous Flight}
	In addition to the requirements above, if the vehicle is flown autonomously, one must also follow the guidelines below.
	\begin{enumerate}
		\item Before commencing any test, ensure all of those involved in the test are familiar with the expected behaviour.
		\item Ensure others in the vicinity of the test are aware of the test and are wearing safety goggles.
		\item Have a safety pilot who is ready to implement the emergency stop at any sign of unexpected behaviour.
		\item The safety pilot must call out each command before sending it (except emergency stop, which needs to be done with no delay). These commands include taking off, switching to the offboard (autonomous) mode, and any other commands relevant to the test.
	\end{enumerate}

\section{Batteries}
Given that all the rollocopters use Lithium ion batteries, procedures for preventing and dealing with battery fires are critical. Hence, the following procedures are in place.

\subsection{Using Batteries}
While using a battery, a battery monitor must be connected and set such that the minimum voltage from each cell is 3.6V, if flying, or 3.8V if running electronics. Note that the discrepancy is due to the slight decrease in voltage when there is a high current drain, such as the propellers. This step is mainly to maintain battery health. If a battery is drained beyond the specified limit, it may cause a fire when next charged.

\subsection{Charging Batteries}
The charging stations are designated areas where the batteries and chargers are kept. This allows for everyone to be aware of where any such issues may arise. When charging a battery, the charger must be in `Balanced Charge' mode, ensuring that all of the cells are equally charged. In addition, the battery must be kept in a LiPo-safe bag. This is to ensure that, if the battery does explode, the fire will be largely contained by the bag. Finally, the person who connects the battery to the charger is responsible for monitoring it and removing it when fully charged. This is unless the person has explicitly transferred the responsibility to another team member in the lab. No batteries can be left charging unattended.

\subsection{Procedures to Mitigate Consequences}
In the case that one anticipates that a battery might explode (if the battery was used beyond the limit, for example), it must be placed in a LiPo-safe bag and inside a bucket of sand until it can be safely disposed of by the fire department. 

There is also a checklist which is used when leaving the lab. This is included in Appendix \ref{apdx:ohs}.

\section{Field Trips}
Each field trip involves a safety team which creates a plan for the tests to be completed during the trip. By combining this plan with information about the location of the field trip, the safety team can evaluate the safety risks and cover the methods for preventing or managing these situations. All those who intend to be involved in the field trip are then required to attend a briefing to cover the safety protocols. After the field day, there is a debrief meeting in which the safety of the event will be discussed. 
	\cleardoublepage
	\chapter{Literature Review}
\attributions{This literature review was completed by the author of this thesis. The work in Section \ref{sec:hybrid_vehicle_lit_rev} is based on work completed by another member of the Guidance and Controls Team.}
\section{Overview}
This chapter provides a review of relevant research and developed tools used in the project. This thesis focuses on development for navigating in challenging environments. This includes working with environments that encourage localisation failures and navigating through areas which require complicated trajectories and endurance.

\section{Hybrid Aerial-Ground Vehicle Configurations}\label{sec:hybrid_vehicle_lit_rev}
Hybrid vehicles reap the benefits of both ground and aerial vehicles such that there can be lower power consumption when the vehicle is in ground mode and higher mobility in aerial mode. There are many different designs of hybrid vehicles, including drones with passive wheels, active wheels, and legs. There are also designs with the drone inside a rotating cage or structures which change configuration to switch between the modes.

\begin{figure}[H]
	\centering
	\includegraphics[width=0.8\textwidth]{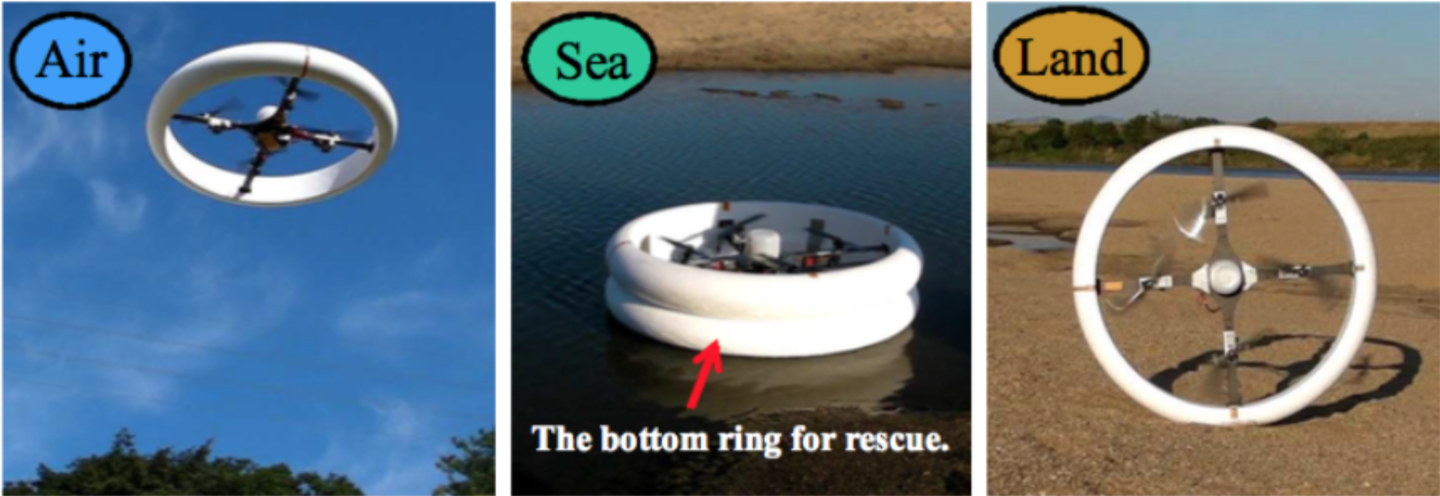}
	\caption{A design with a passive wheel\cite{hybrid_float}}
	\label{fig:passive_wheels}
\end{figure}

An example of a drone with passive wheels is shown in Figure \ref{fig:passive_wheels}.

As shown, this vehicle is capable of traversing land and water, and can fly. While this is a versatile platform, when designing a vehicle for autonomous flight, the sensor configuration must be considered. When the vehicle is in land mode, the front of the wheel would block the view of any sensors in the centre of the wheel. In addition, any side-facing cameras would be constantly spinning and, thus, is susceptible to motion blur.

\begin{figure}[H]
	\centering
	\includegraphics[width=0.5\textwidth]{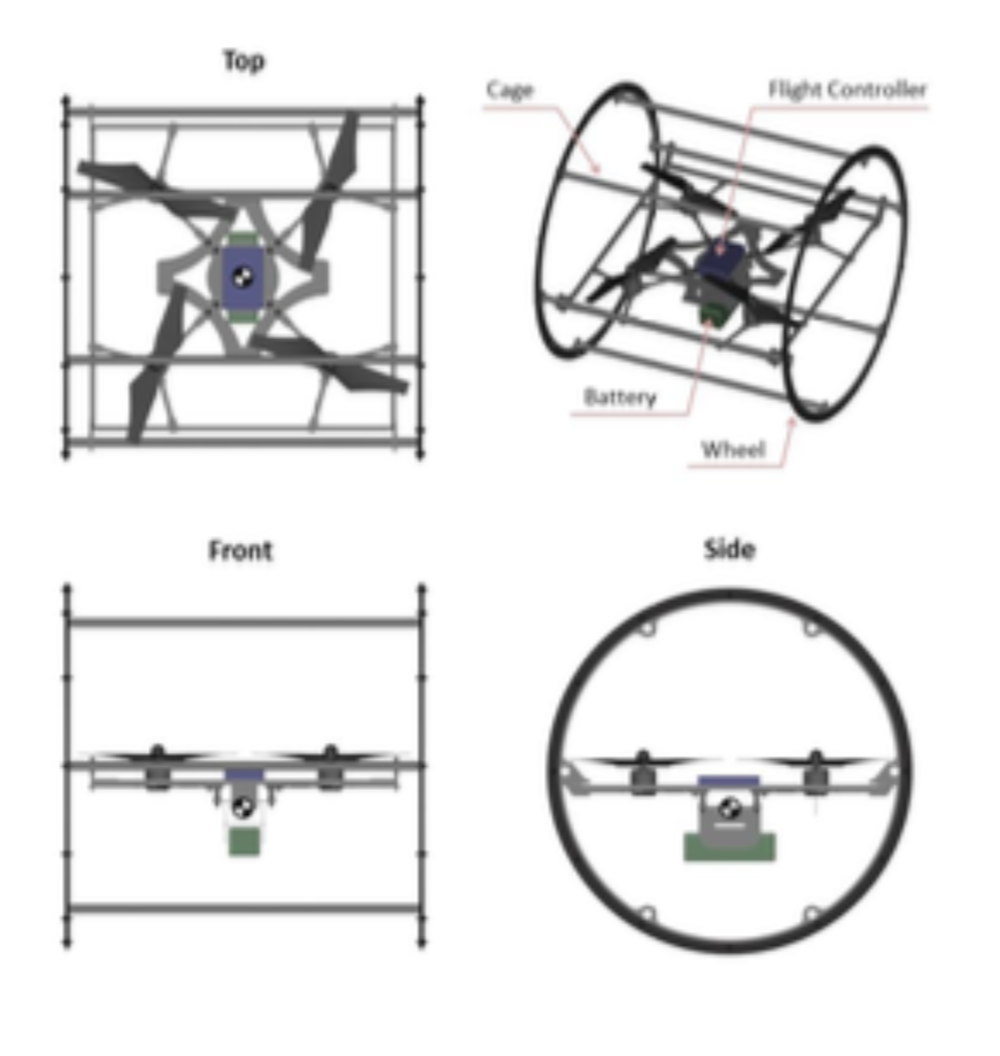}
	\caption{A design with a cage\cite{hytaq}}
	\label{fig:caged}
\end{figure}

An example of the caged drone structure is shown in Figure \ref{fig:caged}.

This vehicle is capable of rolling and flying but also has a similar issue with sensory information. While a camera mounted on the drone would not rotate with the cage, as per the previous design, the beams across the cage would hinder the view of the camera and cause motion blur and difficulty for the visual-inertial odometry algorithms.

\begin{figure}[H]
	\centering
	\includegraphics[width=\textwidth]{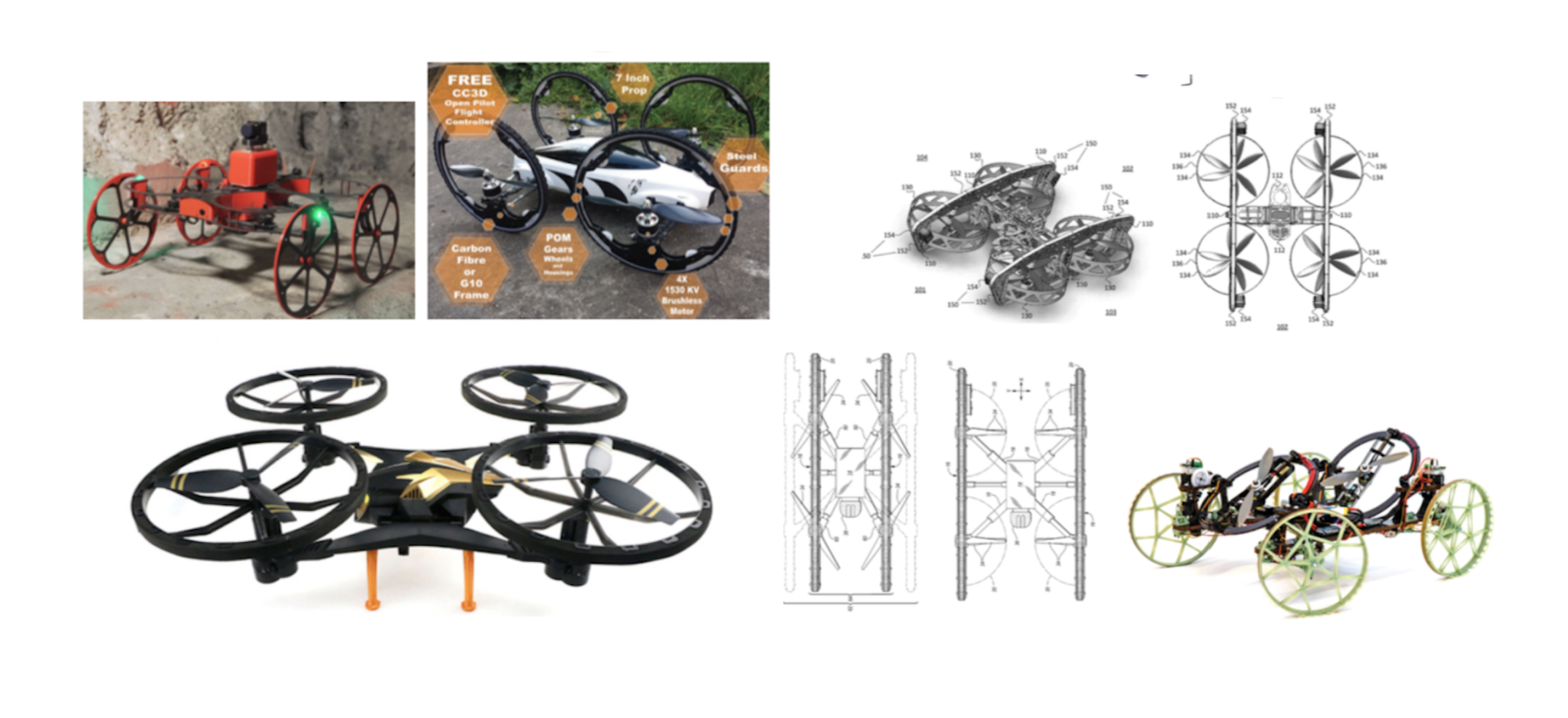}
	\caption{Designs with active wheels\cite{state_2,state_3,state_4,state_5}}
	\label{fig:active_wheels}
\end{figure}

There are many vehicles which appear as drones with active wheels. A common configuration for this is to have the propellers inside the wheels. This allows for the wheels to act as propeller guards. Examples of this are shown in Figure \ref{fig:active_wheels}.

\begin{figure}[H]
	\centering
	\includegraphics[width=0.8\textwidth]{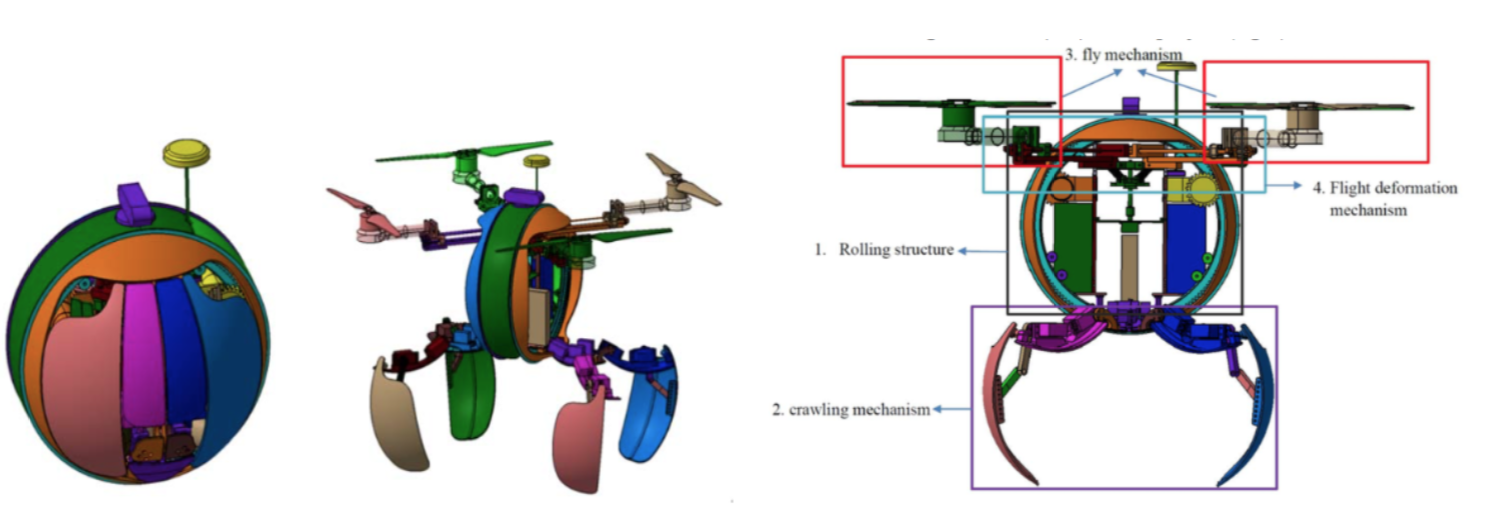}
	\caption{A design with legs\cite{state_10,state_11}}
	\label{fig:legs}
\end{figure}

Finally, a legged drone was designed as shown in Figure \ref{fig:legs}.

This was designed in simulation. The vehicle is capable of rolling, walking, and flying. This design also has the issue with structure occluding the sensors.

\section{Perception and Global Localisation}
The algorithms developed in this thesis rely on sensory information, hence, this section provides some background on the perception subsystem. This subsystem is designed to obtain reliable information about the environment and provide this to the path planning and control subsystems. To ensure reliability, information from multiple different types of sensors is used. For instance, a stereo/infra-red color and depth camera is used in conjunction with a Light Detection and Ranging (LIDAR) system, an inertial measurement unit (IMU), and wheel encoders. The varying types of sensory information are use for different parts of the analysis and some can be fused.

Global localisation refers to identifying the position and orientation of the vehicle with respect to a world reference frame. The linear and angular positions and orientations can be inferred from the fused sensory information, as detailed in this section. While this is simple for the majority of mobile vehicles due to the Global Positioning System (GPS), it is more difficult in underground environments, for which this vehicle is required to operate, due to the lack of access to GPS information. More fundamental methods relying on immediately available sensory information and on board algorithms are required.

\subsection{Sensors}
\subsubsection{RealSense Camera}
An Intel RealSense camera is an RGBD (red, green, blue, depth) camera capable of using stereo and infra-red (IR) sensing to create a color image, depth map and point cloud. The color and depth images can be processed using an algorithm known as ORBSLAM\cite{orbraul2014} which uses simultaneous localisation and mapping (SLAM) to identify and follow the movement of features for localisation. More detail of ORBSLAM is provided in Section \ref{sec:orbslam}.

\begin{figure}[H]
	\centering
	\includegraphics[width=0.5\textwidth]{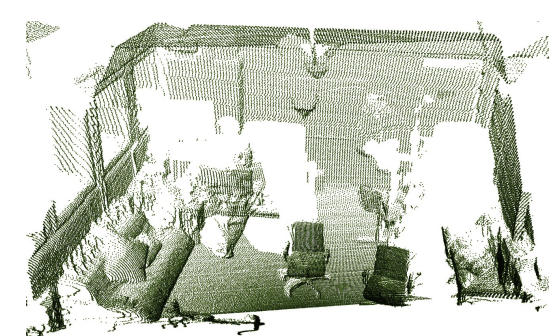}
	\caption{Sample point cloud in the PCL library \cite{pcl}}
	\label{fig:litrevpcl}
\end{figure}

\subsubsection{Light Detection and Ranging (LIDAR)}
The basis of a LIDAR sensor is to send out beams of IR light and process the received signals to obtain depth measurements. There are two main types of LIDAR instruments; 2-dimensional (2D) and 3-dimensional (3D). The 2D instrument, such as the Hokuyo\cite{hokuyo}, sends the IR beams along a single plane. It is commonly attached to a servo motor, allowing it to rotate and combine information from a range of planes. This allows for 3D data to be obtained.

\begin{figure}[H]
	\centering
	\includegraphics[width=0.5\textwidth]{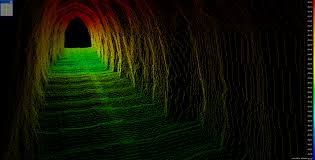}
	\caption{Sample point cloud from a LIDAR sensor\cite{tunnelmap_pic}}
	\label{fig:lit_rev_lidar1}
\end{figure}

The 3D instrument, such as the VLP16\cite{vlp16}, is similar to the 
2D LIDAR but rotates continuously, instead of back and forth like a 
2D LIDAR on a servo motor. It allows for a greater field of view in a 
single rotation without the requirement of more significant temporal 
fusion. The data takes the form in Figure \ref{fig:lit_rev_lidar1}.

\subsubsection{Inertial Measurement Unit (IMU)}
The IMU initially used in this project was included in the pixhawk flight controller board\cite{pixhawkmini}. This contains an accelerometer and a gyroscope. Thus, the orientation and acceleration estimations are reasonably reliable. However, when considering a mission lasting longer than a few minutes, drift becomes an issue. When using the IMU to find velocity, the errors in the acceleration are integrated over time, leading to a significant decrease in the reliability. Hence, the IMU data needs to be fused with other sensory information to obtain an accurate pose estimate.

\subsection{Initial Processing of Sensory Information}\label{sec:orbslam}
The raw images, point clouds and inertial measurements require some processing to make the results more simply applicable to the planning and control systems. Our system runs an ORBSLAM algorithm on the RealSense camera images to identify and track features. It is used for real-time calculation of the trajectory of the camera. This can be used to determine the position, velocity for both translation and rotation in three dimensions.

\subsubsection{How ORBSLAM Works}
The RealSense camera provides an RGBD image to ORBSLAM. The ORBSLAM algorithm then identifies features in the image and calculates the pose of the camera based on a comparison of the current features with those previously recorded. The features and pose are then saved in a structure referred to as a `keyframe'\cite{orbraul2014}. As the camera provides more sensory information, more keyframes are created, the combination of which creates a map. Comparing the current view to the saved keyframes in the vicinity allows for localisation.

\subsection{Fusing Information for Localisation}
To improve the reliability of the localisation, the information from the IMU and ORBSLAM are combined using an algorithm called EKF2\cite{ekf2sameni2007} which is an Extended Kalman Filter (EKF). A Kalman filter (KF) is often used to combine noisy information from sensors with information from a model-based prediction. The model does not account for disturbances, while the measurements will include the results of the disturbances but with noise. Hence, an estimation can be created which aims to filter out the noise without rejecting the disturbances. 

Assuming the model is of the form in Equation \ref{eqn:ekf_model_state}, below, where $x$ is the state vector. The true state can be considered to be in the form of Equation \ref{eqn:ekf_true_state}, where $w$ represents the disturbances in the system. The measurement, $\hat{x}$, can then be represented as in Equation \ref{eqn:ekf_meas_state} where $v$ denotes the noise in the measurements. $B_w$ and $B_v$ are vectors which correspond to how the disturbances impact the different elements of the state. For instance, if the state vector contains two elements, and the disturbances only impact the first element, $B_w$ would be equal to $[1, 0]^T$.

\begin{align}
	\dot{\bar{x}} &= f(x) \label{eqn:ekf_model_state}\\
	\dot{x} &= f(x) + B_w w\label{eqn:ekf_true_state}\\
	\dot{\hat{x}} &= f(\hat{x}) + B_w w + B_v v\label{eqn:ekf_meas_state}
\end{align}

The algorithm for the implementation of the Extended Kalman Filter is included in Appendix \ref{apdx:ekf_derivation}.

\newpage

\section{Collision Avoidance}
Once some form of localisation is achievable, a trajectory can be planned and tracked. A trajectory is a series of points which form an ideal path for the vehicle which must be free of collisions. Collision avoidance encompasses two main processes. The first process is to identify the obstacles and the second is to implement an avoidance mechanism.  

\subsection{Obstacle Detection}
The current system receives point clouds from the sensors.  Hence, obstacle locations must be extracted from the point cloud.  There are a few common methods for achieving this. One must note that when considering if a collision will occur, the most important factor is the minimum distance between a potential path and the nearest obstacle. 

\subsubsection{OctoMap}
OctoMap\cite{octomaphornung2013} is a tool that creates an occupancy grid based on information about the surroundings. Given a point cloud, OctoMap will create a 3D grid of Booleans which specify if the grid point contains an obstacle. While this is computationally efficient\cite{voxbloxoleynikova2017}, this is not ideal for optimal trajectory generation. While it is possible to check if a given point is in collision with known obstacles, it would be more useful to have information about distances to obstacles. This provides the opportunity for further optimization. However, this can be computationally expensive.

\subsubsection{VoxBlox}
\begin{figure}[H]
	\centering
	\includegraphics[width=0.7\textwidth]{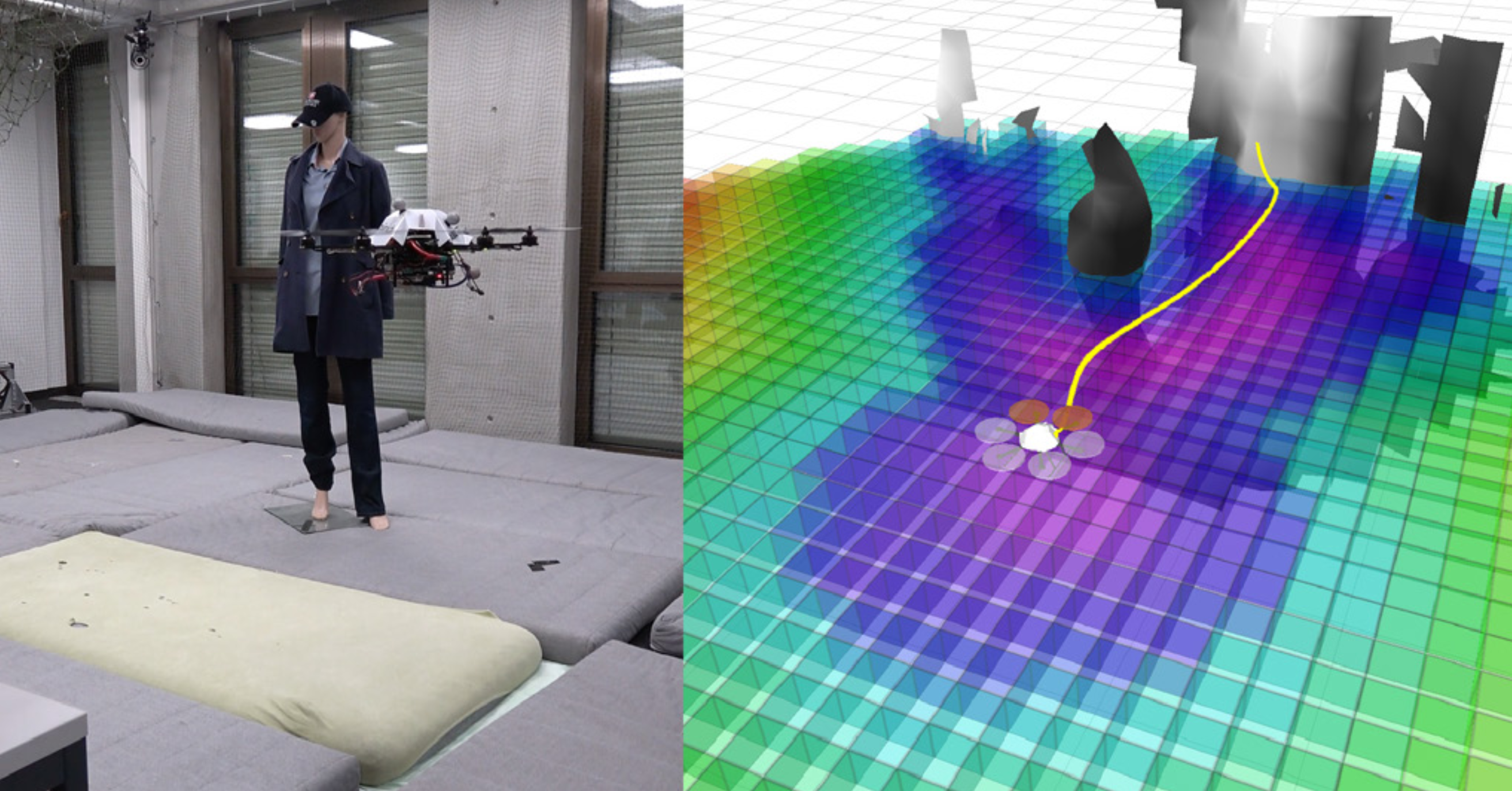}
	\caption{Sample planning using the VoxBlox representation of an 
		ESDF\cite{voxbloxoleynikova2017}}
	\label{fig:voxblox1}
\end{figure}

A tool called VoxBlox\cite{voxbloxoleynikova2017} can be used to determine the location of the nearest obstacles to a given point. VoxBlox creates a Euclidean Signed Distance Field (ESDF). An ESDF is a 3D grid of points (volumetric pixels, or voxels) where each point contains the distance to the nearest obstacle.  This is exemplified in Figure \ref{fig:voxblox1}. 

In this visualisation, the intensity of the colour represents the proximity to an obstacle. The goal was set to be behind the mannequin. As shown, a path can be generated which follows the maximum distance from obstacles while still approaching the goal.

\begin{figure}[H]
	\centering
	\includegraphics[width=0.5\textwidth]{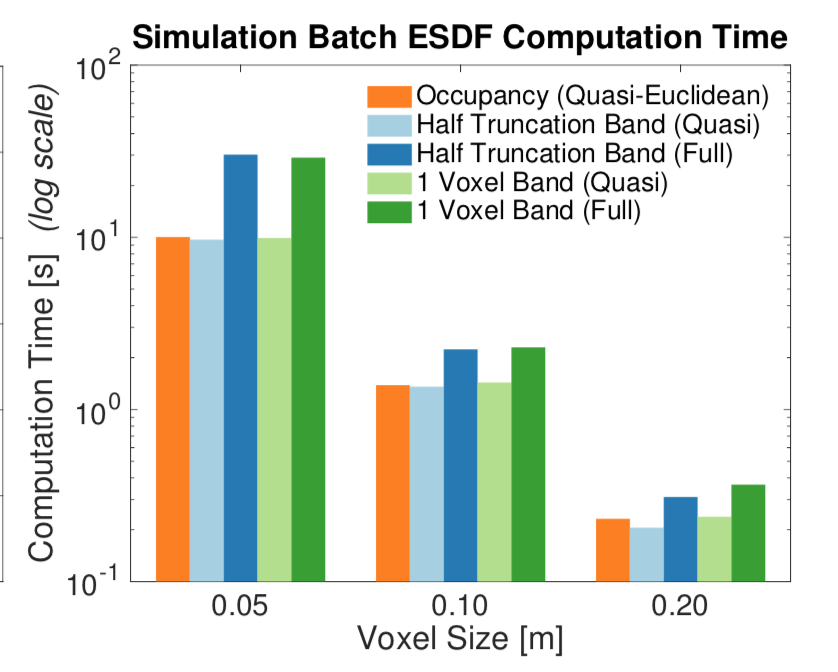}
	\caption{Comparison of computational time for different methods \cite{voxbloxoleynikova2017}}
	\label{fig:voxblox_time}
\end{figure}

As in Figure \ref{fig:voxblox_time}, it has been shown that a Quasi-euclidean VoxBlox grid is less computationally expensive to create. Note that Quasi-euclidean distance refers to the sum of lengths of paths segments between two points, as opposed to the absolute distance (euclidean).

From the graph, a Quasi-Euclidean signed distance field with a resolution of 0.2m took approximately 0.3s to create, while a resolution of 0.05m required 10s. For fine control through confined spaces, it is ideal to be able to have a resolution of approximately 5cm or lower while still maintaining the ability to run in real time. Given this constraint, another method must be considered.

\subsubsection{KD-Trees}\label{sec:lit_rev_kdtree}
\begin{figure}[H]
	\centering
	\includegraphics[width=0.4\textwidth]{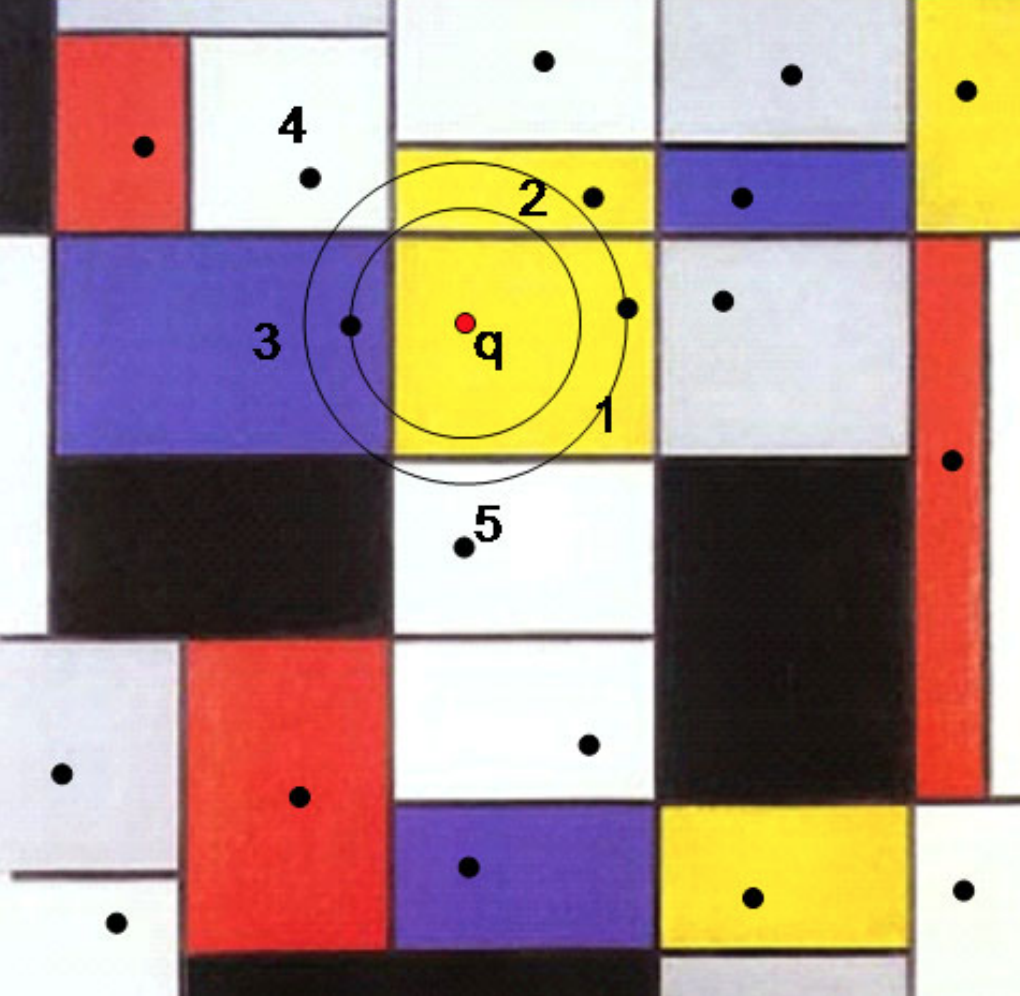}
	\caption{Illustration of the KD-tree structure \cite{pcl_kdtree}}
	\label{fig:lit_ref_pcl_kdtree}
\end{figure}

Another method for finding the nearest obstacles is to perform a ‘$k$-nearest-neighbours’ (KNN) search. The algorithm is exemplified in Figure \ref{fig:lit_ref_pcl_kdtree}. 

First, the points are split into a grid such that each square contains one point, thus forming the $k$-dimensional tree (KD-tree). The structure is then ready to perform searches.

When a query point is provided, the square in which it resides is identified. The distance to the cloud point inside that square is calculated and defines the radius for a circle which is drawn around the query point. This represents the new search space. The search then considers the squares closest to the query point which are overlapped by the circle. In this case of Figure \ref{fig:lit_ref_pcl_kdtree}, these are squares 2, 3, 4, and 5. 

The point inside each of these squares is then identified and the distance to the query point calculated. If this point is outside the circular search space, the point is ignored as per square 2. If the point is within the circular search space, a new search space is defined around the query point with a radius equal to the distance to the new point, such as in square 3. 

The next  and last square which resides inside the new search space is square 4. As the point is outside the radius, it is ignored and the closest point to the query point is the point in square 3. This process is repeated until the $k$ nearest points are identified. 

This algorithm exists within the Point Cloud (PCL) library \cite{pcl_kdtree} and can be used to create and search through a 3D tree based on a point cloud from the sensors.

\subsection{Obstacle Avoidance}
Once obstacles have been identified, action can be taken to avoid collisions. A few methods of varying complexity and reliability are presented in this section.

\subsubsection{Barrier Functions}
Barrier functions\cite{notes5520} use the locations of obstacles to create a spatial cost function which can be evaluated to find the optimal path. The concept is that the cost function is based on the sum of a number of barrier functions which correspond to obstacles. An obstacle location will be converted into a log function such that the cost approaches infinity as the obstacle is approached. This makes the appearance of a ‘barrier’ around the obstacle.

An approach such as Newton’s Method (detailed in Appendix \ref{apdx:newtons_method}), an iterative interior point method, can be used to find the direction of the lowest cost and form a path accordingly. 

\subsubsection{Potential Fields Method}
This method is similar to that of barrier functions in that the obstacle locations are used to devise a spatial cost function and is, in effect, an extension of the method. A path will be defined by following the most negative gradient of the function\cite{cmuPF}. 

\begin{figure}[H]
	\begin{subfigure}[b]{0.4\linewidth}
		\centering
		\includegraphics[width=\linewidth]{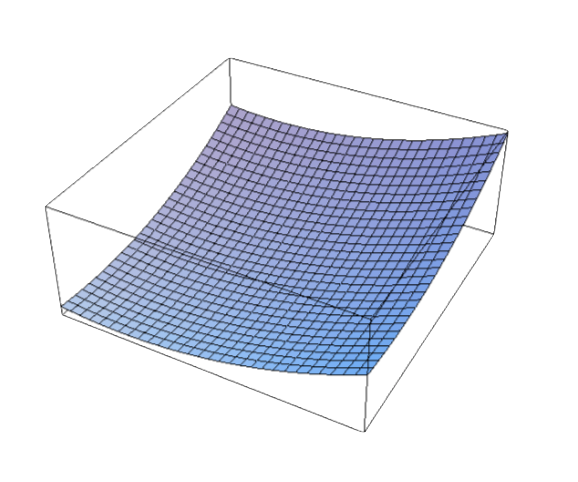}
		\vspace*{-1cm}
		\caption{Attractive function towards the goal}
	\end{subfigure}
	\hfill
	\begin{subfigure}[b]{0.5\linewidth}
		\centering
		\includegraphics[width=\linewidth]{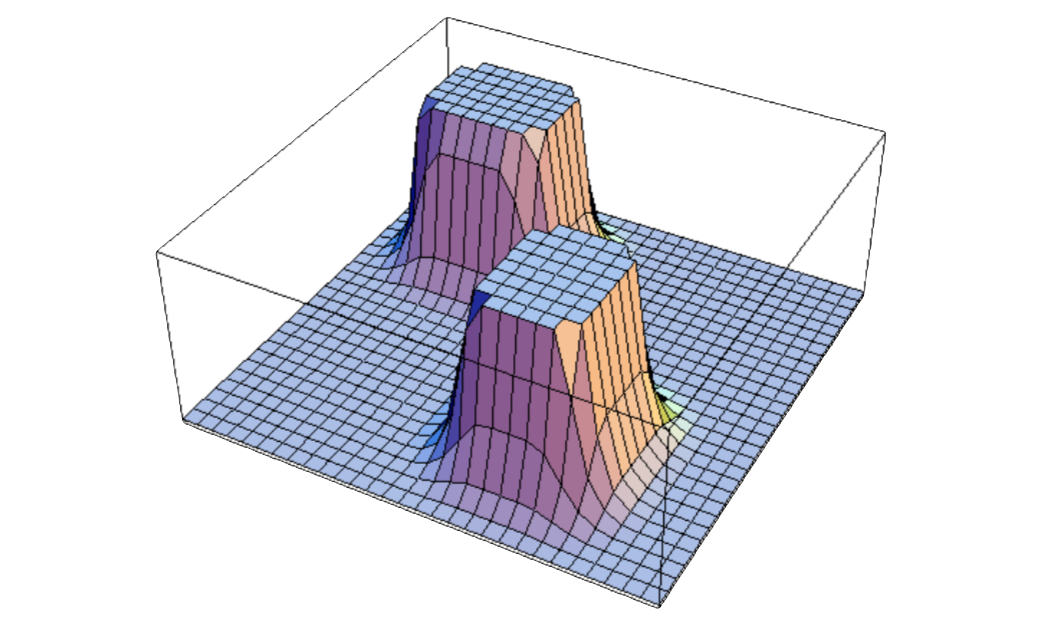}
		\vspace*{-1cm}
		\caption{Repulsive functions away from the obstacles}
	\end{subfigure}
	\caption{Potential field functions \cite{cmuPF}}
	\label{fig:pf_lit}
\end{figure}

\begin{figure}[H]
	\centering
	\includegraphics[width=0.4\textwidth]{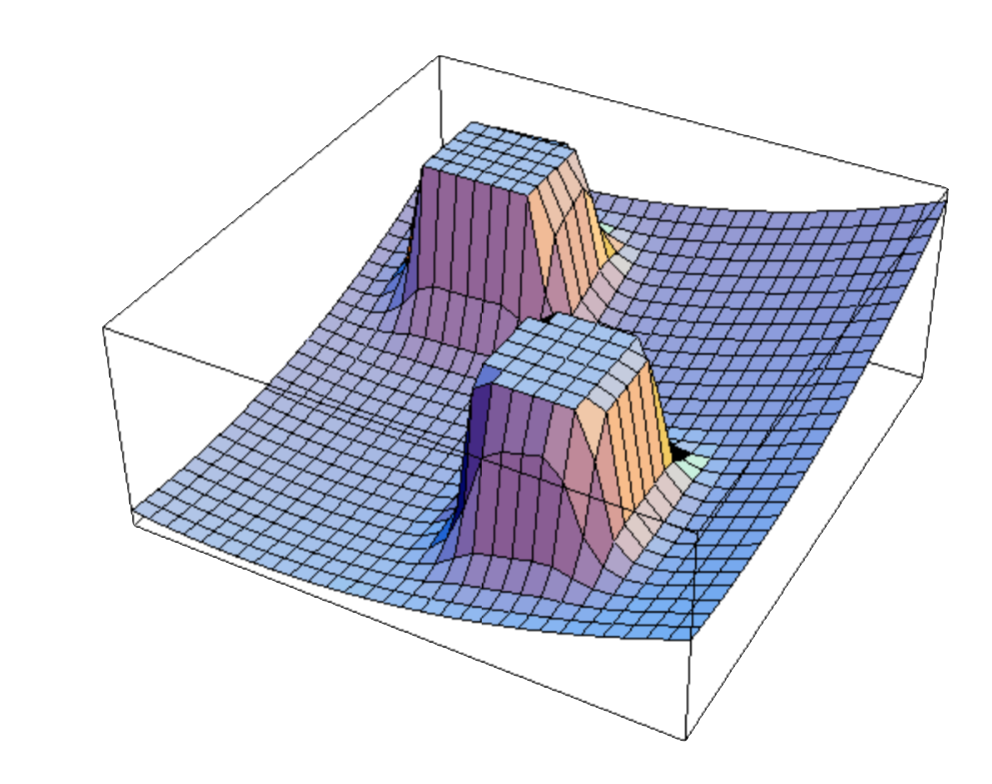}
	\caption{Resulting total potential field function \cite{cmuPF}}
	\label{fig:cmu_pf_tot}
\end{figure}

Attractive and Repulsive potential functions are defined  and can take many forms. For instance, a quadratic function could be determined such that there is a quadratic increase in cost approaching an obstacle or a quadratic decrease approaching the goal. Figures \ref{fig:pf_lit}a and \ref{fig:pf_lit}b demonstrate sample quadratic potential field attractive and repulsive functions. Figure \ref{fig:cmu_pf_tot} shows the combined potential field function which is the total cost function used for finding the optimal paths.

Like barrier functions, Newton’s method is commonly used for evaluating the resulting cost function.

The main advantage to this method is that there is continuity when selecting the path. In addition, optimality can be easily balanced with computational efficiency by tuning the number of iterations of Newton’s method. With more iterations, the variation between the solution will decrease. One can choose a tolerance for this variation which then determines the optimality of the solution.

\begin{figure}[H]
	\centering
	\includegraphics[width=0.4\textwidth]{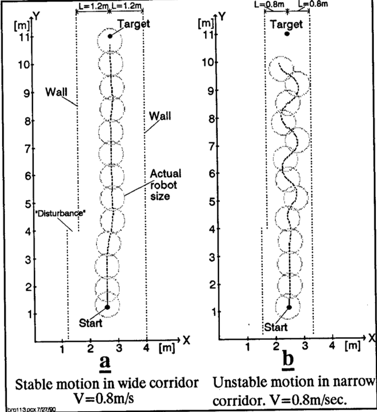}
	\caption{Instability in potential field methods when used in real time\cite{pfbadkoren1991}}
	\label{fig:pfbad}
\end{figure}

While potential field and barrier function methods are ideal theoretically, some challenges have been identified when implemented in real time. For instance, when navigating through narrow passages, if the vehicle turns too far towards one wall, the algorithm will lead to a correction which will cause it to move towards the other wall. This can lead to instability as shown in Figure \ref{fig:pfbad}. 

While this has proven successful on mobile robots moving extremely slowly (0.12m/s)\cite{slow_fieldarkin1989}, Koren and Borenstein\cite{pfbadkoren1991} have shown that in faster-moving systems, the instability exemplified above can occur.

\subsubsection{Polynomial Trajectories}\label{sec:lit_rev_mps}\label{sec:traj_lit_rev}
As the robot is a hybrid aerial-ground vehicle, the planner and controller are designed to be compatible with both the flying and rolling configurations. The initial iteration of the trajectory generator was made for the flying configuration. When flying, the dynamics of the system are significantly less complicated as, like standard drones, the force vector of the drone is in the same direction as the acceleration and there are no constraints on the direction of travel. This is because the vehicle is a free body in space and is powered by only the propellers.

The method chosen for generating trajectories was based on some theory presented in a paper from MIT\cite{quadpolyrichter2016} about how to ensure a dynamically feasible trajectory for a quadcopter.  In this paper, Richter et al. explain how to design a smooth trajectory in a dense indoor environment. The paper discusses how commonly used path planning methods like rapidly- exploring random trees (RRT) and 
RRT* are very effective with path optimisation but not do not consider paths that a vehicle with a high degree of freedom, such as a quadcopter, could follow. 

\begin{figure}[H]
	\centering
	\includegraphics[width=0.8\textwidth]{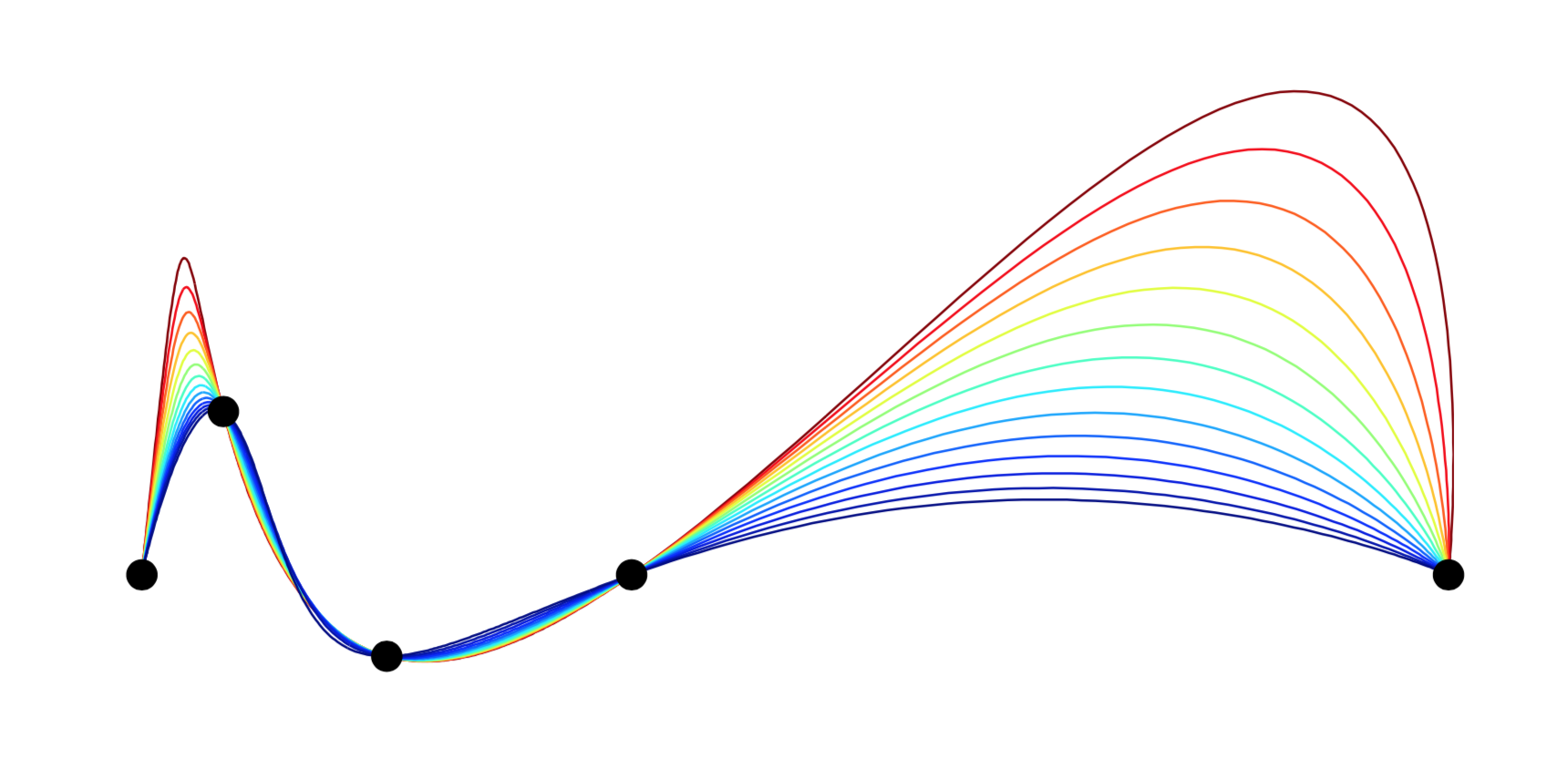}
	\caption{Comparison of trajectories with optimised time compared to 
		snap \cite{quadpolyrichter2016}}
	\label{fig:snaptime_lr}
\end{figure}

Quadcopter dynamics can be shown to be differentially 
flat\cite{minsnapmellinger2011} . This means that a desired path can 
be mapped to the states and control inputs which are necessary to 
follow the path. Given that the orientation of a quadcopter is based 
on the acceleration, smooth changes in acceleration are desired. 
Hence, a low rate of change of jerk (the derivative of acceleration), 
i.e. snap, is ideal. Hence, the algorithm works to minimise snap. To 
do so, 6th order polynomials are generated which take the constraints 
and optimal non-dynamic path (straight line from RRT, for instance) 
and define a feasible dynamic path. The paper also found that there 
was a trade-off between the snap and speed of the trajectory. As 
shown in Figure \ref{fig:snaptime_lr}, the warmer-coloured paths 
represent paths with less snap while the cooler-coloured paths have 
been optimised for time instead. 

\vspace{-0.5cm}


Each of these potential paths is known as a motion primitive. Many 
studies\cite{mp1gray2012,mp2frazzoli2005,mp4hauser2008,mp3yakey2001, 
	mp5bottasso2008} also use motion primitives for trajectory planning. 
Creating a group of primitives then checking for collisions is an 
effective and computationally cheap method of generating semi-optimal 
collision free paths.

The best primitive can be selected according to quality of the path and the proximity to the goal, for instance. The quality of the path is quantized by a cost function which can be based on the proximity to obstacles, for instance.  Determining the distance to the nearest obstacle can be achieved using a search through a KD-tree. A KD-tree can be generated based on a point cloud then points along the primitives can be assigned as query points. The distance is then found as per Section \ref{sec:lit_rev_kdtree} and can be compared to a collision buffer to determine if the path will collide.

\begin{figure}[H]
	\centering
	\includegraphics[width=0.4\textwidth]{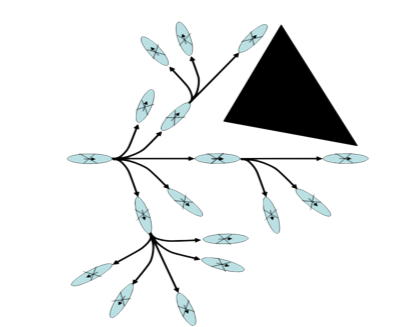}
	\caption{Sample primitives for planning around an obstacle\cite{cmu_prims}}
	\label{fig:lit_rev_cmu_prims}
\end{figure}

An example of this is shown in Figure \ref{fig:lit_rev_cmu_prims}.

In this case, the colliding primitives have been excluded from consideration. The remaining paths can be selected from based on the distance to the goal, for instance.

\newpage


\newpage
\section{Navigation without Localisation}\label{sec:lit_rev_wf}

Previous methods for trajectory generation all assume that localisation information is available and reliable. However, if the robot jolts, the SLAM algorithm can lose track causing the robot to lose the position estimate. Hence, for the robot to be able to continue operating, it needs the ability to navigate without localisation. Few papers exist about the loss of localisation. As stated by A. Howard\cite{navnolochoward1996}, many papers examine how to prevent losing localisation and a few papers investigate regaining localisation but studies investigating navigating without localisation are rare. 

\subsection{Reactive Networks}
\begin{figure}[H]
	\begin{subfigure}[b]{0.4\linewidth}
		\centering
		\includegraphics[width=\linewidth]{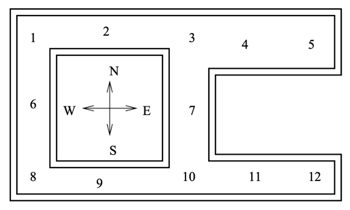}
		\caption{Simple environment}
	\end{subfigure}
	\hfill
	\begin{subfigure}[b]{0.4\linewidth}
		\centering
		\includegraphics[width=\linewidth]{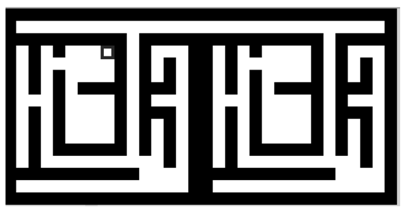}
		\caption{More complicated environment}
	\end{subfigure}
	\caption{Reactive network environments\cite{navnolochoward1996}}
	\label{fig:reactive_networksLR}
\end{figure}

A. Howard\cite{navnolochoward1996} examines a reactive network approach, assuming that the vehicle had some information about orientation provided by a compass, for instance. A series of circumstances were mapped to actions such as ‘travel south until a corner is reached, then move east’. This algorithm was created for the environment shown in Figure \ref{fig:reactive_networksLR}a where the goal was to reach locations 5 or 12. The algorithm was extended to the more complicated scenario shown in Figure \ref{fig:reactive_networksLR}b.  

While this algorithm was successful for the subset of environments tested, the paper admitted that extension to the general case would have some issues.  The issues identified in this paper were:

\begin{itemize}
\item The robot might enter a circumstance not mapped to an action
\item The robot might enter an infinite loop
\item The robot might reach a location that incorrectly resembles the goal
\end{itemize}

The author suggested that the likelihood of the first issue could be reduced by improving the plan when it occurred. The second issue could be avoided by using a more heuristic approach, such as a tree of reactions to avoid repetition. The final case, however, was to be avoided by adjusting the environment which is not always an option in real world applications.

\subsection{Wall Following}

A method which was thought of by our team was to exploit the structure of the environment. Given that the goal of the test is to map the interior of a tunnel, there will always be walls, hence the idea of following walls. A number of papers\cite{wf1roubieu2012,wf2serres2006,wf3santos1995,wf4dwyer2013,wf5van1992,wf6katsev2011} have focussed on wall following and there was little deviation from the method below.

\begin{enumerate}
\item Identify the wall
\item Define a path to follow which is parallel to the wall
\item Follow the path using a controller, typically a proportional controller based on following the path
\end{enumerate}

All these methods were tested in environments with flat walls, like offices or testing environments built specifically for this purpose. As a result, the identification of the wall was simply completed by fitting a plane to the wall. Defining a path parallel to the wall, and following it, was then quite straight forward. 

A limitation to this method is that the walls are required to be planar. Given that this thesis is for a vehicle designed to operate in mines, this method of wall following is not easily applicable. This thesis introduces a method which does rely on a uniformly-shaped wall.

\newpage
\section{Ground Traversability}\label{sec:trav_lit_review}
Analysing the terrain is essential for any ground vehicle designed to operate in off-road environments. This analysis is commonly used for legged robots\cite{terrainbelter2011}. The goal of this analysis is to identify strong footholds. This involves planning a trajectory towards a goal and planning each step. Hence, areas which are traversable must be identified and a path to the goal identified. Once this has been achieved, individual footholds must be identified and used. In the case of this thesis, the robot is wheeled. Hence, the analysis is identical up to the point of defining the path to the goal through traversable areas, but identifying footholds is not required. Hence, this literature survey focused on the methods used for creating a traversability map of the environment.

\begin{figure}[H]
	\centering
	\includegraphics[width=\textwidth]{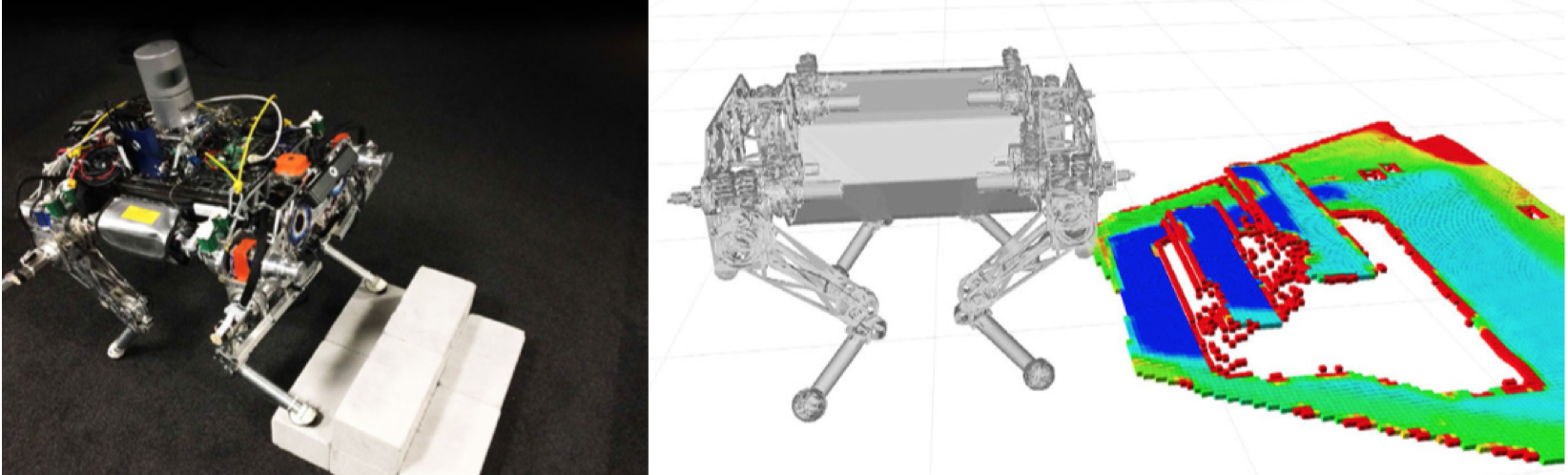}
	\caption{Traversability map\cite{ethz_trav}}
	\label{fig:eth_trav}
\end{figure}

Two common methods were identified. The first method involved creating an elevation map of the environment then analysing this to obtain information about the slope (angle of a fitted plane) and roughness (variance). The intensity of the slope and roughness of the terrain are then combined into a cost function which is used to create a map of traversability. In this map, points with a higher value will be more difficult to traverse\cite{ethz_trav,terrainmeng2018,terrain_trav} An example of this is shown in Figure \ref{fig:eth_trav}.

\begin{figure}[H]
	\centering
	\includegraphics[width=\textwidth]{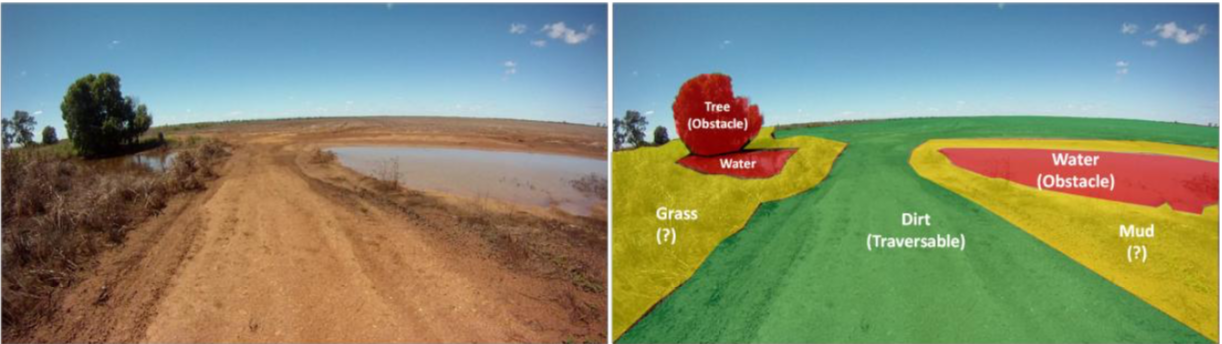}
	\caption{Classified terrain from a machine learning approach \cite{ross_thesis}}
	\label{ross_thesis_trav}
\end{figure}

A second common method was to use machine learning on the images. In the PhD thesis by Patrick Ross\cite{ross_thesis}, different patches were identified using the roughness and colour. This approach was based on machine learning geared towards the outdoor application, to classify the different parts as traversable or not. An example of this is shown in Figure \ref{ross_thesis_trav}.

\section{Summary}
This chapter provided some background into methods which have been developed and tested in previous studies. The conclusions drawn from the comparisons of the different methods helped inform the decisions made during the development completed in this thesis, such as the use of motion primitives for local planning. This thesis also extends on the work that has been covered. For instance, the method for navigating without localisation that is covered in this thesis uses a similar idea to wall following by using the environment but does not have the same restraints on the environment.
	\cleardoublepage
	\chapter{Local Planning for Collision Avoidance}\label{chap:LocalPlanner}
\attributions{The local planning architecture code had been written but not tested or integrated with information from the sensors when the author of this thesis started work on this task. The dust filter described in Section \ref{sec:dust_filter} was developed by another member of the Guidance and Control team. All other work in this chapter was completed by the author of this thesis.}

\section{Overview}
The lowest level of autonomy requires at least a local planner for collision avoidance. The local planner was created to plan short term motion and send commands to the controller. The setpoints provided by the global planner are at a low resolution (such as every 10m) or can be less specific commands, such as `fly forward'. The role of the local planner is to optimise the imminent motion of the vehicle. This involves avoiding obstacles while still moving towards the goal. This is achieved by generating dynamically feasible motion primitives and choosing the optimal primitive based on a cost function. This function designed to balance the distance to the nearest obstacle with the direction of the primitive with respect to the goal. The primitive which is most towards the goal and furthest away from obstacles will be selected. This chapter first covers the structure of the local planner then details the collision avoidance algorithm.

\section{Local Planner Architecture}\label{sec:local_planner}
\begin{figure}[H]
	\centering
	\includegraphics[width=\textwidth]{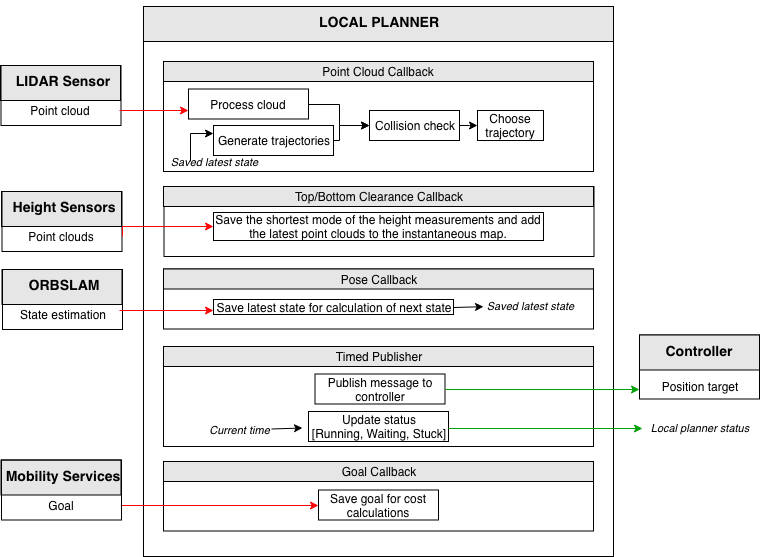}
	\caption{Architecture of the local planner node}
	\label{fig:local_planner_architecture}
\end{figure}

This section details the structure of the local planner node. Figure \ref{fig:local_planner_architecture} shows an overview of the architecture of the node and the required inputs and outputs.

Each section of the local planner is now explained in detail.

\subsection{Point Cloud Callback}
When a raw point cloud is received, it is processed such that the cloud can be used for trajectory selection. In this function, a set of motion primitives are created and the optimal primitive is selected, as detailed in this section.

\subsubsection{Processing the Point Cloud}
\subsubsubsection{Downsampling and Filtering}
When a cloud is received, the first step is to process the cloud and declare if the data is usable. If the cloud is empty or extremely noisy, for instance, the planner should discard the cloud. A Voxel Grid filter is used to downsample and remove noise from the point cloud. The filter downsamples the cloud into a grid of voxels (volumetric pixels) of a chosen size. The filtering is achieved by only creating the voxel in that position if there are above a certain number of points in the voxel. The ratio between the size of the voxel and the number of points needs to be tuned as it acts as a filter. As a result, if there are not enough points inside the area of a given voxel, the points will be rejected as noise.
	
	 \subsubsubsection{Transforming the Cloud}
	The transformation of the point cloud is from the frame of the sensor to the world frame. An affine transform is used with the \textit{pcl::TransformPointCloud} function. This is a matrix of the form in Equation \ref{eqn:affine_transform}.
	
	\begin{align}
	\begin{bmatrix}
	&Rotation&&P_x\\
	&Matrix&&P_y\\
	&&&P_z\\
	0&0&0&1
	\end{bmatrix} \label{eqn:affine_transform}
	\end{align}
	
	In Equation \ref{eqn:affine_transform}, $P_x$, $P_y$, and $P_z$ refer to the required translation and the rotation matrix is a conventional 3x3 rotation matrix based on roll, pitch, and yaw, as illustrated in Equaiton \ref{eqn:rotation_matrix}, where $RM$ is the rotation matrix. and $\phi$, $\theta$, and $\psi$ refer to the roll, pitch and yaw respectively.
	
	\begin{align}
	RM = 
	\begin{bmatrix}
	1&0&0\\
	0&cos(\phi)&-sin(\phi)\\
	0&sin(\phi)&cos(\phi)
	\end{bmatrix} \times 
	\begin{bmatrix}
	cos(\theta)& 0&-sin(\theta)\\
	0&1&0\\
	sin(\theta)&0 &cos(\theta)
	\end{bmatrix} \times 
	\begin{bmatrix}
	cos(\psi)&-sin(\psi)&0\\
	sin(\psi)&cos(\psi)&0\\
	0&0&1
	\end{bmatrix}\label{eqn:rotation_matrix}
	\end{align}
	
	The cloud is first transformed from the sensor frame to the body frame. This involves translating the cloud to the position of the IMU (the origin) on the rollocopter and rotating the cloud to account for any difference between the orientations of the sensor and body frames. In the case of the Velodyne LIDAR, there is no difference in orientation. Hence, the rotation matrix is identity.
	
	The cloud is then transformed from the body frame to the odometry frame. This involves translating the cloud by the latest known position coordinate and rotating the cloud by the current roll, pitch, and yaw.
		
	 \subsubsubsection{Removing the Wheels from the Point Cloud}
	Between the two transforms, when the cloud is in the body frame, the wheels are removed using a \textit{PCL Crop Box Filter}. Given that the position of the wheels is known with respect to the vehicle, points within a box around the wheels are removed from the cloud. This avoids the vehicle perceiving the wheels as obstacles.
	
	 \subsubsubsection{Checking the Status of the Point Cloud}
	The size of the cloud is checked at the end. If the initial cloud was empty, it is decided that the cloud should be empty and that there was simply no obstacles in sight. If the cloud was not empty before processing but is empty after being processed, then the cloud is deemed noisy and therefore unsuitable for use. If the processed cloud is not empty, the cloud is used for collision checking.

\subsubsection{Trajectory Generation}
The trajectory generator receives a start point and goal location. The following procedure is used to generate the trajectories.

\begin{enumerate}
	\item Define a series of end points. These are chosen based on linearly spaced points along a circle defined by the specified planning horizon about the position of the vehicle, as indicated in Figure \ref{fig:prim_endpts}.
	
	\begin{figure}[H]
		\centering
		\includegraphics[width=0.4\textwidth]{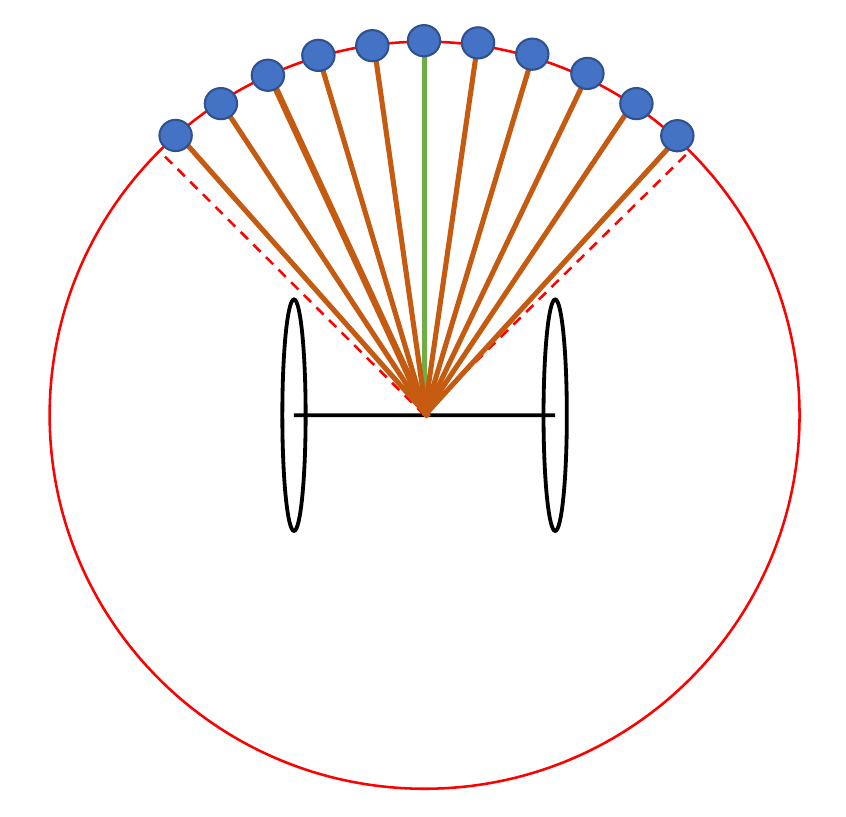}
		\caption{Primitive and endpoint selection}
		\label{fig:prim_endpts}
	\end{figure}

	\item Calculate the coefficients for a constant-snap dynamically feasible trajectory using the minimum-snap model in Equation \ref{eqn:coeff_calcs}. Note that vector $C$ contains the coefficients. This is based on the conditions of the initial and final position, velocity, and acceleration in the given axis. Three polynomials are created per primitive corresponding to the three axes, x, y, and z.
	\begin{align}
	C_{1-3} = [x, \dot{x}, \ddot{x}&]^T\\
	C_{4-6} = A^{-1}B&\\ \label{eqn:coeff_calcs}
	A = 
	\begin{bmatrix}
	T^3&T^4&T^5\\
	3T^2&4T^3&5T^4\\
	6T&12T^2&20T^3
	\end{bmatrix}, 
	&B = 
	\begin{bmatrix}
	\Delta x\\
	\Delta \dot{x}\\
	\Delta \ddot{x}\\
	\end{bmatrix}
	\end{align}
\end{enumerate}

Note that the trajectory coefficients could then be used to calculate the position, velocity, and acceleration in the given axis (x, y, or z) using Equations \ref{eqn:pos} - \ref{eqn:acc}. Note that $t$ is the time along the primitive at which the state is to be calculated.

\vspace{-0.5cm}

\begin{align}
p &= C_1 + C_2t + C_3t^2 + C_4t^3 + C_5t^4 + C_6t^5\label{eqn:pos}\\
v &= C_2 + 2C_3t + 3C_4t^2 + 4C_5t^3 + 5C_6t^4\\
a &= 2C_3 + 6C_4t + 12C_5t^2 + 20C_6t^3\label{eqn:acc}
\end{align}

\subsubsubsection{Collision Checking}
The primitives defined in the previous step are discretised into sample points and a KD-tree is created based on the processed point cloud. The sample points on the primitives are then used as query points for a KD-tree nearest neighbour search. The points are then checked against a specified collision buffer. If there are any points within the collision buffer, the primitive is considered to be in collision.

\subsection{Top/Bottom Clearance Callback}
This function saves the clearance as a global variable. The clearance is then used to plan the motion when in the flying mode. More detail is provided in Chapter \ref{sec:wall_following}.

\subsection{Pose Callback}
This function is called when an odometry message is received. This is the latest estimate of the state and is used by the local planner for the transformation of the sensory information.

\subsection{Timed Publisher}
This function publishes the Position Target message to the controller, the status of the local planner, and a message containing other information for debugging purposes.
The \textit{Position Target} message contains the following information.
\begin{itemize}
	\item The mobility mode of the vehicle (aerial or ground)
	\item The control mode to be used (position control or velocity control) for both motion in the x-y plane and motion along the z axis
	\item The desired state to be reached which could be based on a position and orientation or a velocity and angular velocity.
\end{itemize}

The \textit{Local Planner Status} message details whether the planner is either waiting, running, or stuck. This is used to determine how the position target mode message should be used and can also be used to alert the user of this so that a new goal can be given to the mobility services node.

\subsection{Goal Callback}
This function is called when the local planner receives a goal from the mobility services node. This is used when calculating the cost functions for the selection of the optimal primitive.

\section{Collision Avoidance}
In this section, an approach for avoiding collisions in real time is presented. A requirement for this method is that it cause minimal latency, given that the planner cannot provide paths to the controller without this step being completed.

\subsection{Preliminary Development}
\begin{figure}[H]
	\centering
	\includegraphics[width=0.5\textwidth]{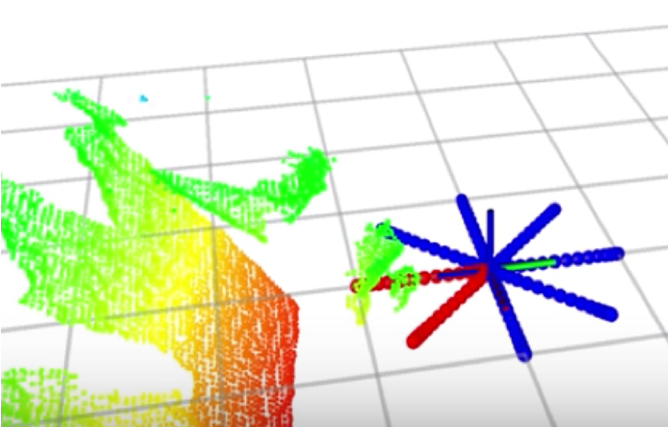}
	\caption{Initial collision avoidance solution}
	\label{fig:local_planner_collision_avoidance}
\end{figure}

The first step was to integrate information from real sensors into this package. An Intel RealSense camera was connected to the local planner code base on the computer and integrated with the collision checker. In addition, a method for visualising the results proved to be a necessity for debugging and was therefore developed in RViz. The visualisation shown in Figure \ref{fig:local_planner_collision_avoidance} was created.

The coloured points represent the point cloud coming into the local planner. The lines forming a star at the origin are the motion primitives being considered. If the primitive is red, it indicates that there is a collision along that path. In this case, it was set to be considered as a collision if there was an obstacle within 30cm of the path.

There were many steps involved in integrating real data and visualising the results. First, the cloud needed to be processed. The following setup was made. First, the cloud is down-sampled from around 100,000 points to 2000 points, increasing the speed of the calculations. This is completed by creating a Voxel (volume-pixel) grid. If a grid point of a specified size has at least a specified minimum number of points, a voxel is placed there. By tuning the size of the grid and the minimum number of points per voxel, one can ensure that outliers are not included and adjust the grid to a more usable number of points with the same filter. This filter also removes all the points at the origin (the location of the camera) and invalid points represented by NaN (not a number).

The next stage of the cloud processing involves transforming the cloud to match the world. The points are extracted in the coordinate system of the camera (herein referred to as the camera frame) and need to be converted to the world coordinate system (the world frame). The world frame is determined using a simultaneous localisation and mapping (SLAM) algorithm. Our system uses ORBSLAM. This algorithm provides the orientation in the same coordinate system based on the initial orientation of the camera, but with a different axis layout. This orientation provides the information required to create an Affine transformation matrix to transform the point cloud from the camera frame to the ORBSLAM frame which, in the early stages of development, was considered to be the same as the world frame. Once this was implemented, rotating the camera led to the point cloud moving around the origin while the features would remain stationary.

Once the cloud had been processed, a KD-tree was generated for it such that the primitives could use a knn-search (k nearest neighbours search) to find the obstacles closest to the sample points along the primitives. The collision status of the primitive (whether or not it would collide with any obstacles) was then correlated to a cost function. This allowed the points representing the primitives to be sent as type PointXYZI, meaning a location coordinate and an intensity, depicting the cost. A point with a higher cost (1, colliding) would be shown as red in RViz while a point with a lower cost (0, not colliding) would be shown in blue.

Once collision avoidance was fully functional and debugged on the computer with the camera connected, the code was transferred to the rollocopter. The rollocopter computer board is an Intel NUC which is effectively a very powerful computer which runs Ubuntu 16.04. Hence, the main task in this was to integrate the sensor with the on board inertial measurement unit (IMU). This is in a gravity-aligned frame. An extended Kalman filter (EKF) had been created which can fuse the IMU data with the ORBSLAM data. The resulting information was then used to transform the point cloud from the camera frame to the ORBSLAM frame, then to the IMU frame, and finally to the world frame.

Once the local planner node was capable of generating commands based on the sensory information from the sensors on the vehicle, it was integrated with the controller. At the point of testing this, the controller was based on a pure pursuit model. While this was not using the trajectories effectively, it had been empirically shown to work effectively in most cases.

\subsection{Hardware Test Results and Improvements for Robustness}
This method was tested in a variety of cases in rolling mode. The following issues were encountered and the corresponding solutions were implemented.

\subsubsection{Field of View Limitations}
\begin{figure}[H]
	\centering
	\includegraphics[width=0.5\textwidth]{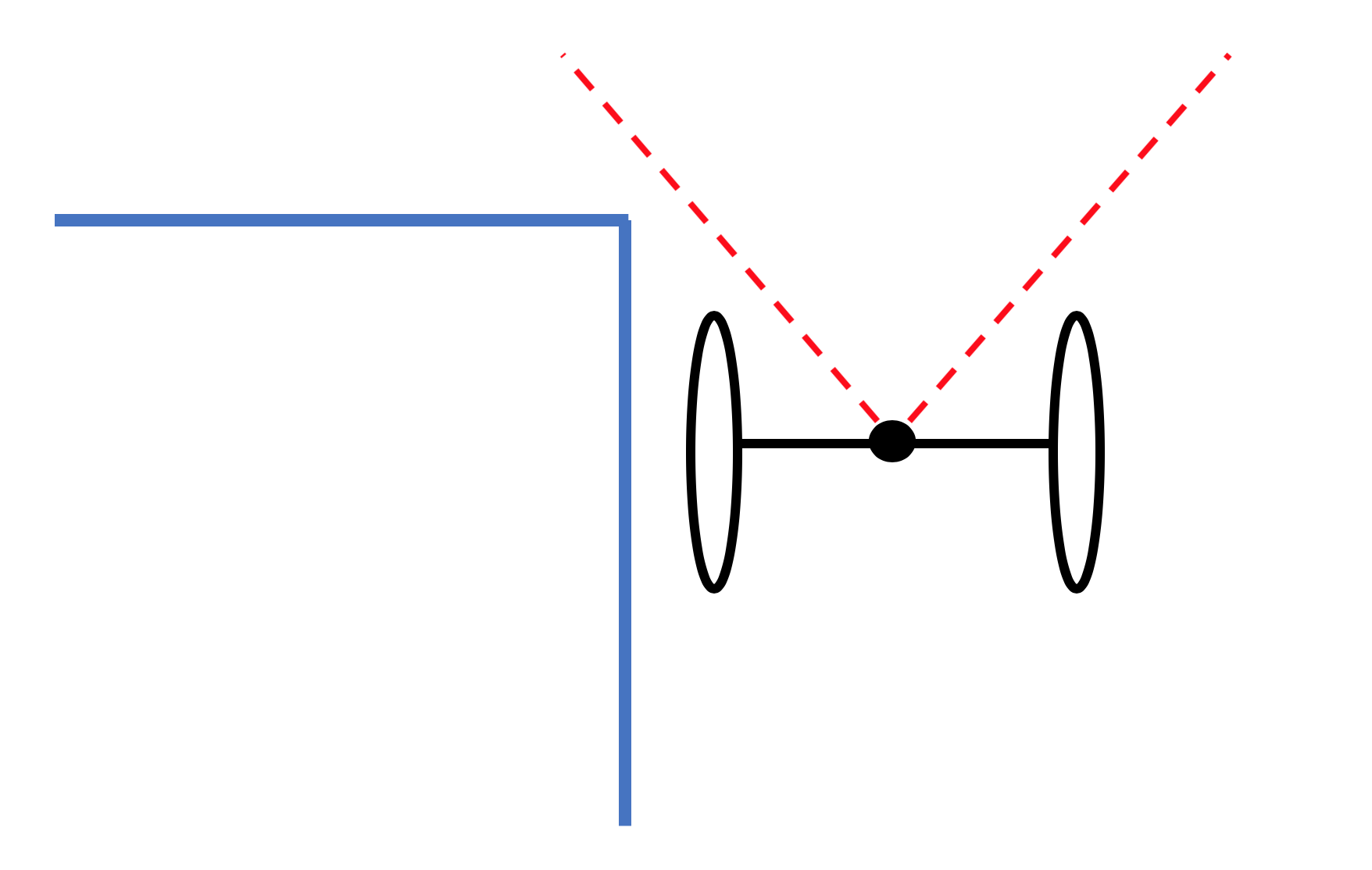}
	\caption{Impact of the limited field of view of the camera, blue = wall, red = field of view}
	\label{fig:fov}
\end{figure}

As shown in Figure \ref{fig:local_planner_collision_avoidance}, there were primitives in all directions despite only having information within the 90$^o$ field of view of the camera. Hence, the primitives are now only generated within the field of view of the camera. In addition, when tested on the vehicle, it was found that there was some jittering when a primitive on the edge of the field of view was selected. This occurs when there was an obstacle near the edge of that primitive. When the vehicle chose the primitive, it would yaw towards it and see the obstacle, hence choosing another path. When it would yaw to the new path, it would then no longer see the previous obstacle and hence choose the primitive on the edge again. This oscillation is what caused the jittering. A case in which this would occur is shown in Figure \ref{fig:fov}.

The solution for this issue was to limit the primitives to be within the field of view of the camera. The given primitive considered to be free could then theoretically be reliably followed without risk of collision.

\subsubsection{Warping Ceiling from Previous Transformation}
Another issue was discovered where the ceiling would appear to warp when the vehicle rolled or pitched. This was fixed by saving the state at the time the cloud was received, instead of using the latest state when transforming. Accounting for this delay allowed for the transformations to be closer to the ground truth and therefore more usable.

\subsubsection{Ground Interference}
When the vehicle pitched forward, the ground was included in the field of view. Given that the collision buffer zone extended 2cm beyond the furthest edge of the wheels, the ground would appear to be in collision with the primitives. The solution to this was to truncate the cloud to occlude the ground region when the transform was accurate.

\subsubsection{Additional Option for When All Primitives are Stuck}
\begin{figure}[H]
	\centering
	\includegraphics[width=0.5\textwidth]{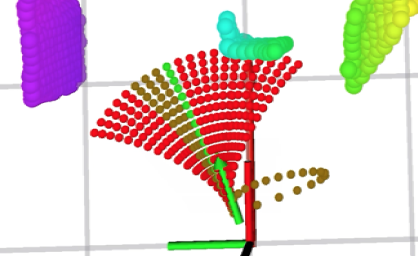}
	\caption{Additional primitive to change the field of view}
	\label{fig:yaw_primitive}
\end{figure}

An additional primitive was added outside the field of view such that when the main primitives were all in collision, the vehicle would have a tendency to yaw. This led to a different field of view, and therefore a higher chance of noticing possible primitives. This primitive is shown in Figure \ref{fig:yaw_primitive}.

\subsubsection{Dusty Environments}\label{sec:dust_filter}
When working in the tunnels and mines, it was found that dust created a significant challenge. The cameras cannot see through the dust and the dust appears in the point cloud and takes a similar form to the wall. A dust filter was required to remove the points that correspond to dust and leave the points which correspond to obstacles. Plotting the raw point cloud in RViz allowed for a comparison of the two clouds.

The Velodyne point cloud is created by 16 lasers rotating around the z axis. Comparing the variance around the rings, it was observed that the variance of the distance to the points along the walls was significantly less than the corresponding variance in dusty areas. Hence, a filter based on the variance of the clouds was used to remove the points which correspond to dust.

\begin{figure}[H]
	\centering
	\includegraphics[width=0.7\textwidth]{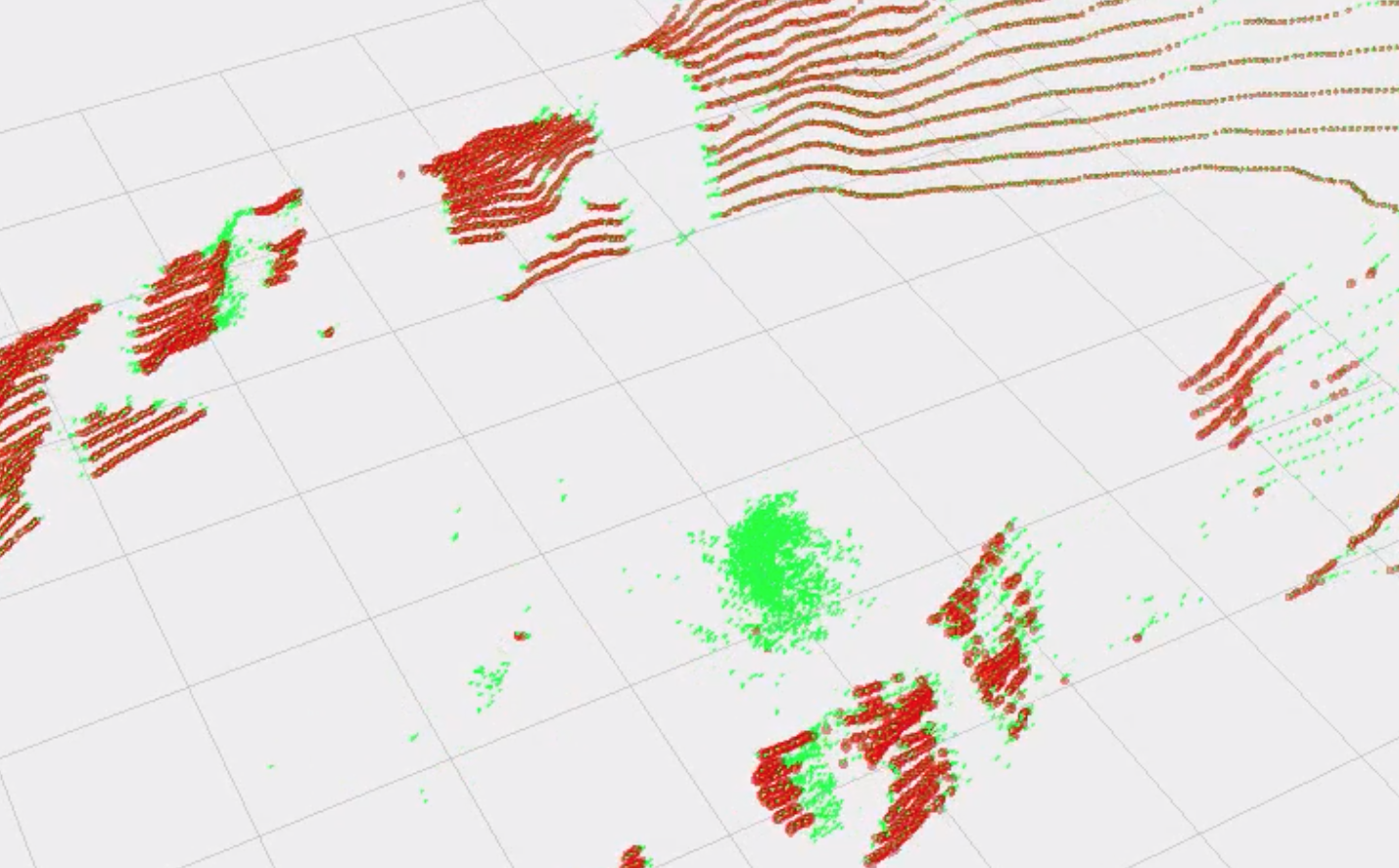}
	\caption{Dust filter in action: green is unfiltered, red is filtered}
	\label{fig:dust_cloud_filtered}
\end{figure}

In addition, the VLP-16 LIDAR has a dual return mode\cite{vlp16_datasheet}. This involves the strongest return point and the last return point being saved. Hence, the high intensity returns from the dust and the far away points from the walls behind the dust were both included in the cloud. Thus, the planner can see the walls and use the dust. Figure \ref{fig:dust_cloud_filtered} shows an example of the dust filter working. The green cloud in the centre is dust. In this case, the holes in the walls are due to the rollocopter wheels blocking the lasers.

\begin{figure}[H]
	\begin{subfigure}[b]{0.5\textwidth}
		\includegraphics[width=\linewidth]{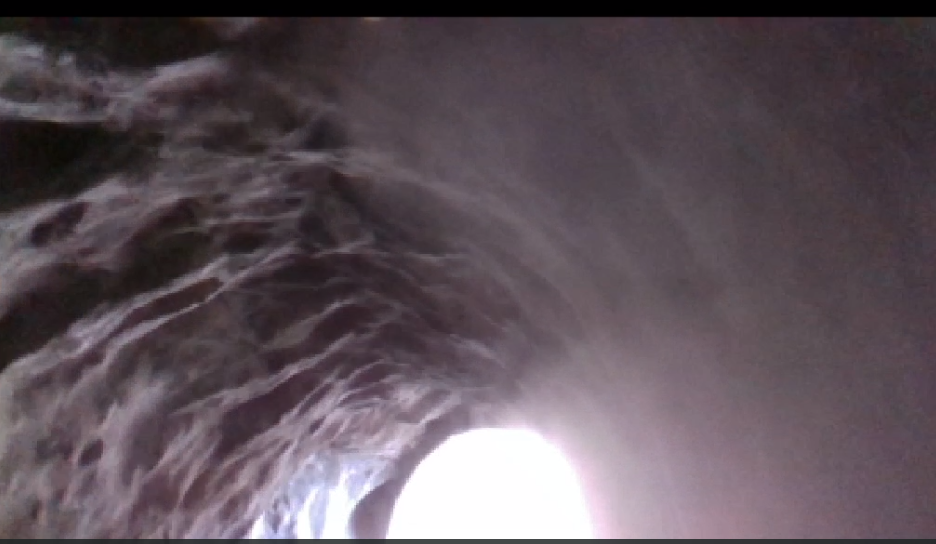}
		\caption{First-person camera view}
	\end{subfigure}
	\hfill
	\begin{subfigure}[b]{0.5\textwidth}
		\includegraphics[width=\linewidth]{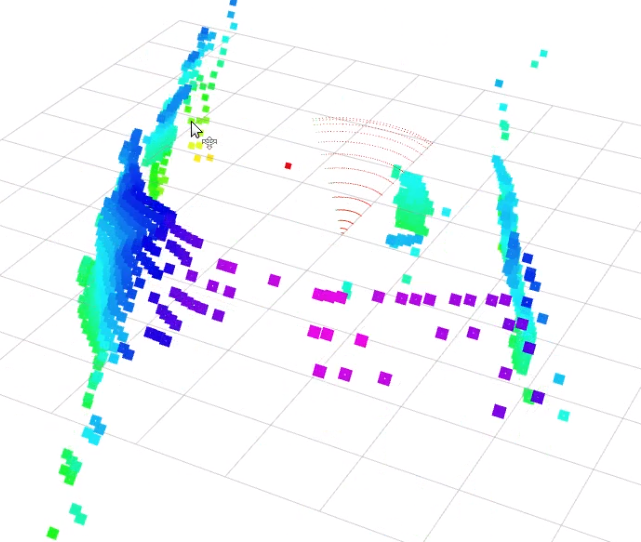}
		\caption{Point cloud}
	\end{subfigure}
	\caption{Dust presented as a wall in the point cloud}
	\label{fig:mueller_dust}
\end{figure}

Figure \ref{fig:mueller_dust} illustrates a test at the Mueller Tunnel which exemplified the requirement for the dust filter. The instantaneous point cloud included the dust. Hence, the dust was treated as an obstacle. This led to the vehicle attempting to avoid the dust which was always a constant distance ahead of the vehicle. Thus, only the side primitives were available to the planner, despite there being no actual obstacles ahead.

\section{Final Results}
With the above improvements implemented, a robust collision avoidance algorithm was created which is capable of avoiding obstacles using instantaneous point clouds. The video found here (\url{https://youtu.be/a9yD9Cap6nA}) shows the collision avoidance algorithm encouraging the vehicle around an obstacle and towards the goal (the red point).
\newpage
\section{Summary}
This chapter covered the development of a local planner capable of collision avoidance for a unique vehicle in extreme environments. The local planning node provides the basic functionality required for the vehicle to navigate autonomously by avoiding obstacles and planning paths towards a specified goal. The main developments through this section of the thesis was to integrate the planning architecture with the vehicle and make the algorithms more robust through testing on hardware and improving the system based on the findings. An additional component which requires addressing is the ability of the local planner to perform without localisation. While the planner is set up to work with instantaneous point clouds, there is still the notion of a goal which needs to be reached. Without localisation, the vehicle can never recognise the arrival at the goal. This issue is addressed in Chapter \ref{sec:wall_following}.
	\cleardoublepage
	\chapter{Navigation without Localisation}\label{sec:wall_following}
\attributions{The hardware testing in this chapter was completed in collaboration with other members of the Guidance and Control team. The local mapper and mid-level planner were developed by another member of the Guidance and Control team. All other work was completed by the author of this thesis.}
\section{Overview}
A major component of this thesis was designing the system to function with unreliable localization. Given the requirement to work in underground environments, access to GPS is unreliable. This leads to several challenges in development. The Perception team on the project have been exploring different avenues such as visual, or LIDAR odometry with and without fusion with an IMU. The current most reliable solution is visual-inertial odometry (VIO), working with ORBSLAM. However, given that the environment is dark and can contain fog, smoke, or dust, VIO cannot be relied upon. Thus, the local planner must be capable of moving the vehicle with minimal access to localisation. As the first year of this challenge involves navigating through tunnels, it was decided that the structure of the environment could be used to control the vehicle. Hence, the local planner in Chapter \ref{chap:LocalPlanner} was extended to include a wall-following behaviour capable of performing without localisation.

\section{Wall Following in Ground Mode}
\subsection{The Initial Method}
At first, this was set up as a simple algorithm in MATLAB which takes a planar wall, the normal of which is known, and converts the wall to a straight line in the body frame in 2D space. A second line, parallel to the wall line, is then defined based on a buffer distance of 30cm, for instance, which becomes the path for the rollocopter to follow.

This algorithm was then converted to C++ and implemented with a RealSense camera connected to a computer. An extra step for this is to extract the wall plane from the real data. The point cloud library (PCL) has a set of functions which use a random sample consensus (RANSAC) method for fitting shapes to the cloud. The plane fitting function was used on the point cloud to define the wall. A visualisation similar to the previous was then developed to show the velocity commands being sent to the controller with respect to the wall.

It was found that this works well for straight walls when the majority of the field of view of the camera is facing the wall. However, if the camera direction was mostly parallel to the wall (defined as a vector pointing straight out of the camera), then the RANSAC algorithm started to fit planes to the floor and the ceiling. To ensure this would not occur, a pass through filter was implemented such that the RANSAC algorithm would be used on a cloud that contained only points within the range in which wall is expected to appear. This involved limiting the z coordinate to above the ground and below the ceiling and keeping only the points with the same y coordinate sign to ensure that the wall on the other side was not being considered too. Once this had been implemented, the path line, and hence velocity commands, stopped changing so rapidly allowing the trajectory to be significantly smoother.

\begin{figure}[H]
	\centering
	\includegraphics[width=0.5\textwidth]{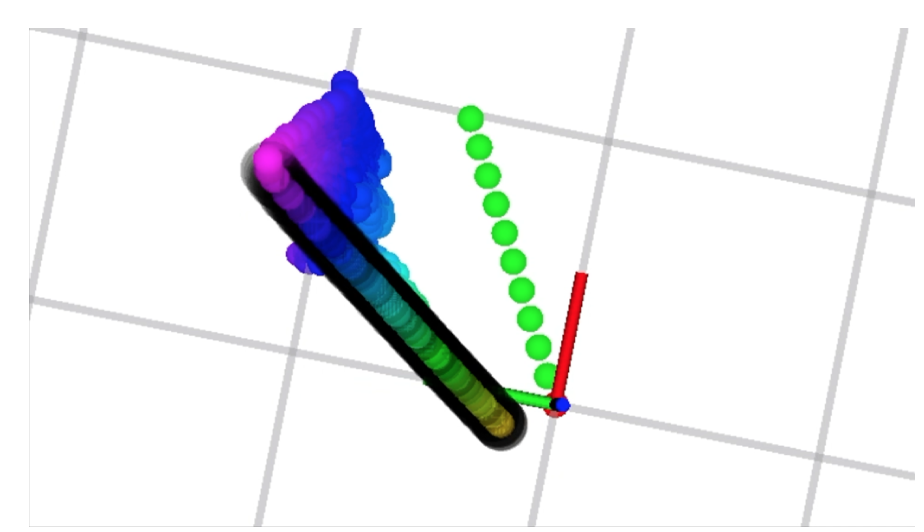}
	\caption{Example wall detected and path defined}
	\label{fig:wf1}
\end{figure}

An example of this is shown in Figure \ref{fig:wf1}.

\subsection{Results from Testing}
This method was tested in a variety of environments with different configurations. This included corners, junctions, and corridors of different widths. This was initially tested with the camera connected to the computer. A video of these results can be found here (\url{https://youtu.be/dhBis5T_pJs}). In this video, the camera is moved towards and away from the wall and a path, based on a proportional gain on the distance to the wall, is created. The desired distance to the wall was 30cm. If the camera was closer than this, the path was pointing away from the wall, and vice versa.

\subsubsection{Issues from Hardware Tests and Improvements}
This test was then implemented on the rollocopter and a few hand-held tests were conducted. This involved carrying the rollocopter near walls and testing the commands sent from the local planner. The following issues were highlighted from these tests.

 \subsubsubsection{Collision Avoidance}
	The wall following algorithm returned one command corresponding to the motion towards or away from the wall. With the current algorithm, there was no check that the path was collision free. Hence, this needed to be combined with the collision avoidance algorithm.
	
	 \subsubsubsection{Field of View Limitations}
	Working with the RealSense camera, the data that could be used was limited to 86$^o$ field of view in the horizontal plane\cite{rs_datasheet}. 
	The algorithm was unable to identify the walls if the wall did not occupy a significant portion of the field of view. As a result, it was found that the vehicle had to be facing less than 45$^o$ away from the wall. Given the non-holonomic constraint, this was not usable on the vehicle when running autonomously.

\begin{figure}[H]
	\centering
	\includegraphics[width=0.35\textwidth]{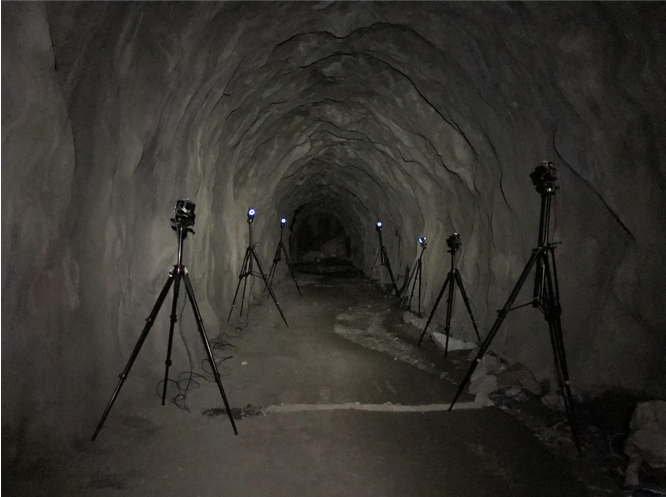}
	\caption{Field test location: Mueller Tunnel, California\cite{jpl_subt}}
	\label{fig:wall_following_mueller_tunnel}
\end{figure}

 \subsubsubsection{Versatility to Different Walls}
	The wall following algorithm worked well for walls which were flat but had not been tested on walls of different shapes. Given that the algorithm relied on plane fitting, it was assumed that the tolerance for fitting walls could simply be adjusted. However, this would lead to the estimated wall position changing significantly at a rapid rate. In addition, this was not compatible with the environment encountered at a field test to be conducted shortly after this development, as shown in Figure \ref{fig:wall_following_mueller_tunnel}.

Given that this man-made tunnel is a reasonable representation of the environments expected in the DARPA challenge, it was decided that this algorithm was unsuitable for the application. Hence, another algorithm had to be designed and implemented.

\subsection{The New Method For Wall Following}\label{sec:new_wf_method}
\begin{figure}[H]
	\centering
	\includegraphics[width=0.6\textwidth]{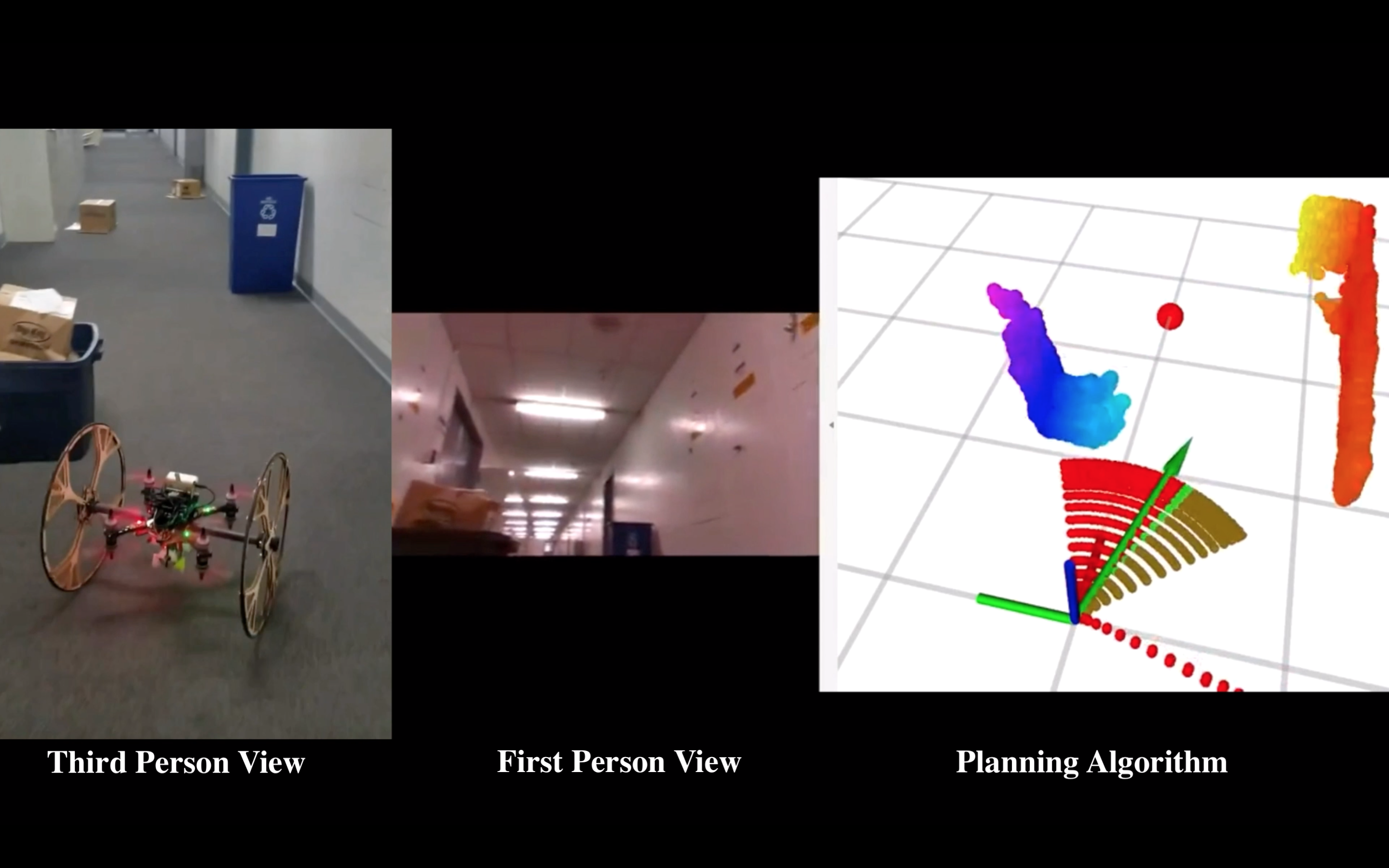}
	\caption{Wall following with collision avoidance}
	\label{fig:wf_snapshot}
\end{figure}

While testing the collision avoidance system, it was found that the walls are also treated as obstacles. As a result, the collision checking algorithm encourages the vehicle to maintain a safe distance form the wall. Hence, the goal is set to be at a constant position with respect to the body frame of the vehicle, at (1,0,0). This leads to the vehicle moving forward while avoiding collisions, thus following the wall. A video of an early test of this behaviour can be found here (\url{https://youtu.be/kLLbE2djRtE}). A preview of this is shown in Figure \ref{fig:wf_snapshot}.

\subsection{Results and Improvements from Hardware Tests}
The tests described in Section \ref{sec:new_wf_method} were very successful. There were no collisions and the overall direction of travel was parallel to the walls, as expected.

Tests were performed to examine the behaviour of the system in more challenging environments such as narrower corridors and corners. The following phenomena were observed.

 \subsubsection{Narrow Corridors}
	The narrower corridor was easier to navigate as there were fewer available primitives which were not in collision. The only option was to continue in the direction of the path, hence encouraging the vehicle to follow the walls precisely.
	
	\subsubsection{Corners}
	 When the vehicle was first tested around corners in the narrow passage, it was observed that the corners would disappear from the field of view. This would cause the vehicle to turn towards this `empty space' and the wheels would brush the corners, sometimes getting caught. This issue was fixed by using the Velodyne 3D LIDAR (with a 360\degr field of view) instead of the RealSense camera (with an 86\degr field of view).
	
	Once this change had been made, it was found that the wheels caused some interference with the cloud. The measurements from the scans through the wheels were showing signs of diffraction around the wheel spokes. Wheels with thinner spokes were made, which appeared to reduce this effect considerably.
	
	\subsubsection{Unstable State Estimation}
\begin{figure}[H]
	 	\centering
	 	\includegraphics[width=0.8\textwidth]{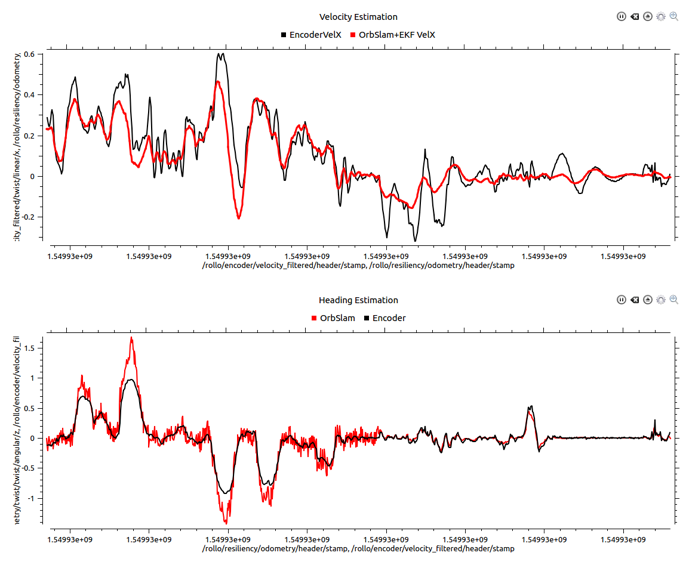}
	 	\caption{State estimation with encoders compared to VIO}
	 	\label{fig:encoders}
	 \end{figure}

	 When rolling, due to the state estimation being unstable, the rollocopter was unable to follow track accurately. This was because the visual odometry was unable to reliably track the motion of the vehicle with both jumps and noise. Encoders were added to both wheels to allow for the body forward velocity and yaw rate to be estimated. Figure \ref{fig:encoders} compares the state estimates from the encoders and the visual-inertial odometry.
	 
	 \vspace{-0.5cm}
	
	As shown, the response is significantly smoother. This leads to the commands being more smoothly tracked. The video above shows how the desired state reached more frequently. Unlike previous tests, the vehicle did not hit any corners.
\newpage
\section{Wall Following in Aerial Mode}
Once the ground mode wall-following behaviour had been implemented, the aerial mode was developed, so as to enable the vehicle to traverse more complicated environments.

\subsection{The Initial Method}
The initial tests for wall following in aerial mode involved implementing the final method used for the ground-mode tests while maintaining a constant altitude. An example of this behaviour can be found here \url{https://youtu.be/cbVKyvlkhpI}.

 This method was tested with localisation information from the VIO algorithm. When testing this algorithm, the safety pilot would manually take off to the desired height, then turn the vehicle to 'Offboard' mode. The autonomous controller would then use the wall following commands from the local planner until the vehicle was stuck. At this point, the safety pilot would activate the emergency-stop and the vehicle would be caught by the tether.

\subsection{Results and Improvements from Hardware Tests}
\subsubsection{Issues Identified Through Testing}
Multiple tests were performed and the following issues were identified.

 \subsubsubsection{Drift in Height}
  Most tests were performed with reliable localisation being used in the control loop (but not the planning loop). In one test, the SLAM algorithm lost track near the start.
 The vehicle followed the walls and exhibited the desired behaviour for approximately 15m, after which the vehicle started drifting downwards. 

 \subsubsubsection{Simplistic Motion}
 This design was very simplistic and allowed the vehicle to fly in a 2D plane only. When using the aerial mode, it is important to have the option of changing the height to move efficiently through the environment, as required. Hence, the range primitives needed to be expanded to provide more options than moving forward or turning in the x-y plane.

\begin{figure}[H]
	\centering
	\includegraphics[width=0.5\textwidth]{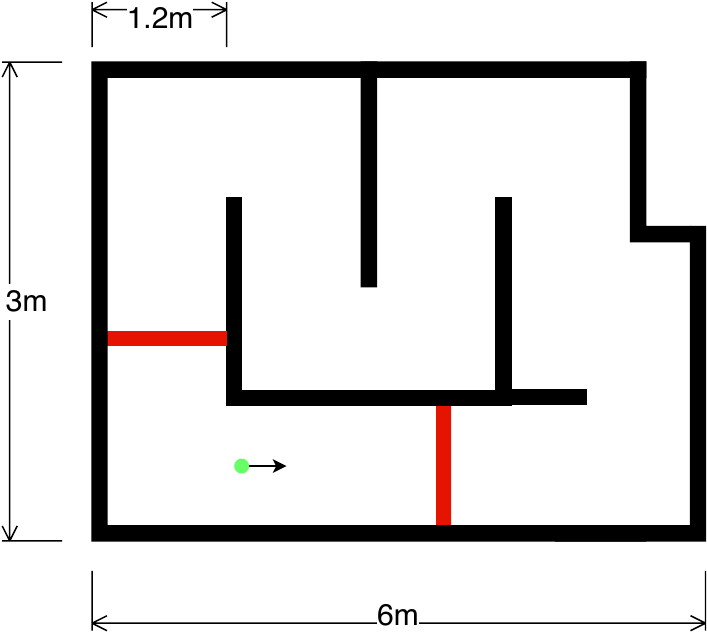}
	\caption{Horizontal sine wave schematic\cite{hybrid_paper}, green = start location, black = maze walls, red = obstacles}
	\label{fig:maze}
\end{figure}

In addition, the project aimed to traverse a variety of complicated environments, such as a horizontal sine wave, a vertical sine wave, and narrow and wide passages. A schematic of the horizontal sine wave maze is shown in Figure \ref{fig:maze}. The vertical sine wave was implemented by putting obstacles on the ground and ceiling inside this maze.

\subsubsection{Improvements Implemented}
To account for the issues and environments above, the following improvements were implemented.

\subsubsubsection{Integration of a Bottom Clearance Sensor}
To account for the drift in height, a bottom sensor was integrated. A RangeFinder 1D LIDAR\cite{range_finder} was integrated with the vehicle and returned the distance to the ground. This allows for the desired height to be planned with respect to the ground. Maintaining a constant height above the ground allows for the option of avoiding drift.

\begin{figure}[H]
		\centering
		\includegraphics[width=0.8\textwidth]{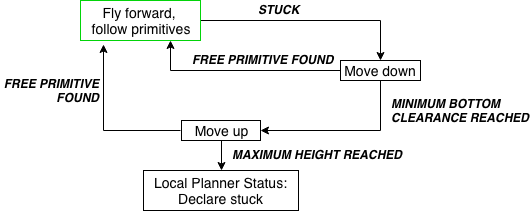}
		\caption{State machine for vertical motion}
		\label{fig:wf_state_machine}
	\end{figure}

 \subsubsubsection{State Machine for Vertical Motion}
 To manage the vertical sine wave, a state machine was created. When the vehicle is stuck, it starts to move down until it reaches a specified minimum distance to the ground. It then moves up until it reaches a defined maximum height. If at any point during this vertical motion the algorithm identifies free primitives, the vehicle will move forward along this primitive and continue in the forward-moving mode. This is illustrated in Figure \ref{fig:wf_state_machine}. The state starts at the green box and aims to remain in this green box at all times.
	
\begin{figure}[H]
		\centering
		\includegraphics[width=0.5\textwidth]{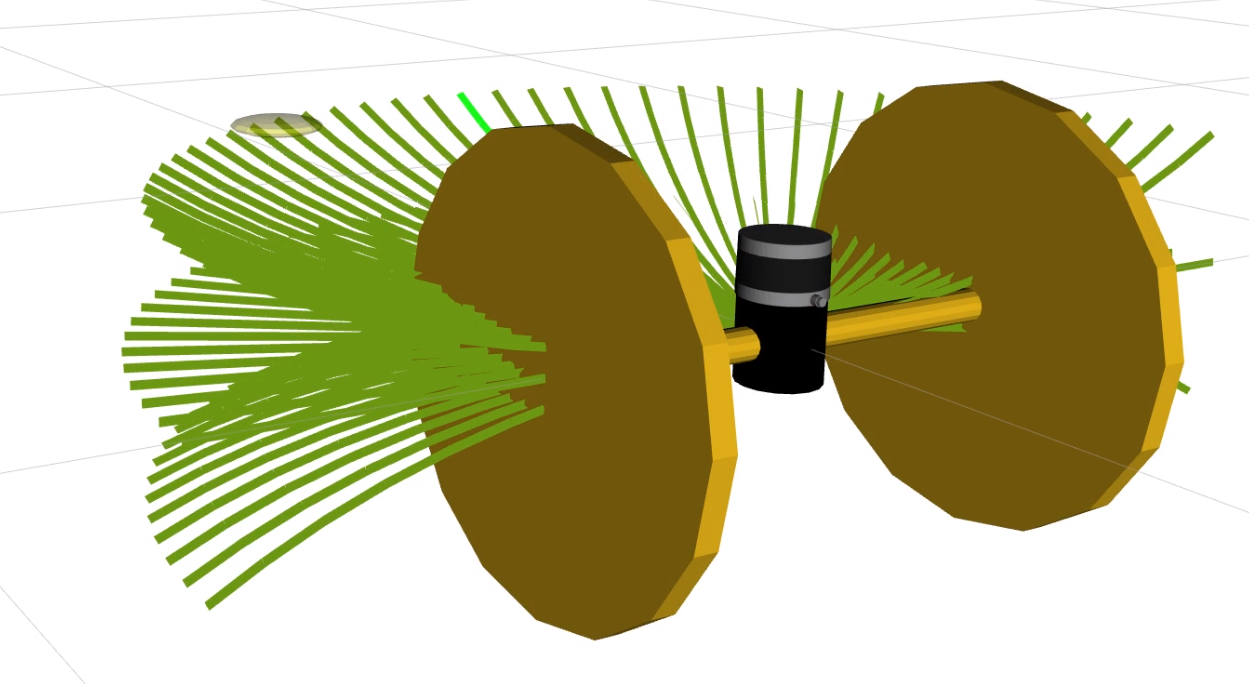}
		\caption{Primitives at different heights}
		\label{fig:height_variation_primitives}
	\end{figure}

	\subsubsubsection{Primitives with End Points of Different Heights}
	 To allow for a greater range of motion, additional primitives were added to allow the vehicle to move up or down. The primitives endpoints were defined as on a sphere around the vehicle. The additional endpoints were added at $\pm 15$\degr to allow for planning in the z axis. This is illustrated in Figure \ref{fig:height_variation_primitives}. The angle of 15\degr was selected as this allows for the primitives to be within the field of view of the LIDAR. This was to avoid the issue with jittering that was observed when first testing the initial collision avoidance algorithm.
	
\begin{figure}[H]
		\begin{subfigure}[b]{0.5\textwidth}
			\includegraphics[width=\linewidth]{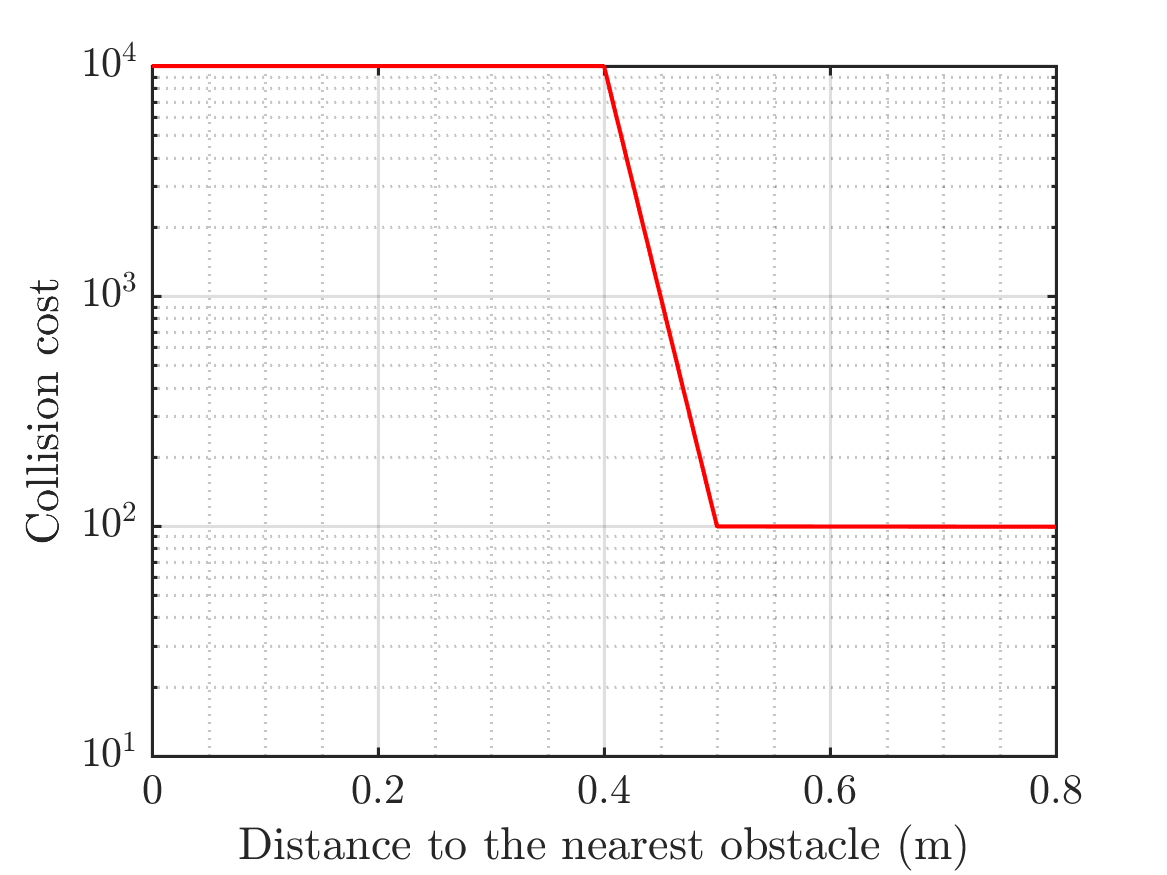}
			\caption{Collision cost}
		\end{subfigure}
		\hfill
		\begin{subfigure}[b]{0.5\textwidth}
			\includegraphics[width=\linewidth]{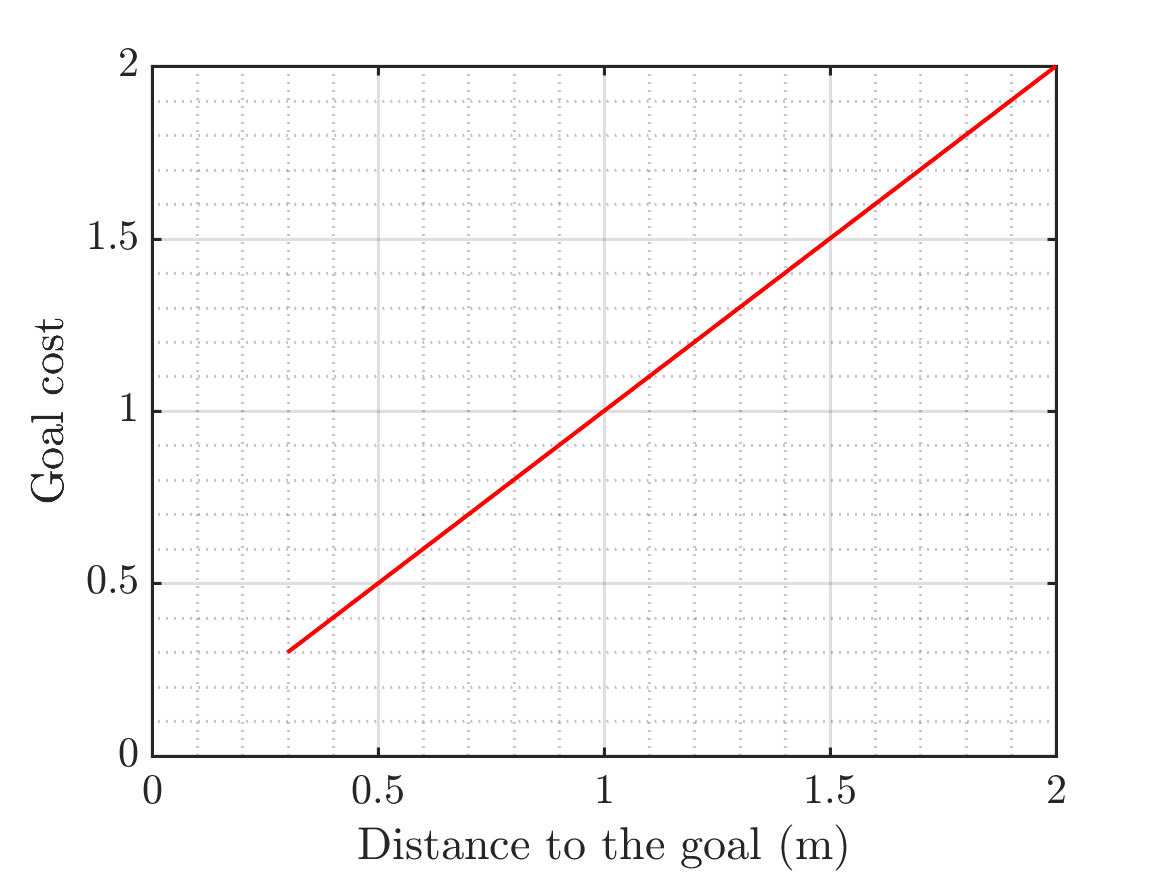}
			\caption{Goal cost}
		\end{subfigure}
		\caption{Cost functions for primitives}
		\label{fig:cost_functions}
	\end{figure}

	 \subsubsubsection{Updated Cost Function with a `Near-Collision' Buffer}
	 It was found that the vehicle was approaching the wall too closely when using the initial collision avoidance algorithm. Hence, it was decided that there would be a `near collision' buffer which would be set as the desired distance from the wall. Hence, a new cost function was devised as the sum of the two plots in Figure \ref{fig:cost_functions}. 
	
	This behaviour causes the planner to prioritise the distance to the nearest obstacle and the distance to the goal. If there are any primitives which do not approach the `near collision' distance, the closest of these to the goal will be selected. If all primitives are either `near collision' or colliding, the best `near collision' primitive (with respect to the goal) will be selected. If all primitives are in collision, the planner will declare that there are no primitives available and will use the state machine. The updated cost function is detailed in Algorithm \ref{alg:new_cost_fcn}. The primitive from which the lowest cost is selected.\\

\begin{algorithm}[H]
	\begin{mdframed}
		\KwIn{Angle between vector to the endpoint and vector to the goal from the current position, $a_g$; distance to the nearest collision, $d_c$, goal cost weight, $c_{gw}$}
		\KwOut{Cost for the primitive}
		\hrule
		
		Set collision cost and near collision cost weights.
		
		\vspace{-1cm}
		
		\begin{align}
		c_{ncol} = c_{gw} \times 100; c_{col} = c_{ncol} \times 100
		\end{align}
		
		\vspace{-0.5cm}
		
		Calculate the collision cost.

		    \If {$d_c$ $\leq$ collision buffer}
		     {
		     	Set collision cost to absolute collision cost.
		     	
		     	\vspace{-1cm}
		     
		     	\begin{align}
		     	C_c = c_{col}
		     	\end{align}
		     	
		     	\vspace{-0.5cm}
		     	
		    } \ElseIf {$d_c$ $\leq$ near collision buffer}
		    {
		    	Set collision cost to near collision cost (100 times less than absolute collision cost)
		    	
		    	\vspace{-1cm}
		    	
		    	\begin{align}
		    	C_c = c_{ncol}-d_c
		    	\end{align}
		    	
		    	\vspace{-0.5cm}
		    	
		    } \Else{Set the collision cost to zero}
		    
		    Calculate the goal cost, based on the axis angle to the goal.

			\vspace{-1cm}

		    \begin{align}
		    C_g = a_g \times c_{gw}
		    \end{align}
		    
		    \vspace{-0.5cm}
		    
		    Calculate the total cost for the primitive.
		    
		    \vspace{-1cm}
		    
		    \begin{align}
		    C = C_c + C_g
		    \end{align}
		    
		    \vspace{-0.5cm}
		    
	\end{mdframed}
	\caption{The updated cost function with the near collision buffer}\label{alg:new_cost_fcn}
\end{algorithm}

\subsubsection{Results from Further Testing}
The following limitations of this new improved method were observed and solutions implemented correspondingly.

 \subsubsubsection{Minimum Depth of the LIDAR}
	Due to the LIDAR having a minimum depth of approximately 50cm, it was found that when an obstacle was closer than this distance to the vehicle, the collision avoidance algorithm would return that the path ahead was still available.

 \subsubsubsection{Assumptions about Ceiling Height}
	Going up to a maximum bottom clearance in the state machine is not robust to different environments and assumes that the distance between the ground and the ceiling is always greater than this specified distance. Hence, it was decided that a top clearance sensor must be added. This allows for the maximum height to be based on the distance to the ceiling. In addition, the top and bottom clearances can be used to maintain a height roughly in the middle, making the vehicle more robust to stalagmites and stalactites. 

 \subsubsubsection{Point-cloud based Height Sensors}
	Another issue regarding the RangeFinder height sensor was that it contained only one measurement. This measurement provided the distance between the height sensor and the point on the ground that was directly underneath the vehicle. Hence, the state estimate of the vehicle varied with the height of the ground and was susceptible to noise. A new height sensor was integrated which returns a point cloud representation of a square below the height sensor. This point cloud can be used to determine the height of the vehicle with respect to the ground and provide insight into any changes in the height of the ground. The cloud from this can also be added to the point clouds used for path planning.

 \subsubsubsection{Limited Field of View in the Elevation Angle}
	While the LIDAR had a 360\degr field of view about the azimuth, the visible range of elevation angles was only  within $\pm15^o$. This meant that the vehicle was then unable to find higher passages which had left the field of view. Initially, a state machine was created to avoid this issue. However, the final solution was to implement a local map. This local map is based on Octomap\cite{octomaphornung2013} and creates a voxel-based point cloud representation of the environment. This map is updated with each new measurement and restarted when localisation is lost. This allows for the map to be used for planning when it can be relied upon, but prevents issues with the map being incorrect with respect to the position of the vehicle. 

\subsubsubsection{Mid-level Planning}
	Creating a local map allows for a more sophisticated planning algorithm. To ensure that an efficient path towards the goal is pursued, the A* planning algorithm was implemented, as described in Algorithm \ref{alg:Astar}, in Appendix \ref{apdx:astar}. Note that the predecessor node refers to the node before the node being selected and the `cost to reach' refers to the distance to the goal from the current node. In this case, the nodes refer to the voxel representation of the environment. When the path has been defined, the next goal is sent to the local planner.

\begin{figure}[H]
	\centering
	\includegraphics[width=0.5\textwidth]{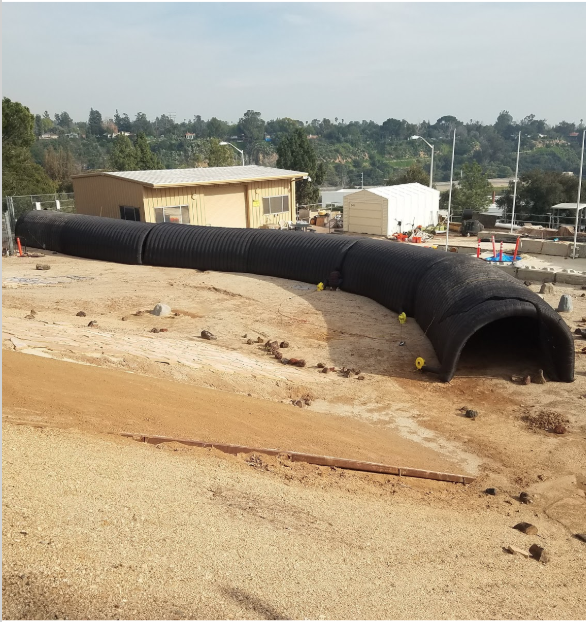}
	\caption{The 6 tunnels through which the rollocopter flew}
	\label{fig:mars_tunnels}
\end{figure}

A video of the aerial wall following behaviour through six tunnels in a dusty representative environment can be found here \url{https://youtu.be/rHJu6qHW_fo}. The environment through which the rollocopter flew was the JPL Mars Yard, as pictured in Figure \ref{fig:mars_tunnels}.

\newpage
\section{Hybrid Wall Following}\label{sec:hybrid_wf}
Given that wall following was functional in both the aerial and ground modes, the final step was to test both modes together.

There were two main considerations when extending to hybrid functionality. There is the planning phase and the execution phase.

\vspace{-0.5cm}

\subsection{Hybrid Planning}
The planning phase involved extending the mid-level planner to account for the difference between rolling and flying and the ability to plan for both.

\begin{figure}[H]
		\begin{subfigure}[b]{0.5\textwidth}
			\centering
			\includegraphics[width=0.8\linewidth]{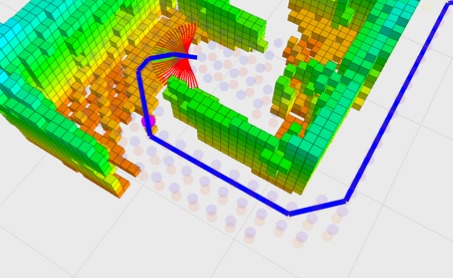}
			\caption{Rolling around the corner}
		\end{subfigure}
		\hfill
		\begin{subfigure}[b]{0.5\textwidth}
			\centering
			\includegraphics[width=0.8\linewidth]{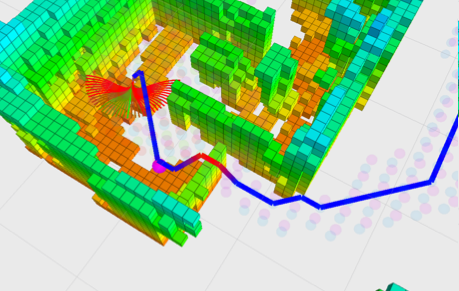}
			\caption{Obstacle appears in field of view}
		\end{subfigure}
		\vfill
		\begin{subfigure}[b]{0.5\textwidth}
			\centering
			\includegraphics[width=0.8\linewidth]{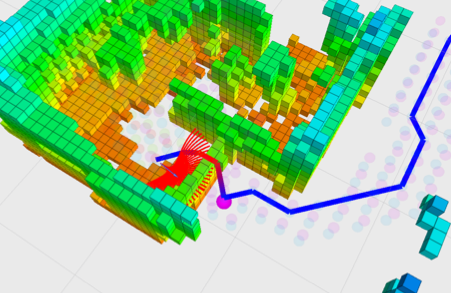}
			\caption{Transitioning to aerial mode}
		\end{subfigure}
		\hfill
		\begin{subfigure}[b]{0.5\textwidth}
			\centering
			\includegraphics[width=0.8\linewidth]{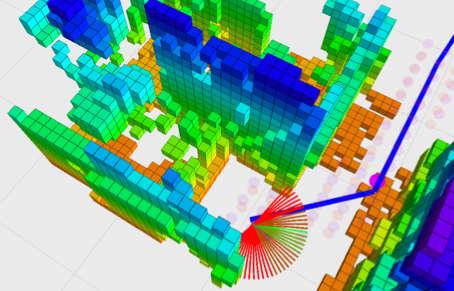}
			\caption{Rolling after the obstacle}
		\end{subfigure}
		\caption{Planning sequence for hybrid motion}
		\label{fig:hybrid_planning_sequence}
	\end{figure}

To encourage the hybrid behaviour, the cost of a node in the air five times greater than the cost of a node on the ground. This leads to the vehicle rolling where possible and only flying over obstacles which cannot be rolled around. This leads to the behaviour demonstrated in Figures \ref{fig:hybrid_planning_sequence}a, \ref{fig:hybrid_planning_sequence}b, \ref{fig:hybrid_planning_sequence}c and \ref{fig:hybrid_planning_sequence}d.

In this sequence, the vehicle rolls around the corner, encounters the barrier, takes off, flies over the barrier, lands, then continues rolling around the remainder of the course.
	
\subsection{Executing Hybrid Functionality}
The main challenge of executing the hybrid behaviour is managing the transitions between rolling and flying.	

Regarding the management of the transitions, a new client for the Mobility Services node, described in Section \ref{sec:mobility_services_overview}, was developed. This client receives a goal from the hybrid planner and converts the transition to the next goal into a sequence of behaviours to be sent to the Mobility Services node.

\begin{itemize}[leftmargin=1cm,rightmargin=1cm]
	\item[] If the previous and next nodes are both in the same mode (aerial or ground mode), the goal is sent as a coordinate to be flown or driven to. 

	\item[] If the previous node was a rolling node and the next node is a flying node, the client will request that the vehicle take off, then fly to the goal. 

	\item[] Finally, if the previous node was a flying node and the next node is a rolling node, the client will request that the vehicle land, then roll to the goal.
\end{itemize}

These commands are then converted to goals for the local planner by the Mobility Services node.

When deciding to land or roll, the top and bottom height sensors are also used. If the top clearance is too low, the vehicle will not take off and will declare the `stuck' status. If the bottom height sensor point cloud has a high variance, it will be decided that the ground is too rough or too sloped to land or roll on and a flying command will be sent.

A video of the hybrid wall following mode working can be found here (\url{https://youtu.be/dFMToIPrCa0}). This is from the same test as in Figures \ref{fig:hybrid_planning_sequence}a, \ref{fig:hybrid_planning_sequence}b, \ref{fig:hybrid_planning_sequence}c and \ref{fig:hybrid_planning_sequence}d.
\newpage
\section{Summary}
This chapter covered the augmentation of the local planner to support a wall following behaviour. This allows for the capability of navigating without localisation. The first stage involved the basic ability to roll along the ground avoiding walls and obstacles. This was then extended to flying and the algorithm enhanced to utilise the increased mobility provided by flying. Finally, the two modes were combined into the hybrid wall following capability which allows the vehicle to move efficiently with the mobility required to traverse the environment. To fully benefit from the hybrid functionality, the vehicle must also consider the local terrain and thus a more reliable estimate of the efficiency of rolling. More detail is provided in Chapter \ref{chap:traversability}. 
	\cleardoublepage
	\chapter{Ground Traversability}\label{chap:traversability}
\attributions{All work in this chapter was completed by the author of this thesis.}
\section{Overview}
When working in the ground mode, or deciding when to transition during hybrid motion, the rollocopter must be able to identify what terrain can be traversed and what cannot. Being robust to different types of terrain involves identifying which types of terrain can be traversed, and defining a cost function to consider the effects of the terrain when selecting the optimal path.

\section{Terrain Classification}\label{sec:trav_categories}
The terrain can be classified into three main categories. 

\begin{itemize}[leftmargin=1cm, rightmargin=1cm]
\item[] \textbf{Easily traversable}\\
 This terrain is flat enough to not have any unexpected effects on the controller. In this case, no traversability cost function is required.

 \item[] \textbf{Difficult to traverse}\\
  The ground is rough or sloped such that the effects of the terrain must be accounted for, but is traversable. This category requires a cost on traversability which can be compared to the cost of taking off.

 \item[] \textbf{Untraversable}\\
  The ground cannot be rolled upon. This is the case if there is extreme slope, large rocks, or water.  The traversability cost of this category must be prohibitive, equal to the cost of a collision.
\end{itemize}

Categorising the terrain can be achieved by performing a traversability analysis. There are two main aspects of this. 

\begin{itemize}
	\item Identifying the capabilities of the hardware
	\item Creating an algorithm capable of processing sensory information about the ground and identifying what can be traversed
\end{itemize}

These tasks are detailed in Sections \ref{sec:trav_hardware_capability} and \ref{sec:trav_analysis}.

\section{Identifying the Capabilities of the System}\label{sec:trav_hardware_capability}
\begin{figure}[H]
	\begin{subfigure}[b]{0.3\textwidth}
		\includegraphics[width=\linewidth]{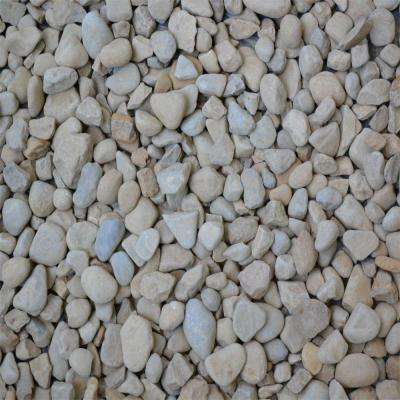}
		\caption{Gravel/Dirt}
	\end{subfigure}
	\hfill
	\begin{subfigure}[b]{0.3\textwidth}
		\includegraphics[width=\linewidth]{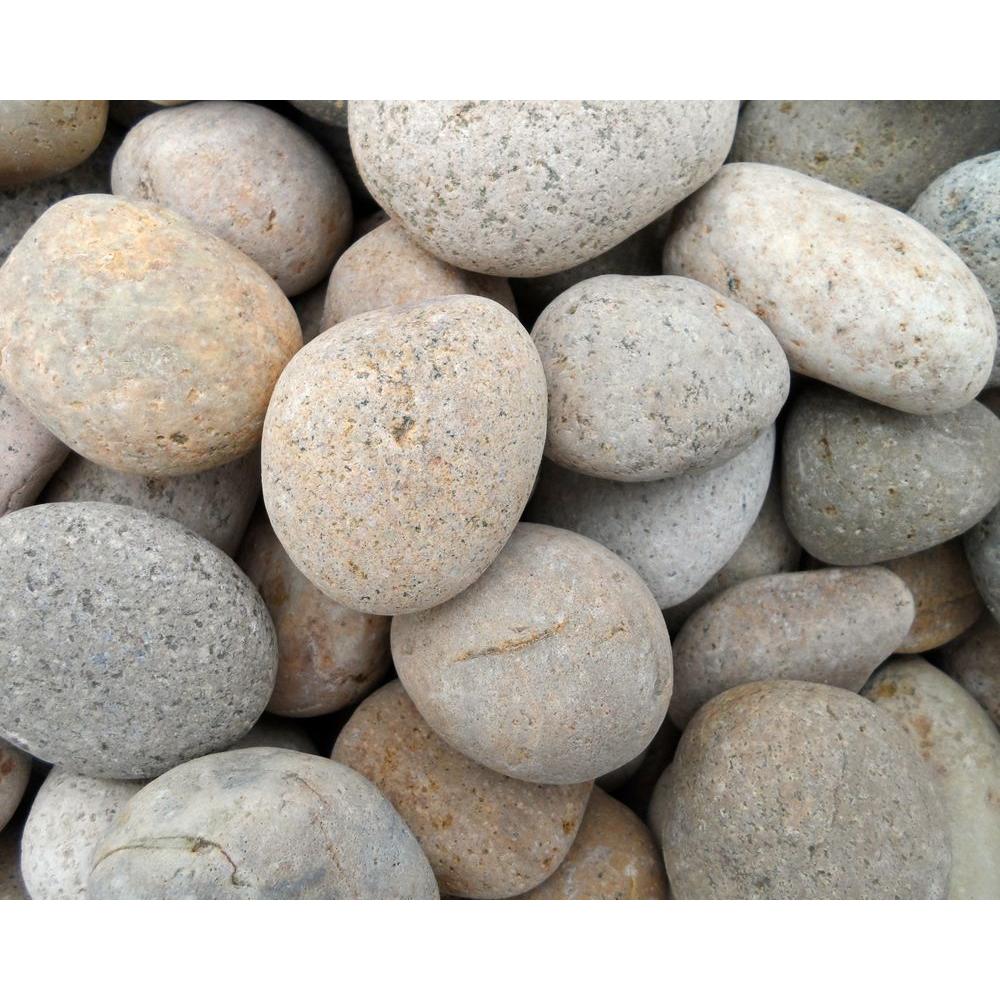}
		\caption{Larger rocks, 3-5cm}
	\end{subfigure}
	\hfill
	\begin{subfigure}[b]{0.3\textwidth}
		\includegraphics[width=\linewidth]{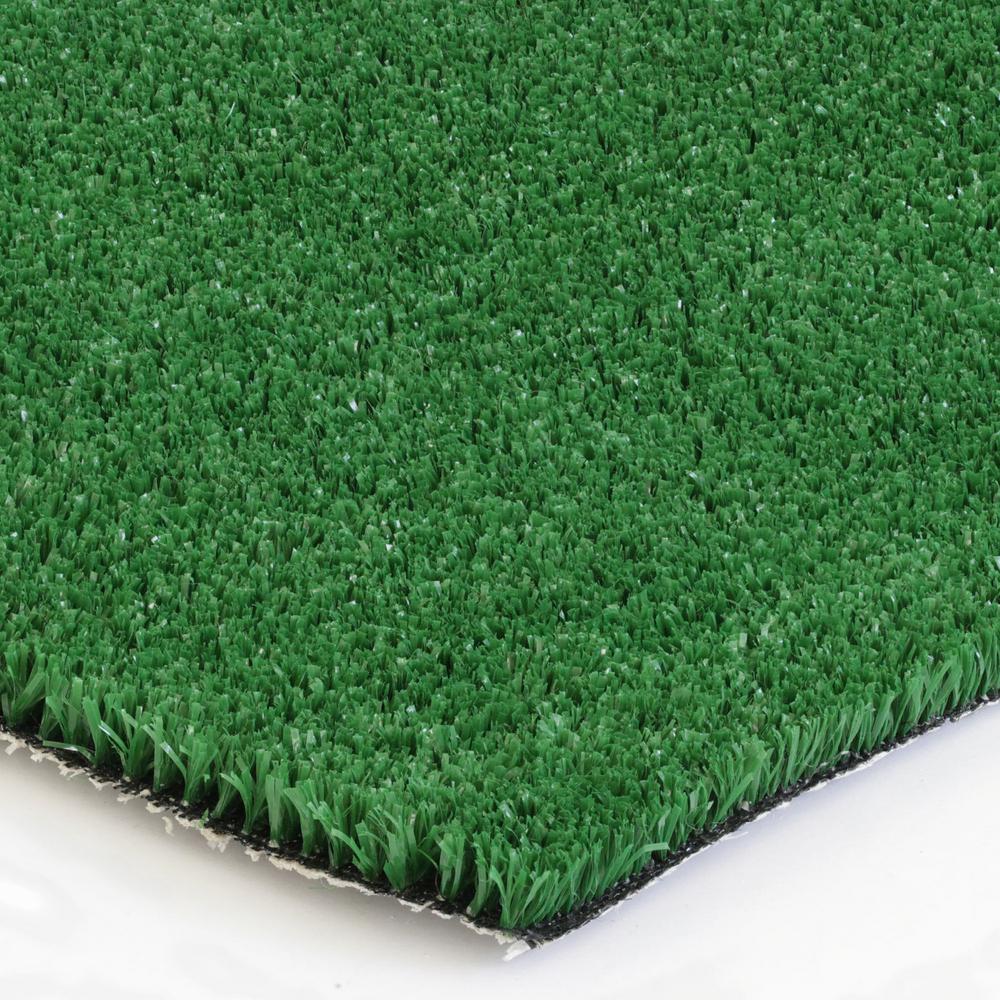}
		\caption{Artificial grass}
	\end{subfigure}
	\caption{Terrain used for testing the rollocopter\cite{home_depot}}
	\label{fig:terrain_types}
\end{figure}

The rollocopter was tested on different types of terrain such as the gravel, rocks, and grass shown in Figure \ref{fig:terrain_types}.

\subsection{Effects of Rough Terrain}
When testing this, the following behaviours were observed.

Firstly, when one wheel would get caught, there would be a sudden yaw motion. This would also cause localisation loss as the images from the camera would be subjected to motion blur which would cause the SLAM algorithm to lose track.

Secondly, if both wheels were stuck, the vehicle would be motionless while still applying the forward force.


\subsection{Solutions for Mitigating the Effects of the Terrain}
The avoid the sudden yaw motions when a single wheel was caught, the proportional gain for the yaw PD controller was increased. This led to the vehicle being less susceptible to such disturbances.

Regarding the effects of both wheels being stuck, the following solutions were considered.

\subsubsection{Adding an Integrator}
 Firstly, it was thought that an integrator could be added to encourage the vehicle towards the goal. However, given that the goal is always in the body frame, the effects of integrator wind-up would be observed. This involves the integral term constantly increasing due to the non-converging error known as integrator wind-up. This would cause a constant increase in the thrust which would lead to instability.

\subsubsection{Increase the Vertical Component of the Thrust force}
 The second solution was to increase the vertical component of the thrust force. With less force exerted on the vehicle by the ground, the vehicle would be able to skim the surface. While this allows for the vehicle to traverse the rougher terrain, it leads to unnecessary power usage when traversing smoother ground.

\subsubsection{Perform a Traversability Analysis}
 A traversability analysis on the ground ahead can be used as a feed-forward term for the controller. If the terrain is rougher or more sloped, the controller can compensate. This compensation could be achieved by having a form of auto-tuning where the controller gains are adjusted based on the nature of the environment.

\subsubsection{Force Detection}
 Force detection is another method which has been considered and pursued in parallel. This involves identifying external forces on the vehicle based on comparing the acceleration from the IMU to the forces exerted by the vehicle. If the vehicle is not moving despite a forward thrust force, it can be determined that there is an external force blocking the vehicle. If the position of this thrust force can be determined, then it can be compensated for by commanding a velocity either tangential to the force or in the opposite direction. This approach is based on using feedback from the ground for planning and control.

\section{Traversability Analysis}\label{sec:trav_analysis}
The goal of this analysis was to classify the terrain into the categories defined in Section \ref{sec:trav_categories}. Through the tests completed above, and the literature review in Section \ref{sec:trav_lit_review}, it was determined that the main characteristics to be observed were the roughness and the slope. The desired output was a map showing sections which are easily traversable, difficult to traverse, or untraversable.  The first step was to create an elevation map of the environment, then roughness and slope maps, which could be combined using a cost function to create the traversability map.

\newpage
\subsection{Creating the Elevation Map}
The sensory input for traversability analysis is the point cloud from the RealSense camera. The cloud was transformed into the gravity-aligned frame using the same method as per Section \ref{sec:local_planner}.

\begin{figure}[H]
\begin{subfigure}[b]{\textwidth}
	\centering
	\includegraphics[width=0.6\textwidth]{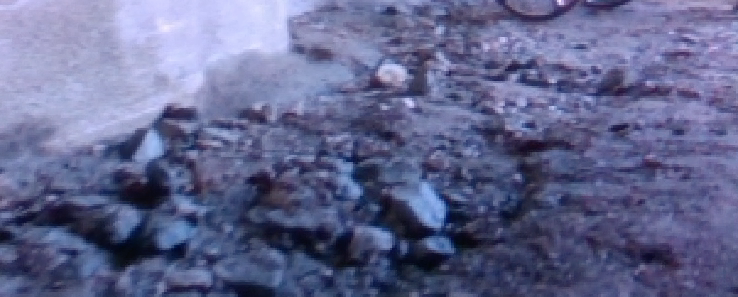}
	\caption{Environment}
	\label{fig:image_trav}
\end{subfigure}
\vfill
\begin{subfigure}[b]{\textwidth}
	\centering
	\includegraphics[width=0.6\textwidth]{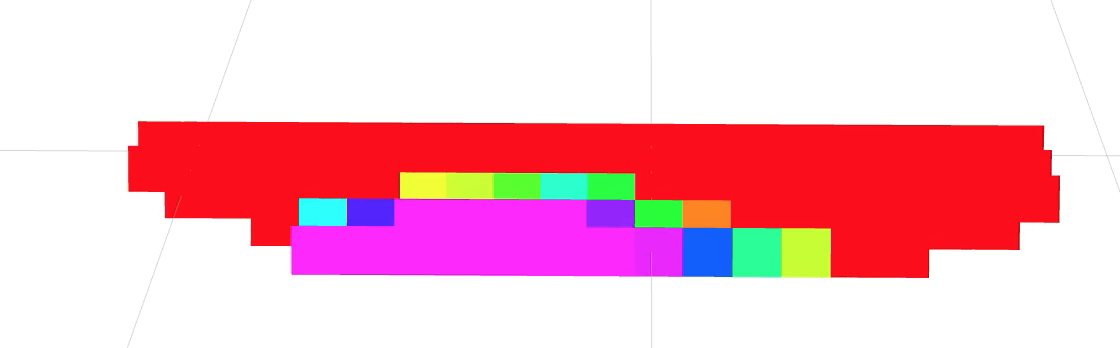}
	\caption{Elevation map}
	\label{fig:elevation_map}
\end{subfigure}
\end{figure}

The cloud was filtered to include only the ground by considering the part of the cloud which was below the height of the vehicle. The heights of these points were then used to create an elevation map, such as that shown in Figure \ref{fig:elevation_map}. The elevation map is represented as a 2 dimensional cloud where the intensity represents the height.

\subsection{Creating the Roughness and Slope Maps}
The elevation map can be discretised into a grid of a chosen resolution. The function loops through all the squares in this grid, and identifies all the points inside the grid square (using a KD-tree search on the 2D elevation map cloud). These points are then analysed as in Algorithm \ref{alg:traversability} to create the roughness and slope maps. The roughness is defined as the variance of the ground to the average plane of the local area and the slope is characterised as the angle between the plane and the horizontal plane.\\

\begin{algorithm}[H]
	\begin{mdframed}
		\KwIn{Small point cloud of a section of the ground}
		\KwOut{Variance and slope of this section}
		\hrule
		\If{the grid square cloud has at least four points}
		{
			Fit a plane to the grid square cloud using the PCL Library RANSAC function and save the coefficients in the form of Equation \ref{eqn:plane_coeffs_form}\;
			
			\vspace{-1cm}
			
			\begin{align}
			ax + by + cz + d = 0\label{eqn:plane_coeffs_form}
			\end{align}
			
			Compute and save the variance of this grid square cloud with respect to the identified plane\;
			
			Calculate the normal vector for the plane as per Equation \ref{eqn:plane_normal_vector}\;
			
			\vspace{-1cm}
			
			\begin{align}
			\begin{bmatrix}
			\frac{a}{\sqrt(a^2 + b^2 + c^2)} & \frac{b}{\sqrt(a^2 + b^2 + c^2)} & \frac{c}{\sqrt(a^2 + b^2 + c^2)}
			\end{bmatrix}\label{eqn:plane_normal_vector}
			\end{align}
			
			Calculate the angle between the normal vector and the vertical vector $[0, 0, 1]$\;
			
			\vspace{-1cm}
			
			\begin{align}
			\theta = acos\left(\frac{|n||v|}{n \bullet v}\right)
			\end{align}
			
			Use this angle to determine the angle between the grid square cloud plane and the gravity-aligned x-y plane\;
		
		}
	\end{mdframed}
	\caption{Roughness and slope map creation}
	\label{alg:traversability}
\end{algorithm}

The following roughness and slope maps were obtained using this information. As per the elevation map, the colour scheme is rainbow and purple corresponds to a low value and red corresponds to a high value.

\begin{figure}[H]
	\begin{subfigure}[b]{0.5\textwidth}
	\centering
	\includegraphics[width=\textwidth]{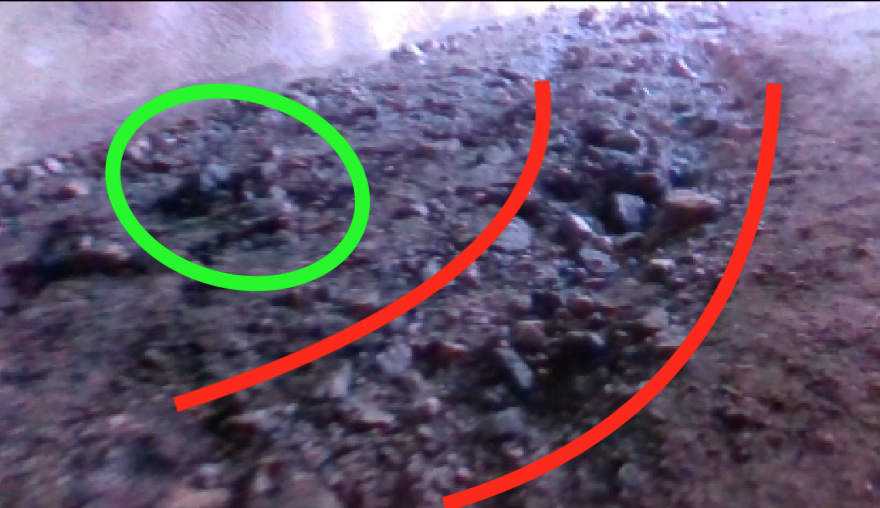}
	\caption{Environment}
	\label{fig:roughness_map}
	\end{subfigure}
	\hfill
	\begin{subfigure}[b]{0.5\textwidth}
	\centering
	\includegraphics[width=\textwidth]{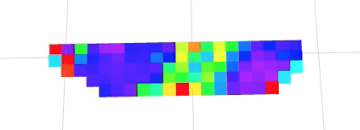}
	\caption{Roughness map}
	\label{fig:slope_map}
	\end{subfigure}
	\caption{Roughness results}
	\label{fig:trav_results}
\end{figure}

The distinct parts of the terrain have been identified with the red lines. As shown in Figure \ref{fig:trav_results}, there is a rough patch, illustrated with high intensity points in the roughness map, in the centre and there are smoother patches outside of this. The highly rough patch on the left, corresponds to the rocks circled green in Figure \ref{fig:trav_results}a.

\begin{figure}[H]
	\begin{subfigure}[b]{0.5\textwidth}
		\centering
		\includegraphics[width=\textwidth]{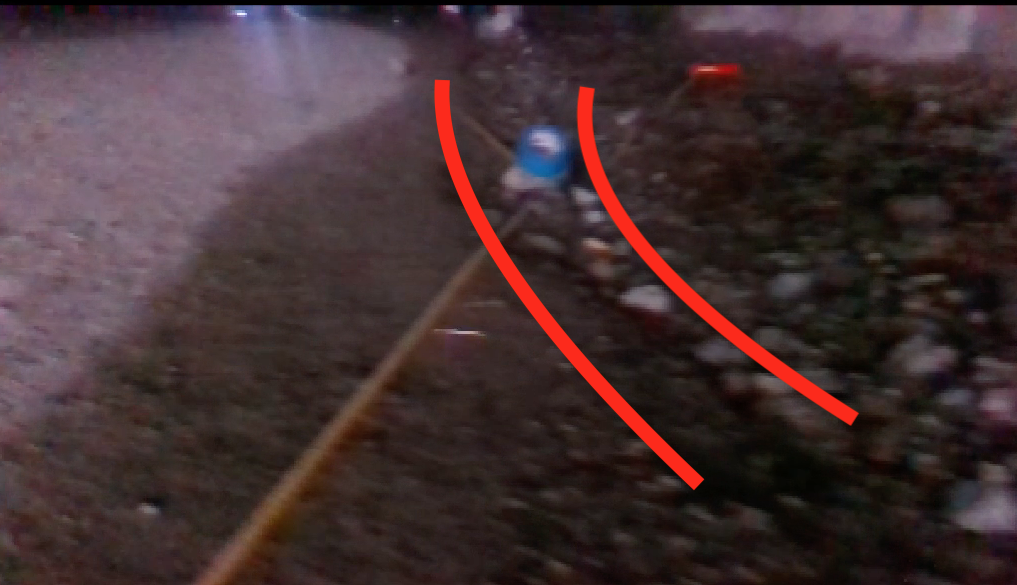}
		\caption{Environment}
		\label{fig:roughness_map1}
	\end{subfigure}
	\hfill
	\begin{subfigure}[b]{0.5\textwidth}
		\centering
		\includegraphics[width=\textwidth]{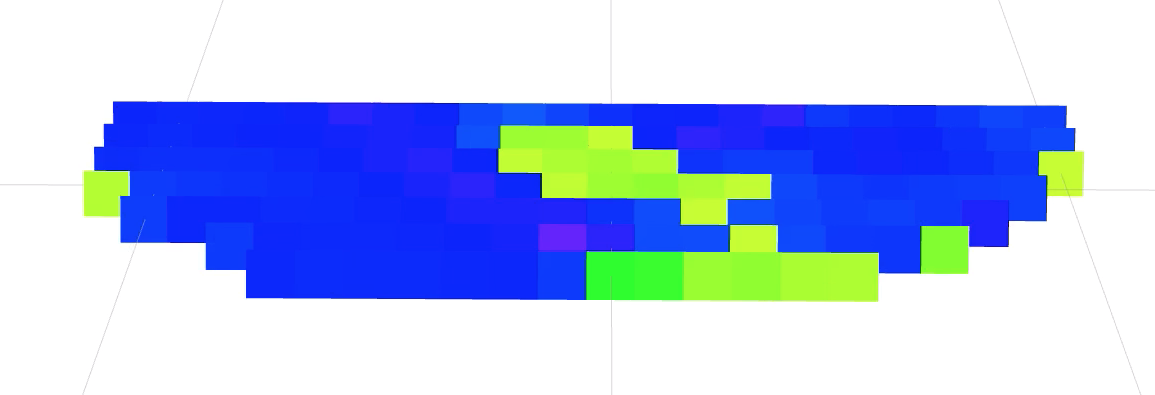}
		\caption{Slope map}
		\label{fig:slope_map1}
	\end{subfigure}
	\caption{Slope results}
	\label{fig:trav_results1}
\end{figure}

In Figure \ref{fig:trav_results1}a, there terrain is mostly flat, except for a sloped area and which leads to a lower elevation. The map in Figure \ref{fig:trav_results1}b shows the two less sloped areas as blue, and the more sloped area in the centre as green. Note that the ground was sloped to the left in Figure \ref{fig:trav_results1}a, hence the slope map in the more flat regions is blue and not purple. The slope map shows the magnitude of the axis-angle between the vector perpendicular to the ground plane in the grid square, and the gravity-aligned vertical vector.
 
\subsection{Grid Resolution}
It is important to choose the grid resolution such that the map is sufficiently detailed but computational time is also reasonable. A grid with a 2cm resolution took around 23 seconds to create. Given that the rollocopter currently moves at approximately 0.3-0.5m/s, the rollocopter would have exited the field of view of the map before the map had been rendered. In comparison, a grid with 40cm resolution may not provide enough information to plan around in a narrow mine shaft. Further testing is required to determine the  optimal resolution for the traversability map. However, it is known that the relationship between the grid resolution (length of the side of each grid square) and the computational time is quadratic. This is because the ground area is discretised into the grid and the algorithm loops through  the grid, processing each square independently.

\section{Current Challenges}
Unfortunately it was decided that the traversability analysis was not required for the project at this time. In addition, the sensor configuration was changed such that the RGBD camera no longer provided information about the ground. This led to the points on the ground being from the Velodyne LIDAR which does not have a high enough resolution for points at a different altitude to the sensor (such as the ground). Hence, this work was stopped. The following tasks and challenges require addressing before this work can be used.

\subsubsection{Integration with the latest planning architecture}
Unfortunately more hardware tests are required to define the variance and slope boundaries for the classifications of the terrain for use in the cost function. In future, the traversability analysis will be combined with the local map of the environment and the terrain cost function will be considered when calculating the optimal path using A*.

\subsubsection{Water and grass-covered patches}
While the roughness and slope are useful characteristics for identifying a variety of untraversable terrains, there are some cases where this analysis is not sufficient. For instance, a body of water is not traversable but will have no slope or variance in a mine. Another case is when there is grass covering the ground. While the grass may present a high variance, it is possible that the ground is quite smooth.

\newpage
\section{Summary}
This chapter provided context on the development of the traversability analysis and the necessary hardware system capability classification. While this has not been implemented on the vehicle, there is potential for the traversability map to be integrated with the local map in detailed in Chapter \ref{sec:wall_following}. This would allow for a more informed decision of when to transition between the aerial and ground modes and reap the full benefit of a hybrid system.
	\cleardoublepage
	\chapter{Unit Test Simulator for Autonomy}
\attributions{All work in this chapter was completed by the author of this thesis. The 3D model of the rollocopter and the Velodyne sensor plugin were provided by another member of the Co-STAR team.}
\section{Overview}
The team project had a high fidelity simulator with a complicated tunnel-like environment but it required an extremely powerful computer (around 20 cores) to run. Hence, it was decided that a second low-fidelity simulator would be developed. This simulator does not accurately represent the vehicle dynamics but allows for all autonomy and planning systems, such as those covered in this thesis, to be tested on the computers commonly used for development.

\section{Software Architecture}
A Gazebo simulator was created containing an environment and a basic model of the rollocopter. A model of the Velodyne VLP-16 LIDAR\cite{simulation_velodyne} was imported and is capable of generating point clouds based on the orientation of the LIDAR and the surrounding environment. In addition, sonar sensors were added to the vehicle to provide the distance between the ground and the wheels. The Gazebo simulator vehicle dynamics were disabled to improve the efficiency of the simulation and avoid interference with the state calculations from the low-fidelity simulator written in C++.

\begin{figure}[H]
	\centering
	\includegraphics[width=\textwidth]{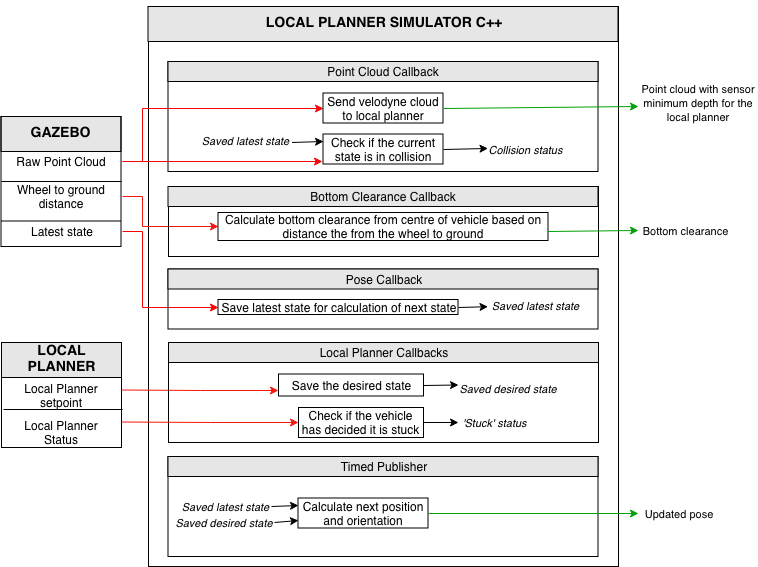}
	\caption{Overall software architecture of the simulator C++ node}
	\label{fig:simulator_software_architecture}
\end{figure}

To work with the autonomy and planning systems, a new node was written which to work with both the Gazebo simulation and the local planner. Figure \ref{fig:simulator_software_architecture} illustrates the overall structure of the node. 

\subsection{Point Cloud Callback Function}
The point cloud callback function handles all cloud processing. The following tasks are completed.

 \subsubsection{Publish Velodyne Cloud}
	The new cloud is created based on the input cloud, but with all points within a specified radius of the origin removed. This is to simulate the minimum depth of the sensor. Empirically, this has been observed as approximately 0.5m. The removal of these points is achieved by first performing a KD-tree radius search (from the point cloud library (PCL)\cite{pcl}) about the origin to identify the points, then extracting the points using the \textit{PCL Extract Indices} class. This new cloud is then published to the topic which is read by the local planner.
	
	 \subsubsection{Process Raw Point Cloud}
	This function processes the initial input cloud and returns a cloud that can be used to check for collisions. This involves transforming the cloud into the world frame using the same algorithm as in the Local Planner (see Chapter \ref{chap:LocalPlanner}). The cloud is also down-sampled and filtered, allowing for a more efficient collision check.
	
	 \subsubsection{Check the Collision Status}
	The function then performs another KD-tree search to check if there are any points within the size of the vehicle. If there is, the simulation is ended as the vehicle has collided with an obstacle.

\subsection{Bottom Clearance Callback Function}
The gazebo simulator has two sonar sensors, one at the base of each wheel, which provide the respective distances to the ground. This can be used to calculate the value equivalent to the bottom clearance provided by the LIDAR-based height sensor on the real rollocopter. The bottom clearance that is then published by the C++ simulator is calculated based on Equation \ref{eq:sim_bottom_clearance}, where $d_{bc}$ is the bottom clearance, $d_{s_x}$ refers to the distance measured by sonar sensor x, and $r_{wheel}$ is the radius of the wheel.
\vspace{-1cm}

\begin{align}
	d_{bc} = min\left[d_{s_1}, d_{s_2}\right] + r_{wheel}\label{eq:sim_bottom_clearance}
\end{align}

\subsection{Pose Callback Function}
This function simply saves the latest state provided by the gazebo simulator as a global variable so it can be used when calculating the next pose.

\subsection{Local Planner Callback Functions}
There are two topics published by the local planner which are required by the simulator.
\subsubsection{Local Planner Status Callback} 
Firstly, there is the local planner status. This topic informs the simulator if the vehicle is stuck. If this is the case, the simulation is ended as the planner cannot find any possible paths.

\subsubsection{Local Planner Setpoint Callback}
The local planner determines the optimal path and sends a corresponding setpoint. This desired state is saved as a global variable to be used when calculating the next pose.

\subsection{Timed Publisher Function}
This function calculates the dynamics based on the time since the previous iteration. The calculation varies for the different modes, as follows.
\begin{itemize}
	\item Mobility mode (aerial or ground)
	\item Control mode (position or velocity control)
	\item Goal frame (whether the desired state is with respect to the odometry frame or the body frame)
	\item Goal frame for z (can be with respect to the odometry frame, body frame, distance to the ground, or distance to the ceiling)
\end{itemize}

The algorithms used in each of the separate cases are included in Appendix \ref{apdx:simulator_algorithms}. In each case, however, the following procedure was followed. Note that the simulator assumes there is no roll or pitch to decrease the computational load, as is the role of the low-fidelity simulator.

\begin{enumerate}
	\item Calculate the time since the last iteration, to use as the time step to the next state.
	\item From the desired state message, calculate the new velocity in the forward direction and yaw in the body frame.
	\item Calculate the corresponding x-y coordinate and yaw in the simulator world frame.
	\item Calculate the new z coordinate based on the desired state.
\end{enumerate}

\newpage
\section{Extending to Multiple-Environment Unit Tests}
For efficiency, an additional feature was created such that the simulator would test the autonomy system with a variety of environments. This was to ensure that the system was not over-designed towards a particular environment. When testing on hardware, significant effort is required to set up testing environments. Hence, versatility in the testing environments is difficult to obtain. The simulator has therefore been designed to test a wide range of environments in one run.

A Gazebo `world' was created based on a 3D model of multiple courses. The rollocopter initial position is set based on the starting position for the initial environment. The autonomy system then attempts the course. The environment is considered complete when one of the following occurs:

\begin{itemize}[leftmargin=1cm,rightmargin=1cm]

 \item[] \textbf{The vehicle gets stuck}\\
	This occurs when the local planner declares that there is no possible path available. In this case, the simulator declares the `stuck' status.
	
	 \item[]  \textbf{The vehicle collides with an obstacle}\\
	As in the local planner, a KD-tree on a point cloud representation of the local environment is used to determine the distance to the nearest obstacle. In the simulator, the query point is set to be the position of the vehicle. If there is any obstacle within a radius of the rollocopter (based on the vehicle size), the `collided' status is declared.
	
	 \item[]  \textbf{The course is successfully completed}\\
	If the vehicle reaches a position within a specified x-y range corresponding to the end of the course passage, the `successful' status is declared. 
	
	 \item[]  \textbf{The simulation reaches a `timeout'}\\
	If a given course has been in progress for more than one minute, it is assumed that the vehicle has entered an infinite loop such as a circular or jittery motion. In this case, the `timeout' status is declared.

\end{itemize}
\newpage

\subsection{User Interface}
The user interface is a YAML file containing parameters which specify which environments to test and what starting coordinates to use. In addition, the x and y coordinate ranges for determining a successful test are also specified here. This allows for different aspects of the system to be tested. For instance, if one is testing the rolling mode, courses which can only be traversed in aerial mode are not applicable. Also, if frontier exploration is to be tested, one is more interested in the identification of forks in the tunnels than the ability of the vehicle to navigate through narrow corridors with obstacles.

\subsection{Environment Management}
The environments are managed in a structure similar to a state machine. 

An \textit{Environment} struct was defined which contains the following information:
\vspace{-0.5cm}

\begin{itemize}
	\item Environment name
	\item Start position
	\item Valid end coordinate ranges
	\item Status (not testing, incomplete, in progress, stuck, collided, successful, timeout)
	\item Start time
	\item Ceiling height (used for collision checking outside the point cloud field of view)
\end{itemize}

When the environments are first loaded, an array of Environment objects is created based on the information in the parameter file. Using the ManageEnvironments() function, the first environment is set. The simulator then runs as before except checking for a status update in each iteration.

\begin{itemize}
	\item When a point cloud is received, the collision check is performed. 
	\item When a LocalPlannerStatus message is received, the status of the local planner (`Running' or `Stuck'), the environment status is updated accordingly, if required.
	\item When a new pose is calculated, this is checked against the end coordinates range to see if the vehicle has successfully completed the course.
	\item In each iteration, the time since the environment was started is checked against the `timeout' time.
\end{itemize}

When there is a status update, the ManageEnvironments() function is called. This checks the status of each environment and chooses the first incomplete environment in the list which is to be tested. When a new environment is selected, the following procedure is followed.

\begin{enumerate}
	\item Set the status of this environment to `IN PROGRESS'
	\item Update the state of the vehicle to the start of this environment
	\item Set the environment start time to the current time
	\item Reset the flags of `collided', `successful', etc. to false
\end{enumerate}

If there are no environments left, the final message is printed to the screen. This includes a list of the environments, the status of each, and, if the vehicle collided or was stuck, the time is provided. This allows for the user to check the recorded data at the time of interest for debugging purposes.

\begin{figure}[H]
	\centering
	\includegraphics[width=0.9\textwidth]{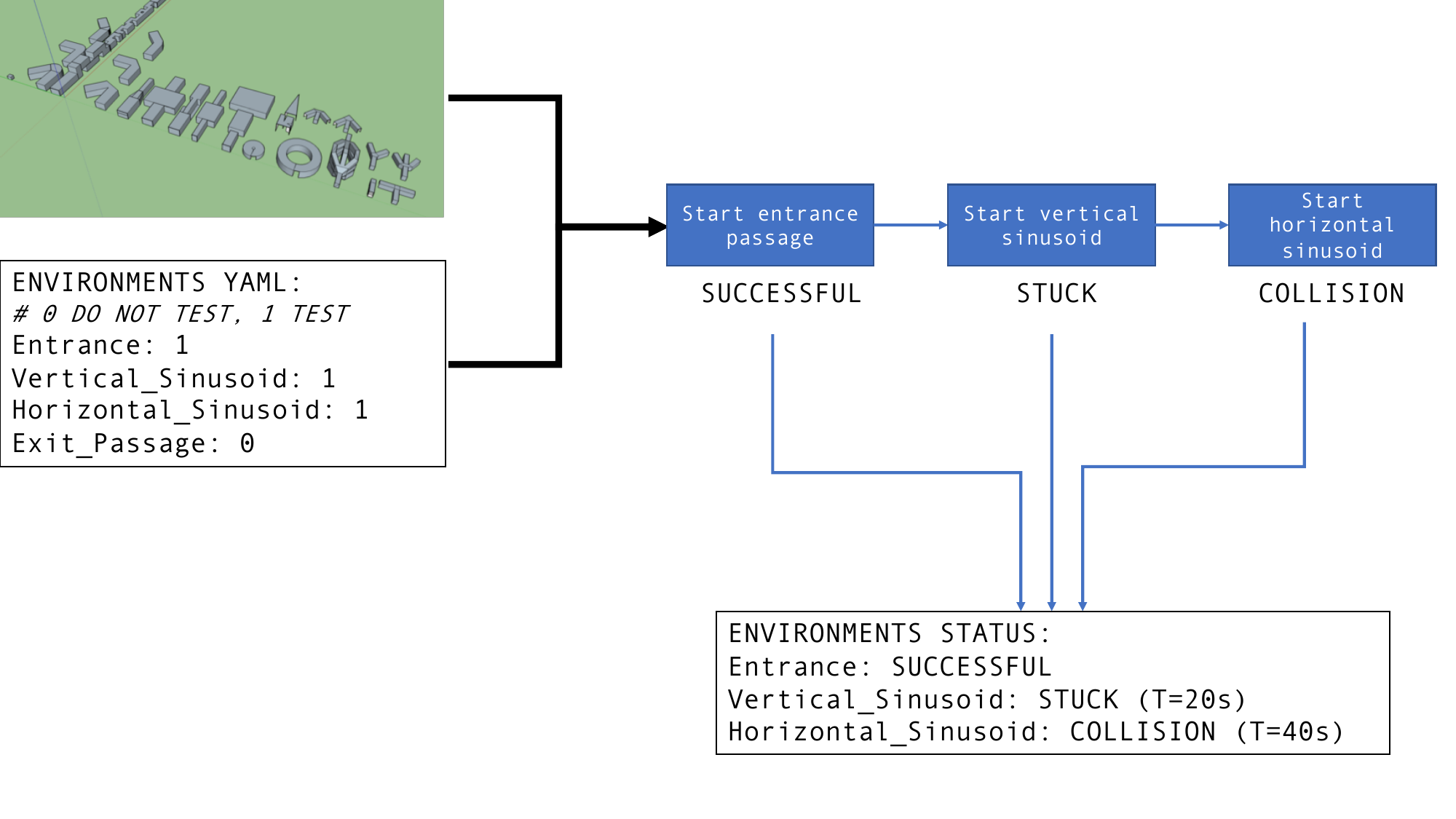}
	\caption{Simulator mulitple-environments process}
	\label{fig:sim_env_process}
\end{figure}

This process is illustrated in Figure \ref{fig:sim_env_process}. A simple test of three environments is exemplified. However, each of the separate entities in the image of the environments at the top left of the image are environments which could be tested.

There are multiple environments, including vertical and horizontal sinusoids, passages with corners and curves of different angles and radii, environments with multiple paths, and corridors of different widths. The extreme variety of environments is to allow the system to be tested in a range of cases. This ensures that the system is not over-tuned to a particular environment that has been tested on hardware. There are also a few environments designed to test particular capabilities of the planning system.

\begin{figure}[H]
	\centering
	\includegraphics[width=0.8\textwidth]{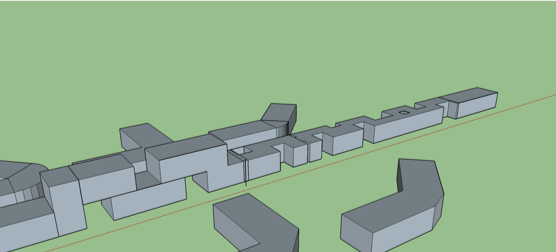}
	\caption{Example environments}
	\label{fig:env_examples}
\end{figure}

The horizontal sinusoid contains a variety of different types of obstacles including doorways, narrow passages, and pillars in the centre of the space. This it to test the wall-following behaviour. The vertical sinusoid is used to test the state machine for vertical motion described in Chapter \ref{chap:LocalPlanner}. The `T' or `Y' shaped environments are designed to test the frontier exploration algorithms. A close up of a few environments are pictured in Figure \ref{fig:env_examples}.

\newpage
\section{Results}
\begin{figure}[H]
	\begin{subfigure}[b]{0.5\textwidth}
	\centering
	\includegraphics[width=\textwidth]{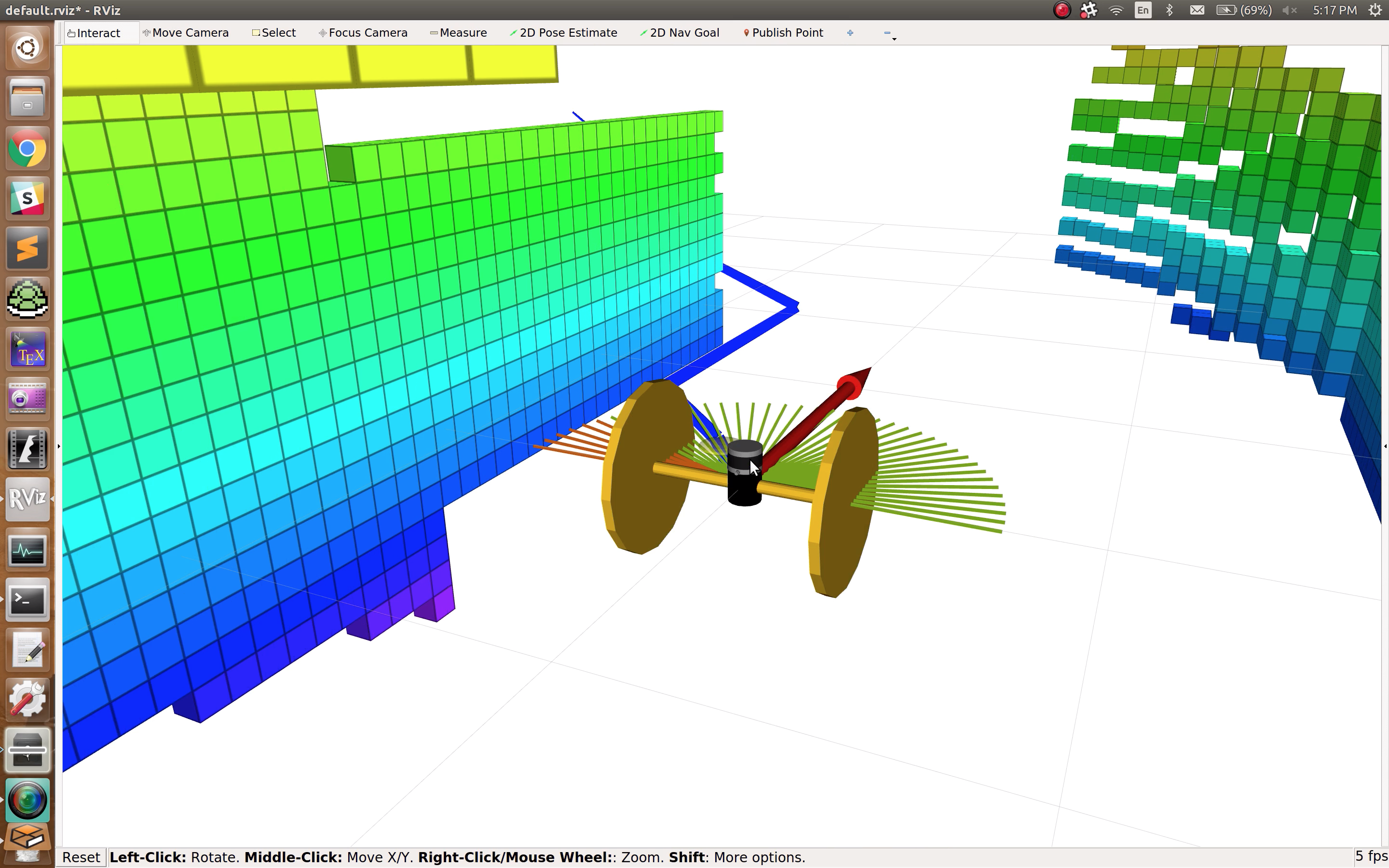}
	\caption{RViz Interface, working with the mapping and planning algorithms}
	\label{fig:sim_rviz}
	\end{subfigure}
\hfill
\begin{subfigure}[b]{0.5\textwidth}
	\centering
	\includegraphics[width=\textwidth]{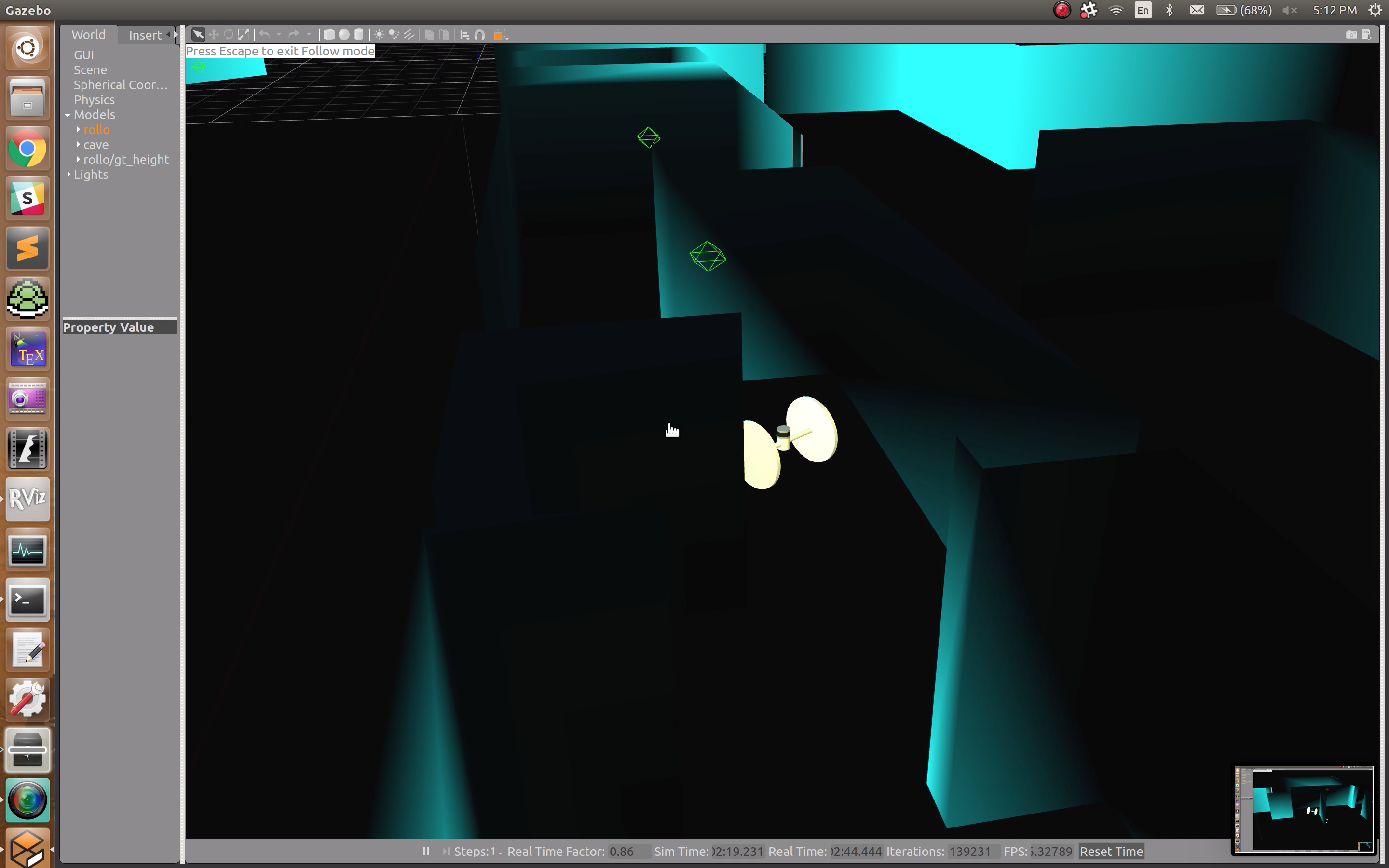}
	\caption{Gazebo interface, horizontal sinusoid, rolling around the corner}
	\label{fig:sim_gazebo}
\end{subfigure}
\caption{The two interfaces for the simulator}
\end{figure}

As this simulator was developed as a tool to be used to testing other subsystems, this section shows some examples of the simulator in use, and provides some detail on what the user experiences. Videos of the simulator in use are available here (\url{https://youtu.be/huvM8Qwed2Q}). 
and here (\url{https://youtu.be/KI8zdz5EZOI}). Figures \ref{fig:sim_rviz} and \ref{fig:sim_gazebo} show the RViz and Gazebo interfaces.

\vspace{-0.5cm}

\begin{figure}[H]
	\begin{subfigure}[b]{0.5\textwidth}
	\centering
	\includegraphics[width=\textwidth]{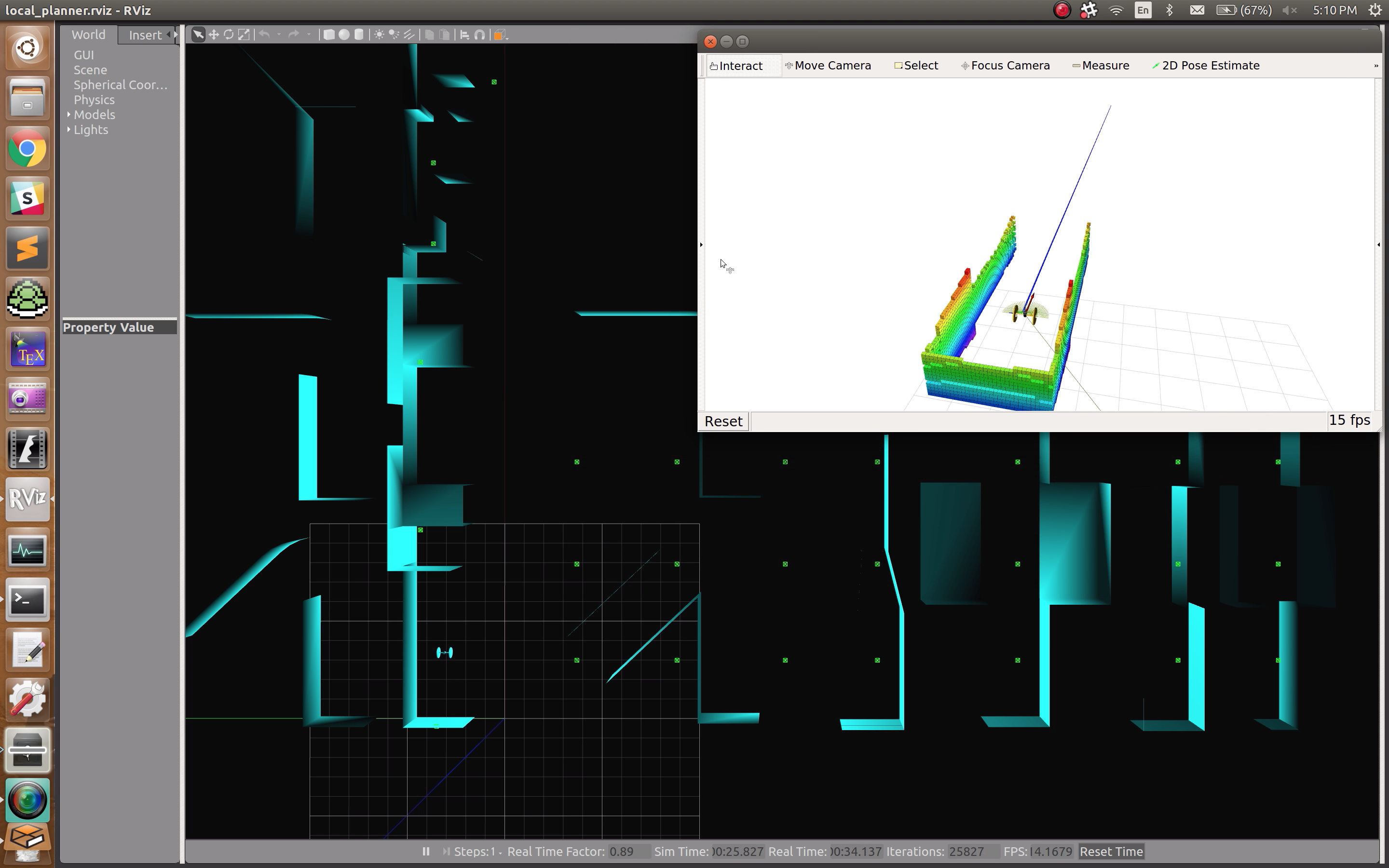}
	\caption{Top-view interface}
	\label{fig:sim_top_gr}
	\end{subfigure}
	\hfill
	\begin{subfigure}[b]{0.5\textwidth}
	\centering
	\includegraphics[width=\textwidth]{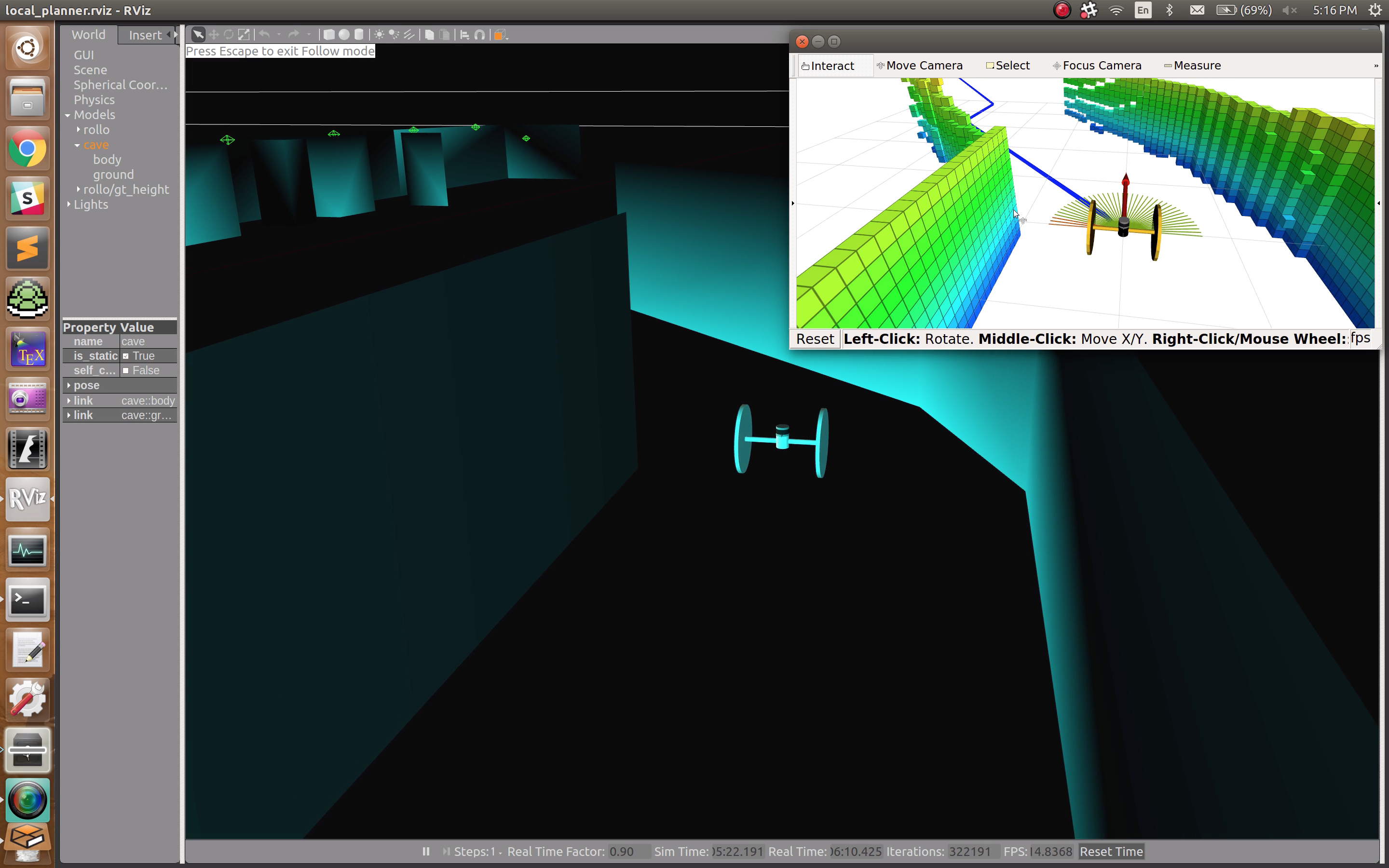}
	\caption{Follow-view interface}
	\label{fig:sim_corner_gr}
	\end{subfigure}
	\caption{The interfaces for the user when running the simulation}
\end{figure}

Figures \ref{fig:sim_top_gr} and \ref{fig:sim_corner_gr} show the views for the user when running the simulator. The top view allows for a heuristic perspective of all the environments while the follow-view allows for the short-term motion of the vehicle to be analysed. 

\section{Summary}
This chapter discussed the development of a simulator which can be used to test the autonomy systems covered in this thesis. While this tool is not directly applicable to the vehicle, it will provide significant aid in the process of ensuring the planning and autonomy algorithms are robust to different types of environments. In addition, creating new environments is a fast process. Hence, a developer can create any environment that can be envisaged and test different aspects of the various algorithms as they are developed. This tool has already been used in improving the robustness of the methods used in navigating without localisation.
	\cleardoublepage
	\chapter{Conclusion and Future Work}

\section{Discussion of Contributions}
A literature review was completed which provided insight into hybrid vehicles, and common methods for local planning, collision avoidance, wall following, and traversability analysis. While many of these methods assume there is reliable localisation, a solution was devised which combined the concepts found in this literature with an idea for navigating without localisation. 

A local planner which is capable of using only instantaneous sensory information to navigate through unknown environments was created. The planner uses a KD-tree search on a 360\degr instantaneous point cloud from the Velodyne LIDAR to generate trajectories in the form of short-term motion primitives and choose the primitives based on a collision avoidance algorithm. This was then extended to support a wall-following behaviour to allow the vehicle to navigate through a variety of unfamiliar passages based on instantaneous information.

The wall following behaviour was first integrated with the ground mobility mode only. Once it had been demonstrated that the vehicle could robustly avoid obstacles and follow the walls, the behaviour was extended to the aerial mobility mode. Initially, this vehicle was using the same method in both modes and simply maintaining a constant height in the aerial mode. To benefit from the enhanced mobility of the aerial mode, this method was adapted to include primitives which would change the height of the vehicle.The algorithm was combined with a local mapper which created a submap including the local area around the vehicle. The planning algorithm could then use the A* global planning method to find a more optimal path towards the goal. This maintained the robustness to localisation failures as the map would be reset if the localisation became unreliable.

Once this map had been created, the algorithm could use the A* planner to determine when to roll or fly. A lower cost was assigned to the nodes on the ground which accounted for the decreased power consumption. This encouraged the behaviour of rolling where possible and flying when there was no possible path that could be rolled along. Hence, the hybrid mobility mode was implemented.

To work with the hybrid mode, a traversability analysis algorithm was developed which would be able to contribute to the cost function based on the traversability of the local terrain. While this effort was abandoned by the project, the algorithm was ready to be integrated with the cost function on the local map.

Finally, a simulator was created which could be used to test and improve the robustness of all the autonomy systems detailed in this thesis. This simulator also contained a state machine capable of testing the planner in multiple environments in one run. A user could set the local planner to attempt a selection of environments and return after a few minutes to observe and analyse the failure modes of the planner.

\section{Comparison to Related Work}
The method used for local planning is based on the method of generating motion primitives and choosing the optimal primitive that was presented in the literature search in Section \ref{sec:lit_rev_mps}. The collision detection method of performing KD-tree searches leads to a lower latency in the processing of the point cloud than a method such as Voxblox\cite{voxbloxoleynikova2017} which involved inserting the point cloud into an existing map. The method used in the local planner identifies the optimal free primitive in less than 10ms. Hence, the updated primitives are published at the frequency of the point clouds. In comparison, at the same resolution of 0.2m, Voxblox runs at a maximum of 20Hz, which is equivalent to up to 50ms latency.

Regarding navigation without localisation, the methods identified in the literature review in Section \ref{sec:lit_rev_wf} include reactive networks\cite{navnolochoward1996} and wall following\cite{wf1roubieu2012,wf2serres2006,wf3santos1995,wf4dwyer2013,wf5van1992,wf6katsev2011}. The approach in this thesis is more reliable than the reactive networks approach because the is no finite list of environments which can be encountered and successfully navigated through. Regarding the wall following methods, the method in this thesis is not limited by the types of walls which can be followed. While the wall following methods in the papers specified above fit planes to the walls, this method follows the walls more passively, leading to a more versatile response. In addition, if there area is empty and there are no obstacles in the vicinity (no walls to follow) the vehicle will continue forward, as is the desired behaviour.

The traversability analysis node was not integrated with the vehicle so the results cannot be compared to other systems with integrated traversability algorithms. The method used is the same as the first found in the literature search in Section \ref{sec:trav_lit_review} which performs statistical analysis on the section of the point cloud which corresponds to the ground. The node in this thesis has successfully differentiated rougher terrain from flat ground. The second method in the literature search used a machine learning approach\cite{ross_thesis} which is able to identify different types of terrain more accurately. For instance, the algorithm in this thesis is unable to differentiate between water and smooth, flat terrain.

The simulator that was developed is a low fidelity version of the simulator that has been developed by DARPA for competitors. The computational load is significantly reduced. This is clear by considering that the low fidelity simulator can run at real time on a standard 4-core Linux computer while the high fidelity simulator crashes on start up on the same computer. This also illustrates the need for the low fidelity simulator which can be used for simple unit testing of planning and autonomy algorithms.

\section{Future work}
The following work could be used to augment the system described in this thesis.

\subsubsection{Force Detection and Proprioceptive Planning}
While this thesis was taking place, another member of the Guidance and Control team was working on external force estimation based on acceleration measurements from the IMU. The idea was to identify the direction in which the force was applied and manoeuvre accordingly. This is applicable when the vehicle is traversing an area which is filled with fog, dust, or smoke. The vehicle will be able to follow the walls and navigate at a lower velocity by colliding with one wall and navigating towards the other wall until colliding with it. A local planner capable of sending open loop commands to the controller was created but has yet to be tested with the force estimates and the hardware. This work also needs to be integrated with the existing local planner architecture.

\subsubsection{Improve the Traversability Analysis}
The traversability analysis algorithm considers the slope and roughness of the point cloud where the ground is expected to appear. This can cause issues in environments were there are bodies of water. At the base of a tunnel, there may be a small lake or puddle which formed over time. This would be recognised as flat terrain by the vehicle but cannot be rolled upon. A method for fixing this would be to use a machine learning approach to identify different aspects of the terrain in the RGB camera video feed and possibly having a downward facing light and checking on the reflectivity of the ground surface.

\subsubsection{Integrate Traversability Mapping with the Planning Architecture}
There is potential for the traversability analysis started in this thesis to be integrated with the local map used for planning. If the traversability map resolution is the same as the resolution of the local map, the cost function could be easily adapted to include the effects of traversability. This would allow for the cost of rolling to be more accurately estimated.

\subsubsection{Integrate the Traversability Analysis with the Controller}
Estimates of the current slope and roughness of the terrain being traversed could be used to control the thrust required. If the surface is more rough or sloped up, more thrust should be used to overcome this. Similarly, if the surface is sloping downwards, the vehicle should use less thrust.

\subsubsection{Landing Site Detection}
This is highly relevant to a hybrid vehicle because the planner will attempt to land when rolling is more efficient than flying. However, it is important to consider the area underneath the vehicle as it lands. As discussed in Section \ref{sec:hybrid_wf}, the bottom height sensor provides a point cloud representation of the area underneath the rollocopter. The traversability analysis code should be integrated with this height sensor to provide a safety check before executing the landing command, to ensure the vehicle does not land on terrain that would damage the vehicle.

\subsubsection{Multi-robot Missions}
The methods discussed in this thesis will be extended to multi-robot missions in which a swarm of robots will perform tasks collaboratively, as required for the challenge. Previously explored methods will be employed for this extension\cite{ali_futurework1,ali_futurework2} to allow the robots to create and localise within a central map of the environment.

\section{Conclusion}
The goal of this thesis was to develop the capability for a hybrid aerial-ground vehicle to robustly navigate without localisation through mines. This work involved the development of a local planner which is capable of navigating with only instantaneous information. This planner has been designed to utilise the full functionality of the hybrid vehicle by having both aerial and ground modes and the ability to transition between these modes in the same test. The robustness of abilities can be improved by testing in a simulator which can provide results in a range of environments, depending on the goals of the test. A planner capable of navigating through any enclosed environment without localisation could be used for many applications beyond the mines and tunnels for which it was developed.

	\cleardoublepage
	\newpage
	\bibliographystyle{aiaa}
	\bibliography{Bibliography}
	\newpage
	\begin{appendices}
		\chapter{Extended Kalman Filter}\label{apdx:ekf_derivation}
\vspace{-1cm}

Algorithm \ref{alg:EKF} details the Extended Kalman Filter. This algorithm is from a previous assignment submission\cite{control_assignment}.

\hspace{-1cm}
\begin{algorithm}[H]
	\begin{mdframed}
		\KwIn{Previous state, measured state, latest input, sample time}
		\KwOut{State estimate}
		\hrule
		Determine the initial estimates of $x$, $\hat{x}$, and $X$\;
		
		\textbf{In a function to be integrated:}\\
		Define the A and C matrices, as normal\;
		Define matrix $B_w$, so as to fit into Equation \ref{eqn:disturbance}, where $w$ is the disturbance\;
		
		\vspace{-1cm}
		
		\begin{align}
		\dot{x} = Ax+B_ww\label{eqn:disturbance}
		\end{align}
		
		The measured ($y$) and modelled($\hat{y}$) state values can then be defined\;
		
		\vspace{-1cm}
		
		\begin{align}
		y = Cx
		\text{\color{white}\large ;;;;; \color{black};\color{white};;;;;\normalsize\color{black}}
		\hat{y} = C\hat{x}
		\end{align}
		Calculate the gain L at this iteration
		\vspace{-1cm}
		
		\begin{align}
		L = XC^TR^{-1}
		\end{align}
		
		Calculate $\dot{x}$, $\dot{\hat{x}}$, and $\dot{X}$ using the gain L, the system model, and the Riccati equation\;
		
		\vspace{-1cm}
		
		\begin{align}
		\dot{x} = Ax+Bu
		\text{\color{white}\large ;;;;; \color{black};\color{white};;;;;\normalsize\color{black}}
		\dot{\hat{x}} = A\hat{x}+Bu + L(y-\hat{y})\\
		\dot{X} = AX+XA^T+B^TQB-XC^TR^{-1}CX		
		\end{align}
		
		Integrate the above (using Runge-Kutta) to get a final estimate of the state at the current time
		
	\end{mdframed}
	\caption{Extended Kalman Filter}\label{alg:EKF}
\end{algorithm}
		\chapter{Newton's Iterative Interior Point Method}\label{apdx:newtons_method}
Algorithm \ref{newtonsapdx} details the method, as in the Advanced Control and Optimisation course notes \cite{notes5520}.

\begin{algorithm}
	 \begin{mdframed}
		 Initialise $t_0 = 0, x_0$ as any interior point.\;
		 \For {a certain number of barrier iterations, indexed by j}
		 {
		 Set 
		 
		 \vspace{-1cm}
		 \begin{align}
		 f_t(x) = t_jcx + \Sigma^m_{i=0} \phi_i(x)
		 \end{align}
		 
		 \For{ a certain number of Newton iterations, indexed by k}
		 {
		 		 Compute gradient\;
		 		 
		 		 \vspace{-1cm}
		 		\begin{align}
		 		g = t_jc + \Sigma^m_{i=0}\frac{\partial\phi_i(x_k)}{\partial x}
		 		\end{align} 
			 Compute Hessian\;
			 
			 \vspace{-1cm}
			 \begin{align}
			  H = \Sigma^m_{i=0}\frac{\partial^2\phi_i(x_k)}{\partial x^2}
			  \end{align}
			 
			 Set
			 
			 \vspace{-1cm}
			\begin{align}
			  x_{k+1} = x_k = H^{-1}g
			\end{align}
		}
		Set $t_{j+1}$ to be a number larger than $t_j$
		}
	 \end{mdframed}
 \caption{Iterative interior point method}
 \label{newtonsapdx}
 \end{algorithm}
		\chapter{Mobility Service Algorithms}\label{apdx:mobility_services}
The mobility services node, covered in Chapter \ref{sec:mobility_services_overview}, uses the following algorithms to generate the goals for the local planner.

\noindent\textbf{Idle Mode}\\
The algorithm for Idle mode is as described in Algorithm \ref{alg:ms_idle}.

\begin{algorithm}[H]
	\begin{mdframed}
		\KwIn{Latest state}
		\KwOut{Goal for the local planner}
		\hrule
		
		Create a goal in the odometry frame\;
		
		\vspace{-1.4cm}
		
		\begin{align}
		x_d = x_{latest};
		y_d = y_{latest};
		z_d = z_{ground}\
		\psi_d = \psi_{latest}
		\end{align}
		\vspace{-0.5cm}
		
		\While{ROS is running and Preempt has not been requested}
		{
			Publish the goal at 30Hz\;
		}
	\end{mdframed}	
	\caption{Idle Mode}
	\label{alg:ms_idle}
\end{algorithm}
\newpage
\noindent\textbf{Hover Mode}\\

\begin{algorithm}[H]
	\begin{mdframed}
		\KwIn{Latest state}
		\KwOut{Goal for the local planner}
		\hrule
				
				Create a goal in the odometry frame\;
				
				\vspace{-1.4cm}
				
				\begin{align}
				x_d = x_{latest};
				y_d = y_{latest};
				z_d = z_{latest};
				\psi_d = \psi_{latest}
				\end{align}
				\vspace{-0.5cm}
				
				\While{ROS is running and Preempt has not been requested}
				{
					Publish the goal at 30Hz\;
				}
	\end{mdframed}	
	\caption{Hover Mode}
	\label{alg:ms_hover}
\end{algorithm}

\noindent\textbf{Take off}\\

\begin{algorithm}[H]
	\begin{mdframed}
		\KwIn{Latest state, desired height}
		\KwOut{Goal for the local planner}
		\hrule
		\While{ROS is running and Preempt has not been requested}
		{
			Create a goal with x and y in the odometry frame and z in the ground frame\;
			Publish the goal\;
			\If{The difference between the desired height above the ground and the bottom clearance is less than the tolerance}
			{
				End while loop and declare takeoff completed\;
			}
			
		}
	\end{mdframed}	
	\caption{Take off Mode}
	\label{alg:ms_take_off}
\end{algorithm}
\newpage
\noindent\textbf{Land}\\

\begin{algorithm}[H]
	\begin{mdframed}
		\KwIn{Latest state}
		\KwOut{Goal for the local planner}
		\hrule
		Create a goal\;
		
		\vspace{-1.4cm}
		\begin{align}
			x_d = x_{latest};
			y_d = y_{latest};
			&z_d = z_{ground};
			\psi_d = \psi_{latest}\\
			z_{control mode} &= VELOCITY\\
			\dot{z}_d &= \dot{z}_{land}
		\end{align}

		\While{ROS is running and Preempt has not been requested}
		{
			Use a low pass filter on the height sensor measurement\;
			\If {The minimum height is less than the bottom clearance}
			{
				Declare that the vehicle has landed\;
			}
		}
		
		Create a new goal with the latest state as the desired state\;
		
	\end{mdframed}	
	\caption{Land Mode}
	\label{alg:ms_land}
\end{algorithm}

\noindent\textbf{Fly to (x, y, z)}\\

\begin{algorithm}[H]
	\begin{mdframed}
		\KwIn{Latest state, desired state}
		\KwOut{Goal for the local planner}
		\hrule
		Create a goal with x and y in the odometry frame and z in the ground frame\;
		Set the goal to be equal to the desired state\;
		\While {ROS is running and Preempt has not been requested}
		{Publish the goal at 30Hz\;}
	\end{mdframed}	
	\caption{Fly to Mode}
	\label{alg:ms_fly_to}
\end{algorithm}
\newpage
\noindent\textbf{Fly Forward}\\

\begin{algorithm}[H]
	\begin{mdframed}
		\KwIn{None}
		\KwOut{Goal for the local planner}
		\hrule
		Create a goal with x and y in the body frame and z in the ground frame\;
		
		\vspace{-1.3cm}
		
		\begin{align}
			x_d = 1;
			y_d = 0;
			z_d = z_{latest}
		\end{align}
		
		\While {ROS is running and Preempt has not been requested}
		{Publish the goal at 30Hz\;}
	\end{mdframed}	
	\caption{Fly Forward Mode}
	\label{alg:ms_fly_fwd}
\end{algorithm}

\noindent\textbf{Drive to (x, y)}\\

\begin{algorithm}[H]
	\begin{mdframed}
		\KwIn{Latest state, desired state}
		\KwOut{Goal for the local planner}
		\hrule
				Create a goal with x and y in the odometry frame and z equal to zero\;
				Set the goal to be equal to the desired state\;
				\While {ROS is running and Preempt has not been requested}
				{Publish the goal at 30Hz\;}
	\end{mdframed}	
	\caption{Drive to Mode}
	\label{alg:ms_drive_to}
\end{algorithm}

\noindent\textbf{Drive Forward}\\

\begin{algorithm}[H]
	\begin{mdframed}
		\KwIn{None}
		\KwOut{Goal for the local planner}
		\hrule
		Create a goal with x and y in the body frame and z in the ground frame\;
		
		\vspace{-1.3cm}
		
		\begin{align}
		x_d = 1;
		y_d = 0;
		z_d = 0
		\end{align}
		\vspace{-0.5cm}
		
		\While {ROS is running and Preempt has not been requested}
		{Publish the goal at 30Hz\;}
	\end{mdframed}	
	\caption{Drive Forward Mode}
	\label{alg:ms_drive_fwd}
\end{algorithm}
		\chapter{Checklist Before Leaving the Lab}\label{apdx:ohs}
\attributions{ This checklist was created based on teachings from the University of Sydney.}
Before leaving the lab in the evening, the last worker in the lab is required to use the following checklist.

\begin{itemize}
	\item Remove all batteries from the charger and store in a LiPo-safe bag
	\item Remove all battery monitors from all the batteries
	\item Turn off robot transmitters and store in a safe place
	\item Ensure all robots are kept in a safe location and powered off
	\item PPE and tools must be returned to their designated locations
	\item Communal work stations must be cleared of clutter
	\item Switch off all lights and lock the door
\end{itemize}

This aims to eliminate the risk of unmonitored batteries or robots causing difficulty and ensures a smooth start in the lab the next morning.
		\chapter{Simulator State Calculation Algorithms}\label{apdx:simulator_algorithms}
The following algorithms detail the procedures and calculations used to calculate the state updates in the low fidelity simulator.

Algorithm \ref{alg:simulator_dynamics_ground} details the calculations for simulation in the ground mode.

\begin{algorithm}[hbpt]
	\begin{mdframed}
		\KwIn{Latest state, desired state}
		\KwOut{Next state}
		\hrule
		$\Delta t = t_{now} - t_{prev}$, T = planning time horizon\; \tcc{calculate the body velocity and change in yaw}
		\If {Ground mobility mode}
		{
			\If{Goal in odometry frame}
			{
				\If {Position control mode}
				{
					\vspace{-0.8cm}
					\begin{align}
						\Delta \psi &= \psi_d - \psi_L\\
						v_{x_d} &= \frac{x_d - x_L}{T} ; v_{y_d} = \frac{y_d - y_L}{T} \\
						v_{x_b} &= v_{x_d}cos(\psi_L) + v_{y_d}sin(\psi_L)
					\end{align}
					\vspace{-1cm}
					
				}
				
				\If {Velocity control mode}
				{
					\vspace{-0.8cm}
					\begin{align}
						\Delta \psi &= \dot{\psi_d} \times \Delta t\\
						v_{x_b} &= v_{x_d}cos(\psi_L) + v_{y_d}sin(\psi_L)
					\end{align}
					\vspace{-1cm}
				}
			}
			
			\If{Goal in gravity-aligned body frame}
			{
				\If {Position control mode}
				{
					\tcc{Not applicable.}
				}
				
				\If {Velocity control mode}
				{
					\vspace{-1cm}
					\begin{align}
					\Delta \psi &= \dot{\psi_d} \times \Delta t\\
					v_{x_b} &= v_{x_d}
					\end{align}
					\vspace{-1cm}
				}
			}\tcc{calculate the new pose}
			\vspace{-1cm}
			\begin{align}
				\psi_{t+\Delta t} &= \psi_L + \Delta \psi\\
				x_{t+\Delta t} &= x_L + x_bcos(\psi_{t+\Delta t})\\
				y_{t+\Delta t} &= y_L + x_bcos(\psi_{t+\Delta t})\\
				z_{t+\Delta t} &= r_{wheel}
			\end{align}
			\vspace{-1cm}
		}
	\end{mdframed}	
	\caption{Dynamics calculations, ground mode}
	\label{alg:simulator_dynamics_ground}
\end{algorithm}

Algorithm \ref{alg:simulator_dynamics_aerial} is used to calculate the next pose when the vehicle is in aerial mode and the goal is in the odometry frame.

\begin{algorithm}[H]
	\begin{mdframed}
		\KwIn{Latest state of x and y, desired state for x and y}
		\KwOut{Next state for x and y}
		\hrule
		dt = $t_{now}$ - $t_{prev}$\;
		\If{Aerial mobility mode}
		{
			\If{Goal in odometry frame}
			{
				\If {Position control mode}
				{
					\vspace{-0.8cm}
					\begin{align}
					x &= x_L + \frac{x_d-x_L}{T} \times \Delta t ; y &= y_L + \frac{y_d-y_L}{T} \times \Delta t
					\end{align}
					\vspace{-1cm}
				}
				
				\If {Velocity control mode}
				{
					\vspace{-0.8cm}
					\begin{align}
					x &= x_L + v_{x_d} \times \Delta t ; y &= y_L + v_{y_d} \times \Delta t
					\end{align}
					\vspace{-1cm}
				}
				
				\If{Negative change in yaw requested}
				{
					\vspace{-0.8cm}
					\begin{align}
						\psi_{t+\Delta t} = \psi_L - dt * \dot{\psi}_o
					\end{align}
					\vspace{-1cm}
					\If{$\psi_{t+\Delta t}<\psi_d$}
					{
						\begin{align}
						\psi_{t+\Delta t}=\psi_d
						\end{align}
					}
				}
				\If{Positive change in yaw requested}
				{
					\vspace{-0.8cm}
					\begin{align}
					\psi_{t+\Delta t} = \psi_L + \Delta t * \dot{\psi}_o
					\end{align}
					\vspace{-1cm}
					\If{$\psi_{t+\Delta t}>\psi_d$}
					{
						\begin{align}
						\psi_{t+\Delta t}=\psi_d
						\end{align}
					}
				}
			}
		}
		\end{mdframed}	
		\caption{X and Y dynamics calculations, aerial mode, goal in the odometry frame}
		\label{alg:simulator_dynamics_aerial}
	\end{algorithm}
	
	\newpage
	Algorithm \ref{alg:simulator_dynamics_aerial2} is used to calculate the next pose when the vehicle is in aerial mode and the goal is in the body frame.
	
	\begin{algorithm}[H]
		\begin{mdframed}
			\KwIn{Latest state of x and y, desired state for x and y}
			\KwOut{Next state for x and y}
			\hrule
			dt = $t_{now} - t_{prev}$\;
			\If{Aerial mobility mode}
			{
				\If{Goal in gravity-aligned body frame}
				{
					\If {Position control mode}
					{
						\tcc{Not applicable.}
					}
					
					\If {Velocity control mode}
					{
							\vspace{-1cm}
							\begin{align}
							x_{t+\Delta t} &= x_L + x_bcos(\psi_L)\\
							y_{t+\Delta t} &= y_L + x_bcos(\psi_L)\\
							z_{t+\Delta t} &= r_{wheel}
						\end{align}
						\vspace{-1cm}

					}
				}
				\If{Negative change in yaw requested}
				{
					\vspace{-0.8cm}
					\begin{align}
					\psi_{t+\Delta t} = \psi_L - \Delta t * \dot{\psi}_o
					\end{align}
					\vspace{-1cm}
					\If{$\psi_{t+\Delta t}<\psi_d$}
					{
						\begin{align}
						\psi_{t+\Delta t}=\psi_d
						\end{align}
					}
				}
				\If{Positive change in yaw requested}
				{
					\vspace{-0.8cm}
					\begin{align}
					\psi_{t+\Delta t} = \psi_L + \Delta t * \dot{\psi}_o
					\end{align}
					\vspace{-1cm}
					\If{$\psi_{t+\Delta t}>\psi_d$}
					{
						\begin{align}
						\psi_{t+\Delta t}=\psi_d
						\end{align}
					}
				}
			}
		\end{mdframed}	
		\caption{X and Y dynamics calculations, aerial mode, goal in the body frame}
		\label{alg:simulator_dynamics_aerial2}
	\end{algorithm}
	
	\newpage
	The Algorithm \ref{alg:simulator_dynamics_aerial_z} is used to calculate the new z coordinate.

	\begin{algorithm}[H]
	\begin{mdframed}
		\KwIn{Latest state of z, desired state for z}
		\KwOut{Next state for z}
		\hrule
	\If {Z goal in body frame}
	{
		\If {Z position control mode}
		{
			\tcc{Not applicable.}
		}
		
		\If{Z velocity control mode}
		{
			\vspace{-0.8cm}
			\begin{align}
				z_{t+\Delta t} = z_L + \frac{z_d}{T} \times \Delta t
			\end{align}
			\vspace{-1cm}
		}
	}
	
	\If {Z goal in odometry frame}
	{
		\If {Z position control mode}
		{
			\If{Negative change in z requested}
			{
				\vspace{-0.8cm}
				\begin{align}
				z_{t+\Delta t} = z_L - \Delta t * \dot{z}_o
				\end{align}
				\vspace{-1cm}
				\If{$z_{t+\Delta t}<z_d$}
				{
					\begin{align}
					z_{t+\Delta t}=z_d
					\end{align}
				}
			}
			\If{Positive change in z requested}
			{
				\vspace{-0.8cm}
				\begin{align}
				z_{t+\Delta t} = z_L + \Delta t * \dot{z}_o
				\end{align}
				\vspace{-1cm}
				\If{$z_{t+\Delta t}>z_d$}
				{
					\begin{align}
					z_{t+\Delta t}=z_d
					\end{align}
				}
			}
		}
		
		\If{Z velocity control mode}
		{
			\vspace{-0.8cm}
			\begin{align}
			z_{t+\Delta t}=z_L + \dot{z}_d \times \Delta t
			\end{align}
			\vspace{-1cm}
		}
	}
	\end{mdframed}	
	\caption{Dynamics calculations, Aerial Mode, Z for the body or odometry frames}
	\label{alg:simulator_dynamics_aerial_z}
\end{algorithm}

If the goal for z is with respect to the distance to the ground, Algorithm \ref{alg:simulator_dynamics_aerial_z_g} is used.

\begin{algorithm}[H]
	\begin{mdframed}
	\If {Z Goal in Ground Frame}
	{
		\If {Z position control mode}
		{
			\If{Negative change in z requested}
			{
				\vspace{-0.8cm}
				\begin{align}
				z_{t+\Delta t} = z_L - \Delta t * \dot{z}_o
				\end{align}
				\vspace{-1cm}
				\If{$z_{t+\Delta t}<z_d + z_{ground}$}
				{
					\begin{align}
					z_{t+\Delta t}=z_d +  z_{ground}
					\end{align}
				}
			}
			\If{Positive change in z requested}
			{
				\vspace{-0.8cm}
				\begin{align}
				z_{t+\Delta t} = z_L + \Delta t * \dot{z}_o
				\end{align}
				\vspace{-1cm}
				\If{$z_{t+\Delta t}>z_d + z_{ground}$}
				{
					\begin{align}
					z_{t+\Delta t}=z_d + z_{ground}
					\end{align}
				}
			}
		}
		
		\If{Z velocity control mode}
		{
			\vspace{-0.8cm}
			\begin{align}
			z_{t+\Delta t}=z_L + \dot{z}_d \times \Delta t
			\end{align}
			\vspace{-1cm}
		}
	}
		\end{mdframed}	
		\caption{Dynamics calculations, Aerial Mode, Z for the ground frame}
		\label{alg:simulator_dynamics_aerial_z_g}
	\end{algorithm}
%
If the goal for z is with respect to the distance to the ground, Algorithm \ref{alg:simulator_dynamics_aerial_z_c} is used.

	\begin{algorithm}[H]
		\begin{mdframed}
	\If {Z Goal in Ceiling Frame}
	{
		\If {Z position control mode}
		{
			\If{Negative change in z requested}
			{
				\vspace{-0.8cm}
				\begin{align}
				z_{t+\Delta t} = z_L - \Delta t * \dot{z}_o
				\end{align}
				\vspace{-1cm}
				\If{$z_{t+\Delta t}<z_{ceiling} = z_d$}
				{
					\begin{align}
					z_{t+\Delta t}=z_{ceiling} = z_d
					\end{align}
				}
			}
			\If{Positive change in z requested}
			{
				\vspace{-0.8cm}
				\begin{align}
				z_{t+\Delta t} = z_L + \Delta t * \dot{z}_o
				\end{align}
				\vspace{-1cm}
				\If{$z_{t+\Delta t}>z_{ceiling} = z_d$}
				{
					\begin{align}
					z_{t+\Delta t}=z_{ceiling} = z_d
					\end{align}
				}
			}
		}
		
		\If{Z velocity control mode}
		{
			\vspace{-0.8cm}
			\begin{align}
			z_{t+\Delta t}=z_L + \dot{z}_d \times \Delta t
			\end{align}
			\vspace{-1cm}
		}
	}
	\end{mdframed}	
	\caption{Dynamics calculations, Aerial Mode, Z for the ceiling frame}
	\label{alg:simulator_dynamics_aerial_z_c}
\end{algorithm}
		\chapter{A* Path Planning Algorithm}\label{apdx:astar}

\vspace{-1.5cm}

Algorithm \ref{alg:Astar} details the path planning method used to generate the global plan within the local map.

	\begin{algorithm}[H]
		\begin{mdframed}
			\KwIn{Starting node, goal location, voxel representation of the environment}
			\KwOut{Path cell array, cost array, predecessor array}
			\hrule
			
			Set all the `Cost to reach' values to $\infty$ except the starting node `Cost to reach', to 0\;
			Set all predecessor nodes to NaN\;
			\hrule
			\While{The end node is unchecked}
			{
				\For{each unchecked node}
				{
					Temporarily save the node costs to reach and the cost to go and respective indices in Q into the empty arrays\;
				}
				Find node with the minimum cost in the list of unsolved nodes\;
				\For {each node, NodeI}
				{
					\If {there is a path form the current node to NodeI}
					{
						\If {the path from the current node to NodeI is less than the current saved cost to NodeI}
						{	
							Update this to be the new cost\;
							Save the predecessor of NodeI to be the current node
						}
					}
				}
			}
			
			Work backwards from the predecessor array to make an array of nodes on the optimal path
		\end{mdframed}
		\caption{A-star path selection algorithm}\label{alg:Astar}
	\end{algorithm}
	\end{appendices}
	
\end{document}